\documentclass{article}
\usepackage{hyphenat}

\usepackage[preprint]{neurips_2026}

\usepackage[utf8]{inputenc} 
\usepackage[T1]{fontenc}    
\usepackage{hyperref}       
\usepackage{url}            
\usepackage{booktabs}       
\usepackage{amsfonts}       
\usepackage{nicefrac}       
\usepackage{microtype}      
\usepackage[table]{xcolor}        

\usepackage{multirow}
\usepackage{rotating}
\usepackage{pgfplots}
\pgfplotsset{compat=1.18}
\usepackage{graphicx}
\usetikzlibrary{arrows.meta,positioning,fit,calc}
\usepackage{wrapfig}
\usepackage[most]{tcolorbox}
\usepackage{lmodern}
\usepackage{amsmath}
\usepackage{array}
\usepgfplotslibrary{groupplots}
\usetikzlibrary{patterns}
\usetikzlibrary{patterns.meta}
\usepackage{subcaption}
\usepackage{colortbl}
\RequirePackage{xspace}
\usepackage{float}
\usepackage{bbm}
\usepackage{booktabs}
\usepackage{makecell}
\usepackage[table]{xcolor}
\usepackage{colortbl}
\usepackage{xfp}
\usepackage{arydshln}

\makeatletter
\DeclareRobustCommand\onedot{\futurelet\@let@token\@onedot}
\def\@onedot{\ifx\@let@token.\else.\null\fi\xspace}

\def\eg{\emph{e.g}\onedot} 

\def\ie{\emph{i.e}\onedot}

\makeatother

\title{VL-DocIR: A Benchmark for Vision-Based Long Document Information Retrieval}

\author{%
  \textbf{%
    Simon Ott\textsuperscript{1,}\thanks{Equal contribution.},~
    Yufan Chen\textsuperscript{1,\ensuremath{\ast},}\thanks{Corresponding author:
      Yufan Chen (\texttt{yufan.chen@kit.edu}).},~
    Ruiping Liu\textsuperscript{1},~
    Junwei Zheng\textsuperscript{1},~
    Jiale Wei\textsuperscript{1}}\\
  \textbf{%
    Di Wen\textsuperscript{1},~
    Kunyu Peng\textsuperscript{1},~
    Jiaming Zhang\textsuperscript{2},~
    Rainer Stiefelhagen\textsuperscript{1}} \\
  \textsuperscript{1}Karlsruhe Institute of Technology
  \quad
  \textsuperscript{2}Hunan University\\
  \texttt{simon.ott@student.kit.edu,yufan.chen@kit.edu}
}

\begin{document}

\maketitle

\begin{abstract}
Vision-based document retrieval has improved rapidly, but current evaluation settings still emphasize short documents and single relevant pages. This leaves an important gap between leaderboard performance and realistic long-document retrieval, where the evidence needed to answer a query may span multiple pages or even multiple documents. We present VL-DocIR, a page-level benchmark for evidence-complete visual long-document retrieval. VL-DocIR is built from 29,641 HTML-born documents rendered into 388,548 page images from Wikipedia, arXiv, PubMed, and SEC proxy statements. It contains 271,760 queries across 23 domains and six evidence structures, covering single-page, within-document multi-page, and cross-document evidence configurations. Evidence annotations are grounded to rendered pages and HTML element identifiers, and queries are filtered with a cleaning pipeline targeting clarity, correctness, closedness, and the absence of explicit layout references. We evaluate leading single-vector and multi-vector visual retrievers together with diagnostic retrieval settings such as document-context scoring, query splitting, reranking, and round-robin ensembles. The strongest evaluated configuration reaches 69.3 nDCGAll@10 and 80.6 RecallAll@10 under structure-balanced Macro evaluation, but source- and evidence-structure breakdowns show that cross-document retrieval remains substantially harder than single-page retrieval, and that document-context scoring is especially helpful for within-document multi-page queries. VL-DocIR exposes these failure modes and provides a more demanding target for visual long-document retrieval. The new dataset and code will be made publicly available at \href{https://yufanchen96.github.io/VL-DocIR/}{VL-DocIR Project}.

\end{abstract}

\section{Introduction}
\label{sec:introduction}

Document retrieval is increasingly moving from OCR-heavy text pipelines to page-image retrieval, due to information lost in plain-text representations for charts, tables, figures, and layouts. This shift has produced strong retrievers for visual documents, while the evaluation remains incomplete. A system can perform well at the page level while still failing on the long documents that matter in practice, where the evidence for a query is distributed across multiple pages, mixed across text and visuals, or split across separate documents.

This gap is especially acute for scientific articles, encyclopedic web pages, and financial reports. These documents are long, visually heterogeneous, and structurally uneven, \ie, relevant information may begin in prose, continue in a table, and be disambiguated by a later figure. Benchmarks that reward retrieving any relevant page do not test whether a system can recover the full evidence set required to answer queries.

We introduce \textbf{VL-DocIR,} a benchmark for vision-based long document retrieval, which contains 29,641 rendered documents, 388,548 page images, and 271,760 queries over 23 domains. Each benchmark item consists of a natural-language query, multiple evidence pages, and the image corpus. The benchmark covers six query types ranging from single-page single-evidence queries to cross-document queries whose evidence spans multiple documents. This structure can evaluate not only overall retrieval quality, but also the performance for distributed evidence. 

With the VL-DocIR benchmark, we have evaluated leading single-vector and multi-vector document retrievers together. 
Cross-document retrieval remains the hardest setting by a wide margin, while document-context scoring is most beneficial when evidence remains inside the same long document. These results show that long document visual retrieval is not only a scale problem, but also a different evaluation regime with different bottlenecks.

Our contributions are fourfold:
\begin{itemize}
    \item We introduce a large-scale vision-based long document retrieval benchmark with 29,641 documents, 388,548 rendered pages, 271,760 queries, and 23 domains. 
    \item We define a benchmark construction pipeline that combines HTML-grounded page rendering, evidence-page configuration, VLM-based query generation, and query cleaning to produce page-level retrieval supervision at scale.
    \item We provide a systematic evaluation of leading visual document retrievers and complementary OCR-based, text-based, and hybrid baselines, together with benchmark-oriented analysis settings, including query splitting, document-context scoring, reranking, and simple ensembles.
    \item We show that multi-vector retrieval is consistently stronger than single-vector retrieval, that document-context scoring is especially effective for within-document multi-page queries, and that cross-document evidence coverage remains limited among the evaluated configurations. 
\end{itemize}

\section{Related Work}
\label{sec:related_work}

\subsection{Vision-based Document Retrieval}
Visual document retrieval has shifted from OCR-first pipelines toward direct page-image embedding. DSE treats screenshots as retrieval units~\cite{ma2024dse}, while ColPali introduced late interaction over visual page embeddings together with the ViDoRe benchmark~\cite{faysse2024colpali}. The ViDoRe line has continued with harder visual retrieval and RAG settings, including ViDoRe V2 and V3~\cite{mace2025vidorev2,loison2026vidore}. Recent visual document retrieval work further studies multilingual and long-document retrieval~\cite{chen2025visrbench,dong2025mmdocir,cho2025m3docvqa,suri2025visdom,wang2025vidorag} and efficient representations~\cite{yan2026colparse,yan2026colchunk}. Broader reasoning-intensive multimodal retrieval benchmarks and pipelines show that retrieval still struggles when visual inputs require reasoning beyond semantic matching~\cite{abdallah2026mmbright,kasem2026marvel}. Visual RAG systems~\cite{yu2025visrag} and recent multimodal retrieval methods~\cite{zhang2024gme,guenther2025jinaembeddingsv4universalembeddingsmultimodal,teiletche2025modernvbert,nomicembedmultimodal2025,huang2025tomoroqwen3,xiao2026metaembed,uzan2026gqr} further strengthen page-image retrieval.

\subsection{Long Document and Visual Document Understanding}
Document VQA benchmarks\cite{mathew2021docvqa,mathew2021infographicvqa,singh2019textvqa,biten2019scene,mishra2019ocrvqa,vanlandeghem2023dude,masry2022chartqa,methani2020plotqa} evaluate answering over document, scene-text, infographic, and chart images. Structured-document QA datasets~\cite{zhu2021tatqa,talmor2021multimodalqa,chen2021websrc,tanaka2023slidevqa} extend the setting to web pages, slides, and mixed textual-visual evidence. MP-DocVQA extends document VQA to multi-page documents~\cite{tito2023mpdocvqa}, while long document benchmarks~\cite{ma2024mmlongbenchdoc,wang2025mmlongbench,deng2025longdocurl} test long-context multimodal understanding, reasoning, and locating. These benchmarks establish that layout, tables, figures, and cross-page evidence matter for downstream understanding.

\subsection{Document AI and Vision-Language Models}
Visual document systems build on Document AI models that combine text, layout, and image signals. LayoutLM series~\cite{xu2020layoutlm,xu2021layoutlmv2,huang2022layoutlmv3} established pretraining objectives for visually rich document understanding, while generative methods~\cite{appalaraju2021docformer,kim2022ocrfree,lee2023pix2struct,blecher2024nougat,wang2024docllm} explore multimodal and OCR-free document understanding modeling. General vision-language models~\cite{radford2021learning,li2023blip2,liu2023visual,bai2023qwenvl,bai2025qwen25vl,bai2025qwen3} provide the perception and language capabilities increasingly used for document parsing, query generation, and visual retrieval.

\subsection{Retrieval and Embedding Benchmarks}
RAG systems~\cite{karpukhin2020dense,xiong2021approximate,formal2021splade,khattab2020colbert,santhanam2022colbertv2,izacard2022unsupervised,lewis2020retrieval} rely on sparse, dense, and late-interaction retrieval foundations. Text retrieval benchmarks~\cite{thakur2021beir, muennighoff2022mteb} broaden evaluation across tasks and domains. VLM2Vec~\cite{jiang2025vlm2vec} extends embedding evaluation to multimodal inputs, and later universal multimodal embedding work~\cite{jiang2024e5v,lin2024mmembed} further adapts MLLMs to retrieval settings. Long-context benchmarks~\cite{yang2018hotpotqa,bai2023longbench} test reasoning over multiple textual contexts. Document-image benchmarks~\cite{dong2025mmdocir, chen2025visrbench, ma2024dse} further shift evaluation toward retrieving evidence. These resources motivate heterogeneous evaluation and expose the limits of single-domain leaderboards.

\section{Benchmark}
\label{sec:benchmark}

VL-DocIR is a page-level document retrieval benchmark for visually-rich long documents. It pairs a corpus of rendered document pages with natural-language questions whose supporting evidence is localized to one or more pages. Unlike standard page-level retrieval evaluations that reward retrieving any relevant page, VL-DocIR measures the evidence-complete retrieval, \ie, a system succeeds only when it retrieves the full set of evidence pages.


\subsection{Task Definition and Stratification}
Each benchmark item is a tuple $(q, E(q), \mathcal{C})$, where $q$ is a natural-language query, $E(q)$ is the set of one or more ground-truth evidence pages, and $\mathcal{C}$ is the full corpus of rendered page images. At evaluation time, the retriever ranks pages from $\mathcal{C}$, and the item is considered evidence-complete at cutoff $k$ only if every page in $E(q)$ appears in the top-$k$ list. 
VL-DocIR stratifies benchmark items along three dimensions, \ie, evidence structure, source provenance, and domain coverage.

\noindent\textbf{Evidence structure.}
Evidence structure defines the retrieval target and forms the task taxonomy with six compact evidence types, \ie, \textbf{Page-Single} for one evidence on one page, \textbf{Page-Multi} for multiple evidences on one page, \textbf{Doc-Adjacent} for adjacent pages, \textbf{Doc-Near} for nearby non-adjacent pages, \textbf{Doc-Far} for far-apart pages in the same document, and \textbf{Cross-Doc} for pages across documents. This taxonomy
separates ordinary page lookup from retrieval settings where the complete evidence set is distributed across pages or documents.

\noindent\textbf{Source provenance.}
Our VL-DocIR benchmark 
aims to evaluate vision-based information retrieval methods on an extensive, diverse document corpus. To this end, we select documents from four data sources: \textit{Wikipedia}, \textit{arXiv}, \textit{PubMed}, and \textit{SEC Proxy Statements}. These sources differ in document length, layout regularity, visual density, and distractor density. Source-level evaluation provides diagnostic slices of retrieval performance under different document regimes rather than merely reporting metadata. 

\noindent\textbf{Domain coverage.}
Domain coverage mainly focuses on the semantic setting of each benchmark query across 23 domains. It is used to analyze whether retrieval performance generalizes across content regimes, not only across evidence structures or data sources. Domain labels prevent the benchmark from being dominated by a single application area, support domain-level failure analysis, and allow future work to report robustness beyond aggregate retrieval scores. Domain-level comparisons should therefore be interpreted with source provenance.



\subsection{Visual Document Database Creation Pipeline}
\label{subsec:document_dataset}
The intention behind creating the visual document database is to build a large corpus of long document images with rich visual content, including figures, tables, formulas, diagrams, and structured layouts. We render HTML-born documents directly so that visual layout and document structure remain available as retrieval signals. The retrieval unit in VL-DocIR is a rendered page image. In the benchmark, models are allowed to use visual, OCR-based, text-only, or hybrid internal representations to return ranked page identifiers from the same corpus. In this paper, our primary experiments focus on vision-based retrievers. 


\noindent\textbf{Document Index Creation.}
We first build a source-specific document index that records the source, document identifier, domain label, and collection metadata needed for HTML document rendering and evidence grounding. The indexing stage is designed to provide broad domain coverage, related-document neighborhoods for cross-document question generation, and documents that can be rendered with
element-level evidence alignment. The collection strategy is source-specific, \eg,  Wikipedia is expanded from seed articles through linked-article selection, arXiv combines survey-seed expansion with category-level sampling to ensure enough papers, PubMed uses open-access articles with license metadata, and SEC uses one recent proxy statement per
selected company to reduce year-to-year duplication. Domain labels are assigned before rendering and propagated to pages and generated questions.


\noindent\textbf{Document Rendering and Page Construction.}
We render documents with a headless Chrome browser using Playwright\footnote{\url{https://playwright.dev/}}. Before rendering, we trigger lazy-loaded content, remove invisible HTML elements, remove source-specific clutter such as banners, inject identifiers into HTML elements, and record their bounding boxes. These identifiers and bounding boxes are later used to map VLM-selected evidence elements to rendered pages. We then split each full-document screenshot into page images of $2000 \times 4000$ pixels, using a rendering scale factor of 2 for sharper images. A naive fixed-height cut can split tables, figures, or text blocks across pages. To reduce such artifacts, the splitter uses whitespace and bounding-box heuristics to choose page boundaries. When necessary, it adds whitespace at the bottom of a page to move cut-off content to the next page. The stored bounding boxes are updated after this adjustment so that the rendered image and element coordinates remain aligned.

\begin{wraptable}[9]{r}{0.4\textwidth}
\vspace{-0.8em}
\centering
\small
\setlength{\tabcolsep}{4.5pt}
\renewcommand{\arraystretch}{0.95}
\begin{tabular}{@{}lrr@{}}
\toprule
\textbf{Source} & \textbf{Docs} & \textbf{Avg. Pages} \\
\midrule
Wikipedia  & 11{,}988 & 6.49  \\
arXiv      & 10{,}846 & 15.50 \\
PubMed     & 4{,}815  & 12.26 \\
SEC Proxy  & 1{,}992  & 41.97 \\
\midrule
\textbf{Total} & \textbf{29{,}641} & \textbf{13.11} \\
\bottomrule
\end{tabular}
\caption{Benchmark statistics.}
\label{tab:data_source_document_count_and_average_length}
\vspace{-5.0em}
\end{wraptable}

\noindent\textbf{Visual Document Data Statistics.}
The final document dataset contains 29{,}641 documents and 388{,}548 rendered pages. As shown in Tab.~\ref{tab:data_source_document_count_and_average_length}, Wikipedia
contributes many short documents, with an average length of 6.49 pages. arXiv and PubMed documents are longer, averaging 15.50 and 12.26 pages, respectively. SEC proxies are fewer in number but much longer, averaging 41.97 pages. This source-level length variation allows VL-DocIR to distinguish ordinary page lookup from retrieval in long and structurally repetitive documents. The domain distribution is broad but not uniform. The largest domains include Astrophysics, Science, and Sports, while smaller domains remain represented for evaluation.


\subsection{Benchmark Query Creation Pipeline}
\label{subsec:question_dataset}

\begin{figure}[htbp]
    \centering
    \includegraphics[
        trim=2cm 4cm 2cm 4cm,
        clip,
        width=1.0\linewidth
    ]{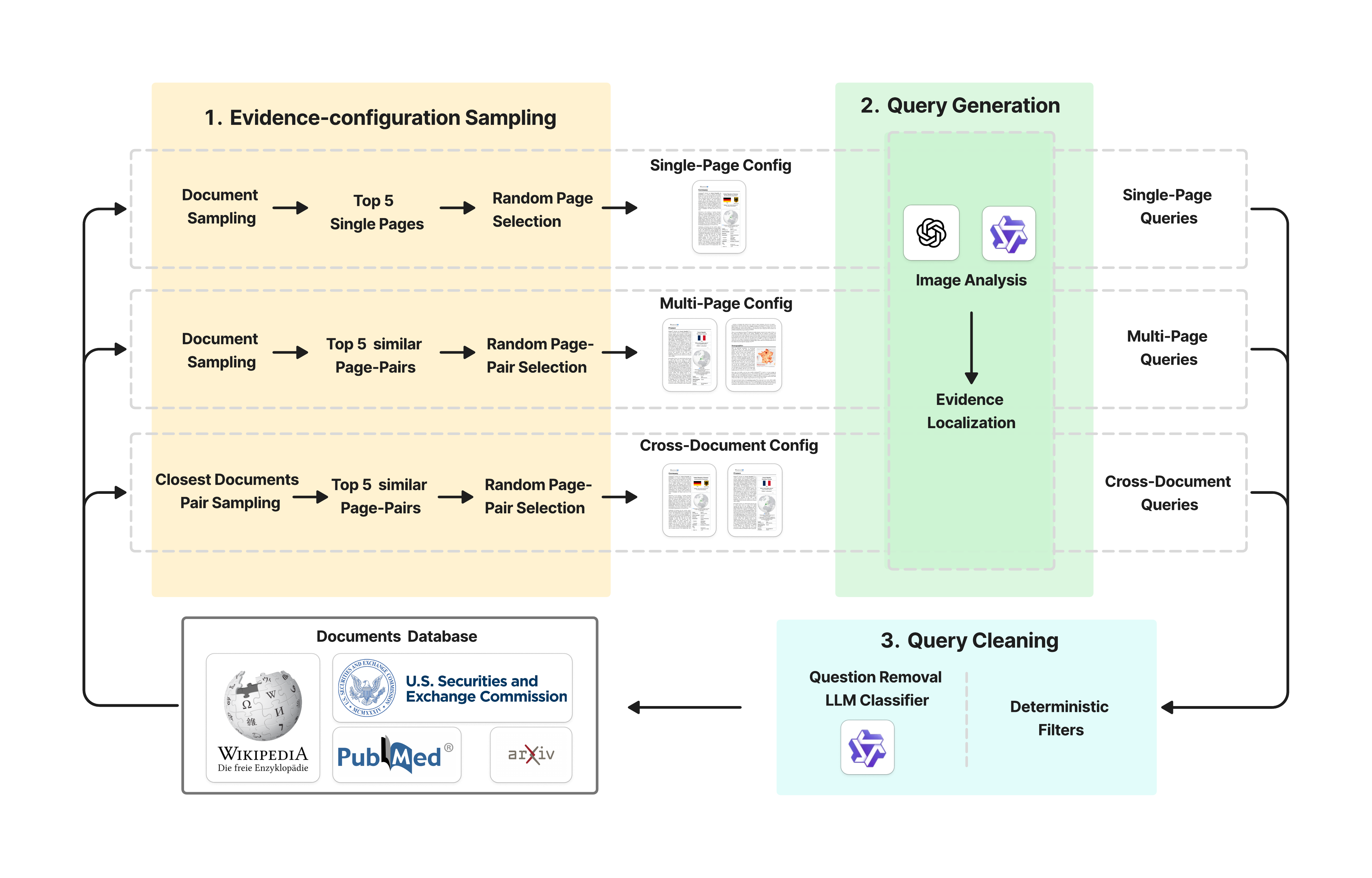}
    \caption[Benchmark Query Creation Process]{Benchmark query creation process overview.}
    \label{fig:questions_dataset_creation_process}
\end{figure}

Our benchmark query creation pipeline consists of three stages: evidence-configuration sampling, VLM-based query generation, and query cleaning. Figure~\ref{fig:questions_dataset_creation_process} summarizes the pipeline. The goal is to generate natural-language queries whose evidence can be localized to one or more rendered document pages.

\noindent \textbf{Evidence-configuration Sampling.}
The first stage selects the evidence pages used to generate queries. The sampling strategy depends on the evidence structure. For \textit{Page-Single} and \textit{Page-Multi} queries, we sample from the top five pages of a document according to a visual-content heuristic score. This biases generation toward visually rich pages while still allowing text-based evidence. For \textit{Page-Adjacent}, \textit{Page-Near}, and \textit{Page-Far} queries, we further select candidate page pairs using TF-IDF cosine similarity~\cite{tfidf} and add a small content-richness term to avoid low-content pages. For \textit{Cross-Doc} queries, we first pair nearby documents using text embeddings of the document start text, and then apply the same page-pair selection strategy as in the within-document setting.

\noindent \textbf{Query Generation.}
We generate the queries and corresponding answers based on the document pages with the help of Vision Language Models (VLMs). For each selected evidence configuration, the pipeline first creates a detailed analysis of the relevant page images. A VLM then receives the rendered page images, the underlying HTML with injected element identifiers, and the page-image analysis. The prompt asks the model to produce queries, answers, and evidence element identifiers. Prompts are specialized by source and
evidence structure so that single-page, multi-page, and cross-document queries follow different generation constraints. After generation, we resolve valid HTML element identifiers to page-relative bounding boxes and evidence pages. Invalid generations, including malformed JSON outputs or invalid element identifiers, would be discarded.

\noindent \textbf{Query Cleaning.}
Since VLM-generated queries can be ambiguous, underspecified, or incorrectly grounded, we apply a cleaning pipeline before constructing the final benchmark set. We evaluate each query using four criteria, \ie, query clarity, answer correctness or plausibility when direct verification is unavailable, closedness, and absence of explicit layout references such as ``in the table'' or ``in the figure''. We use LLM-generated criterion scores and an LLM-based binary removal classifier as the final filtering strategy. To evaluate the removal strategy, we manually annotate a query-cleaning evaluation set with 1{,}438 samples. As an additional deterministic filter, multi-page and Cross-Doc queries must contain evidence across different pages. After cleaning, the dataset is reduced from 388{,}389 generated queries to 271{,}760 final benchmark queries.

\setlength{\columnsep}{0.6em}
\begin{wraptable}[10]{r}{0.54\linewidth}
\vspace{-0.8em}
\centering
\scriptsize
\setlength{\tabcolsep}{2.0pt}
\renewcommand{\arraystretch}{0.90}
\begin{tabular}{l | r r r r | r}
\toprule
\multicolumn{1}{l}{}
& \multicolumn{4}{c}{Data Source}
& \\
\cmidrule(lr){2-5}
Evidence Structure 
& Wiki 
& arXiv 
& PubMed 
& SEC 
& Total \\
\midrule
Page-Single 
& 26,306 
& 14,027 
& 14,128 
& 31,735 
& 86,196 \\

Page-Multi
& 34,222 
& 26,990 
& 29,620 
& 29,792 
& 120,624 \\


Doc-Adjacent
& 3,283 
& 2,556 
& 3,360 
& 2,773 
& 11,972 \\

Doc-Near
& 1,172 
& 634 
& 1,000 
& 586 
& 3,392 \\

Doc-Far 
& 255 
& 461 
& 613 
& 1,234
& 2,563 \\

Cross-Doc 
& 14,598 
& 9,001 
& 9,743 
& 13,671 
& 47,013 \\
\midrule
Total
& 79,836
& 53,669
& 58,464
& 79,791
& 271,760 \\
\bottomrule
\end{tabular}
\caption{Benchmark Query Distribution Overview.}
\label{tab:final_question_count_distribution}
\vspace{-1.0em}
\end{wraptable}

\noindent\textbf{Benchmark Query Statistics.} 
Table~\ref{tab:final_question_count_distribution} shows the final query distribution by source and evidence configuration. The dataset contains 206{,}820 single-page queries and 64{,}940 distributed-evidence queries. This mixture preserves a large base of page-level retrieval examples while explicitly testing whether systems can recover complete evidence sets that span multiple pages or documents.

\subsection{Annotation Quality and Adjudication}
\label{subsec:annotation_quality}

We conducted a stratified manual review of 800 entries covering all combinations of data source and evidence structure. For each source, we sampled 50 Cross-Doc entries and 30 entries for each of the other five structures to evaluate query construction and the correctness and necessity of the annotated evidence pages. Overall, 746 entries were entirely correct, yielding an observed accuracy of 93.25\%, while the corresponding accuracy for Cross-Doc entries was 88.5\%. Additionally, we evaluated the consistency between page mappings and bounding boxes for all 47,013 Cross-Doc queries and their 100,504 evidence elements. For 84.23\% of the queries, the stored and configured page assignments matched perfectly with the assignments derived from the bounding boxes.

\section{Experiment}
\label{sec:experiment}
\definecolor{viridisA}{RGB}{68,1,84}
\definecolor{viridisB}{RGB}{33,145,140}
\definecolor{viridisC}{RGB}{253,231,37}

\definecolor{bestcell}{RGB}{0,0,255}
\definecolor{secondbestcell}{RGB}{33,145,140}
\definecolor{thirdbestcell}{RGB}{0,0,0}

Our experiments are designed for three purposes: 
\begin{itemize}
    \item How current visual retrievers perform when all evidence pages must be retrieved.
    \item Whether multi-vector page-image matching improves over single-vector embedding.
    \item For which retrieval-stage modifications help evidence coverage.
\end{itemize}


\subsection{Experimental Setup}

\noindent \textbf{Evaluation Metrics.} Standard Recall@k and nDCG@k \cite{jarvelin2002} are informative for page-level relevance, but they can
overstate success for distributed-evidence queries. For example, retrieving one of two required evidence pages is useful, but it is not sufficient for answering the item. We therefore report evidence-complete variants, \textbf{RecallAll@k} and \textbf{nDCGAll@k}.

For a query $q$, let $\pi_k(q)=(p_1,\ldots,p_k)$ be the ranked top-$k$ list returned by a retriever, and let $R_k(q)=\{p_1,\ldots,p_k\}$ be the corresponding set of retrieved pages.
Let $E(q)$ denote the set of ground-truth evidence pages. We define the evidence-completion indicator

\begin{equation}
    c_k(q) = \mathbbm{1}\!\left[E(q) \subseteq R_k(q)\right].
\end{equation}

RecallAll@k is the item-level completion indicator:
\begin{equation}
    \mathrm{RecallAll@}k(q) = c_k(q).
\end{equation}
For nDCGAll@k, we first compute binary-relevance nDCG@k over the ranked list $\pi_k(q)$, where evidence pages have gain 1, and all other pages have gain 0:
\begin{align}
    \mathrm{DCG@}k(q) &= \sum_{j=1}^{k}
    \frac{\mathbbm{1}[p_j \in E(q)]}{\log_2(j+1)}, \\
    \mathrm{IDCG@}k(q) &= \sum_{j=1}^{\min(|E(q)|,k)}
    \frac{1}{\log_2(j+1)}, \\
    \mathrm{nDCG@}k(q) &= \frac{\mathrm{DCG@}k(q)}{\mathrm{IDCG@}k(q)} .
\end{align}
The evidence-complete version is
\begin{equation}
    \mathrm{nDCGAll@}k(q) = c_k(q) \cdot \mathrm{nDCG@}k(q).
\end{equation}

\noindent\textbf{Aggregation protocol.}
Our primary aggregate is structure-balanced Macro, which gives
equal weight to the six evidence structures. For an item-level
metric $m(q)$ and the query set $Q_e$ of structure $e$, we compute
\begin{equation}
\mathrm{Macro}(m)
=
\frac{100}{6}\sum_{e=1}^{6}
\frac{1}{|Q_e|}\sum_{q\in Q_e}m(q).
\end{equation}
Micro is the mean over all queries multiplied by 100 and is reported
as a secondary aggregate. Table~\ref{tab:overall_metric_summary}
presents both aggregates at $k\in\{3,5,10\}$ for the same retrieval
runs. 


\noindent \textbf{Evaluated Models.}
We evaluate the following information retrieval methods:
\begin{itemize}
    \item \textbf{Single-Vector Visual Models:} GME Qwen2-VL~\cite{zhang2024gme}, Jina v4 Single-Vector~\cite{guenther2025jinaembeddingsv4universalembeddingsmultimodal};
    \item \textbf{Multi-Vector Visual Models:} ColPali v1.3~\cite{faysse2024colpali}, ColQwen2.5~\cite{faysse2024colpali}, ColSmol500~\cite{faysse2024colpali}, Jina v4 Multi-Vector~\cite{guenther2025jinaembeddingsv4universalembeddingsmultimodal}, ColModernVBERT~\cite{faysse2024colpali}, Colnomic 7B~\cite{nomicembedmultimodal2025}, Tomoro ColQwen3~\cite{huang2025tomoroqwen3};
    \item \textbf{Text-based Models:} Tesseract 5.5.1~\cite{tesseract} with BM25~\cite{bm25} and BGE-M3~\cite{bge-m3}, DOM with BM25~\cite{bm25};
    \item \textbf{Hybrid Methods:} Colnomic 7B~\cite{nomicembedmultimodal2025} with BM25~\cite{bm25} by equal-weight Reciprocal Rank Fusion (RRF)~\cite{RRF}.
\end{itemize}
All models are retrieved from the same corpus and receive only the query at evaluation.

\noindent \textbf{Diagnostic retrieval settings.} In addition to base retrievers, we evaluate four retrieval-stage diagnostic settings. These are used to characterize which failure modes current retrievers expose on our VL-DocIR benchmark.
\begin{itemize}
    \item \emph{Query Splitting} decomposes a query into sub-queries with an LLM and retrieves pages for each sub-query independently. We merge sub-query ranked lists in round-robin order while removing duplicate pages. This setting tests whether explicit query decomposition helps when a question requires evidence from multiple pages or documents.
    \item \emph{Document-context scoring} augments page-level late interaction with a document-level context score. For each document, we cluster its page embeddings into a fixed number of document-context vectors using spherical $k$-means. At query time, a page $p$ from document $d$ receives the score
    \begin{equation}
    s(q,p,d)
    =
    \underbrace{
    \sum_{i=1}^{N_q}
    \max_{j \leq N_p}
    \left\langle E_q^{(i)}, E_p^{(j)} \right\rangle
    }_{s_{\mathrm{page}}(q,p)}
    +
    \underbrace{
    \sum_{i=1}^{N_q}
    \max_{j \leq N_d}
    \left\langle E_q^{(i)}, E_d^{(j)} \right\rangle
    }_{s_{\mathrm{doc}}(q,d)} .
    \end{equation}
    Here $N_q$, $N_p$, and $N_d$ denote the number of query vectors, page vectors, and document-context vectors. $E_q^{(i)}$, $E_p^{(j)}$, and $E_d^{(j)}$ are normalized query, page, and document-context embeddings.
    \item \emph{Reranking} retrieves the top 25 pages with a multi-vector retriever and reranks these candidates with Jina Reranker M0~\cite{jina2025jinarerankerm0}.
    \item \emph{Round-robin ensembles} combine ranked lists from multiple settings by alternating between lists and skipping duplicate pages.
\end{itemize}

\subsection{Results and Analysis}
We organize the results into four diagnostic views. First, we compare representative retrievers across standard and evidence-complete metrics to quantify the overall performance gap. Second, we analyze retrieval-stage settings, including query splitting, document-context scoring, reranking, and round-robin ensembles. Third, we break down the average result by source and evidence structure to expose the long-document stress cases. Fourth, we aggregate performance by evidence structure to show the transition from single-page lookup to distributed-evidence retrieval. The full table over all models, metrics, and $k \in \{3,5,10\}$ is reported in the Appendix.

\paragraph{Overall performance.}
Table~\ref{tab:overall_metric_summary} shows that evidence-complete
retrieval remains substantially harder than ordinary page-level
relevance. The comparison between Jina V4 single-vector and
multi-vector variants shows that the multi-vector version improves
nDCGAll@10 and RecallAll@10 by 11.6 and 10.5 percentage points
under Macro, respectively (12.9 and 10.0 under Micro), suggesting
that token-level or patch-level matching is important for recovering
complete evidence sets. Among base visual retrievers, ColQwen3 and
ColNomic are the strongest, while the strongest evaluated
configuration, ColNomic (Ensemble), reaches 69.28 nDCGAll@10 and
80.64 RecallAll@10 under Macro (75.64 and 86.23 under Micro).
The gain comes from combining page-local matching, document
context, and reranking within the existing retrieval backbone.
Among the text baselines, OCR-BM25 achieves the highest Macro
RecallAll@10 of 61.69. Its fusion with ColNomic (Ctx) increases
Macro RecallAll@10 from 78.19 to 78.67, while reducing
nDCGAll@10 from 66.03 to 65.13.

\begin{table*}[t]
\centering
\resizebox{\textwidth}{!}{
\setlength{\tabcolsep}{2.0pt}
\renewcommand{\arraystretch}{0.95}
\begin{tabular}{@{}llrrr rrr rrr rrr@{}}
\toprule
\multirow{2}{*}{Model} & \multirow{2}{*}{Aggregate}
& \multicolumn{3}{c}{nDCG@k} & \multicolumn{3}{c}{Recall@k}
& \multicolumn{3}{c}{nDCGAll@k} & \multicolumn{3}{c}{RecallAll@k} \\
\cmidrule(lr){3-5}\cmidrule(lr){6-8}\cmidrule(lr){9-11}\cmidrule(l){12-14}
& & @3 & @5 & @10 & @3 & @5 & @10 & @3 & @5 & @10 & @3 & @5 & @10 \\
\midrule
\multirow{2}{*}{ColSmol 500} & Macro & 49.43 & 53.36 & 56.26 & 52.89 & 61.04 & 68.62 & 35.30 & 42.11 & 47.33 & 39.01 & 49.10 & 58.15 \\
 & Micro & 54.98 & 57.76 & 59.95 & 61.35 & 67.70 & 73.97 & 50.63 & 53.87 & 56.58 & 57.03 & 63.45 & 69.81 \\
\multirow{2}{*}{ColModernVBERT} & Macro & 47.33 & 50.91 & 53.72 & 50.48 & 57.93 & 65.30 & 33.28 & 39.31 & 44.29 & 36.73 & 45.70 & 54.35 \\
 & Micro & 51.69 & 54.39 & 56.55 & 57.73 & 63.92 & 70.11 & 47.45 & 50.57 & 53.20 & 53.51 & 59.73 & 65.96 \\
\multirow{2}{*}{GME Qwen2-VL} & Macro & 55.27 & 59.51 & 62.72 & 59.35 & 68.10 & 76.42 & 41.08 & 48.82 & 55.10 & 45.33 & 56.70 & 67.34 \\
 & Micro & 60.18 & 63.16 & 65.44 & 67.20 & 73.94 & 80.35 & 55.34 & 59.01 & 62.11 & 62.37 & 69.38 & 76.22 \\
\midrule
\multirow{2}{*}{Jina V4 SV} & Macro & 48.31 & 52.61 & 55.80 & 52.40 & 61.33 & 69.69 & 34.70 & 42.09 & 48.03 & 38.80 & 49.86 & 60.24 \\
 & Micro & 52.55 & 55.75 & 58.24 & 59.58 & 66.87 & 74.01 & 48.17 & 51.93 & 55.10 & 55.16 & 62.60 & 70.00 \\
\multirow{2}{*}{Jina V4 MV} & Macro & 60.92 & 64.99 & 67.79 & 64.43 & 72.78 & 79.99 & 45.69 & 53.56 & 59.59 & 49.61 & 60.92 & 70.77 \\
 & Micro & 67.24 & 69.83 & 71.74 & 73.36 & 79.17 & 84.47 & 61.80 & 65.15 & 67.98 & 68.01 & 74.16 & 80.03 \\
\midrule
\multirow{2}{*}{ColQwen3} & Macro & 65.54 & 69.09 & 71.36 & 68.15 & 75.35 & 81.13 & 50.39 & 57.86 & 63.20 & 53.74 & 64.11 & 72.42 \\
 & Micro & 72.54 & 74.53 & 75.89 & 77.44 & 81.80 & 85.49 & 67.13 & 69.92 & 72.15 & 72.21 & 77.00 & 81.24 \\
\midrule
\multirow{2}{*}{ColPali v1.3} & Macro & 42.03 & 45.47 & 48.26 & 45.04 & 52.18 & 59.52 & 28.93 & 34.55 & 39.41 & 32.14 & 40.58 & 49.13 \\
 & Micro & 46.10 & 48.74 & 50.92 & 51.76 & 57.80 & 64.08 & 42.21 & 45.20 & 47.85 & 47.87 & 53.90 & 60.26 \\
\multirow{2}{*}{ColPali v1.3 (Rerank)} & Macro & 57.25 & 59.73 & 60.92 & 58.96 & 63.94 & 66.94 & 43.30 & 48.42 & 51.12 & 45.88 & 52.93 & 57.16 \\
 & Micro & 62.97 & 64.26 & 64.95 & 66.38 & 69.19 & 71.03 & 58.31 & 60.18 & 61.33 & 61.93 & 65.05 & 67.17 \\
\midrule
\multirow{2}{*}{ColQwen2.5} & Macro & 58.68 & 62.56 & 65.15 & 62.08 & 70.02 & 76.72 & 44.15 & 51.48 & 56.72 & 47.93 & 58.51 & 67.23 \\
 & Micro & 64.94 & 67.41 & 69.21 & 70.84 & 76.37 & 81.40 & 59.84 & 62.97 & 65.51 & 65.82 & 71.63 & 77.04 \\
\multirow{2}{*}{ColQwen2.5 (Rerank)} & Macro & 65.82 & 69.22 & 71.15 & 68.55 & 75.45 & 80.31 & 51.33 & 58.45 & 62.88 & 54.69 & 64.60 & 71.62 \\
 & Micro & 72.61 & 74.49 & 75.65 & 77.26 & 81.38 & 84.50 & 67.33 & 70.02 & 71.94 & 72.14 & 76.73 & 80.36 \\
\midrule
\multirow{2}{*}{ColNomic} & Macro & 63.85 & 67.61 & 70.05 & 66.84 & 74.53 & 80.78 & 48.73 & 56.38 & 61.82 & 52.36 & 63.06 & 71.76 \\
 & Micro & 70.46 & 72.75 & 74.35 & 75.95 & 81.05 & 85.46 & 65.06 & 68.17 & 70.68 & 70.71 & 76.24 & 81.24 \\
\multirow{2}{*}{ColNomic (Ctx)} & Macro & 64.85 & 69.37 & 72.57 & 68.26 & 77.51 & 85.72 & 49.33 & 58.64 & 66.03 & 53.28 & 66.31 & 78.19 \\
 & Micro & 69.75 & 72.51 & 74.48 & 75.70 & 81.88 & 87.38 & 63.50 & 67.15 & 70.07 & 69.69 & 76.28 & 82.26 \\
\multirow{2}{*}{ColNomic (Rerank)} & Macro & 66.82 & 70.50 & 72.76 & 69.77 & 77.24 & 83.00 & 52.25 & 59.84 & 65.00 & 55.77 & 66.37 & 74.62 \\
 & Micro & 73.77 & 75.87 & 77.29 & 78.72 & 83.33 & 87.15 & 68.44 & 71.42 & 73.76 & 73.52 & 78.63 & 83.11 \\
\multirow{2}{*}{ColNomic (Ensemble)} & Macro & \textbf{68.42} & \textbf{72.68} & \textbf{75.45} & \textbf{72.03} & \textbf{80.67} & \textbf{87.73} & \textbf{53.58} & \textbf{62.60} & \textbf{69.28} & \textbf{57.43} & \textbf{70.04} & \textbf{80.64} \\
 & Micro & \textbf{75.00} & \textbf{77.39} & \textbf{79.03} & \textbf{80.59} & \textbf{85.85} & \textbf{90.30} & \textbf{69.43} & \textbf{72.89} & \textbf{75.64} & \textbf{75.02} & \textbf{80.95} & \textbf{86.23} \\
\midrule
\multirow{2}{*}{OCR-BM25} & Macro & 51.30 & 55.40 & 58.47 & 54.99 & 63.50 & 71.53 & 37.56 & 44.59 & 50.24 & 41.45 & 51.91 & 61.69 \\
 & Micro & 56.06 & 59.16 & 61.53 & 62.70 & 69.76 & 76.48 & 51.55 & 55.20 & 58.26 & 58.18 & 65.36 & 72.34 \\
\multirow{2}{*}{OCR-dense (BGE-M3)} & Macro & 22.90 & 24.97 & 26.59 & 24.58 & 28.90 & 33.16 & 14.14 & 16.92 & 19.35 & 15.92 & 20.26 & 24.70 \\
 & Micro & 23.37 & 24.95 & 26.24 & 26.66 & 30.28 & 34.00 & 20.81 & 22.45 & 23.86 & 24.10 & 27.55 & 31.12 \\
\multirow{2}{*}{DOM-BM25} & Macro & 47.81 & 51.49 & 54.34 & 51.12 & 58.75 & 66.19 & 35.24 & 41.67 & 46.96 & 38.82 & 48.37 & 57.44 \\
 & Micro & 51.11 & 53.86 & 56.00 & 57.07 & 63.34 & 69.41 & 47.04 & 50.31 & 53.07 & 53.01 & 59.44 & 65.72 \\
\midrule
ColNomic (Ctx) with & Macro & 62.88 & 67.70 & 71.23 & 67.11 & 77.01 & 86.10 & 47.87 & 57.14 & 65.13 & 52.30 & 65.60 & 78.67 \\
OCR-BM25 (RRF) & Micro & 67.74 & 70.97 & 73.26 & 75.00 & 82.26 & 88.62 & 62.27 & 66.34 & 69.75 & 69.53 & 77.13 & 84.24 \\
ColNomic (Rerank) with & Macro & 61.38 & 65.84 & 69.10 & 65.41 & 74.60 & 82.96 & 46.78 & 54.95 & 61.86 & 51.07 & 62.90 & 74.39 \\
OCR-BM25 (RRF)& Micro & 67.30 & 70.39 & 72.61 & 74.29 & 81.25 & 87.37 & 62.16 & 66.07 & 69.34 & 69.16 & 76.47 & 83.30 \\
\bottomrule
\end{tabular}
}
\caption{\textbf{Average Retrieval Results} on our VL-DocIR benchmark. Each model family is reported with the best variants under structure-balanced Macro (primary) and query-averaged Micro (secondary). SV/MV denotes single-vector and multi-vector retrieval. Ensemble denotes a combined setting. Bold indicates the best score within each aggregation protocol. The more detailed results and analysis are reported in the Appendix.}
\label{tab:overall_metric_summary}
\end{table*}

\paragraph{Retrieval-stage setting effects.}
Table~\ref{tab:setting_effects} shows that retrieval-stage
modifications improve different aspects of the task. Reranking
generally improves nDCGAll over the base retrievers by refining
candidate ordering. Document-context scoring often improves
RecallAll by helping retrieve a more complete evidence set when
relevant pages belong to the same document.
For ColPali, Ctx improves Macro RecallAll@10 from 49.13 to 65.69
(from 60.26 to 69.90 under Micro), while Rerank gives the best
nDCGAll@10 under both protocols.
For ColQwen2.5, Ctx gives the best Macro RecallAll@10, whereas
B+Ctx gives the best Micro RecallAll@10; Rerank gives the best
nDCGAll@10 under both protocols.
For ColNomic, B+Ctx gives the best Macro nDCGAll@10 and
RecallAll@10 among the settings in this table. Under Micro,
Rerank gives the best nDCGAll@10, while B+Ctx gives the best
RecallAll@10. The three-list ColNomic ensemble further improves
both metrics under both protocols, as shown in
Table~\ref{tab:overall_metric_summary}.
These differences highlight distinct challenges in candidate
generation, document-level coverage, and candidate ordering.

\begin{table}[t]
\centering
\small
\setlength{\tabcolsep}{12pt}
\renewcommand{\arraystretch}{0.95}
\resizebox{\textwidth}{!}{
\begin{tabular}{@{}llccccc@{}}
\toprule
Family & Aggregation & Base & Q-Split & Ctx & B+Ctx & Rerank \\
\midrule

\multirow{2}{*}{ColPali}
& Macro
& 39.41/49.13
& --
& 50.87/\textbf{65.69}
& 48.69/62.55
& \textbf{51.12}/57.16 \\

& Micro
& 47.85/60.26
& --
& 53.48/\textbf{69.90}
& 53.14/69.12
& \textbf{61.33}/67.17 \\
\midrule

\multirow{2}{*}{ColQwen2.5}
& Macro
& 56.72/67.23
& 57.48/69.19
& 62.80/\textbf{75.78}
& 62.74/75.68
& \textbf{62.88}/71.62 \\

& Micro
& 65.51/77.04
& 65.64/78.63
& 66.85/80.10
& 67.81/\textbf{81.41}
& \textbf{71.94}/80.36 \\
\midrule

\multirow{2}{*}{ColNomic}
& Macro
& 61.82/71.76
& --
& 66.03/78.19
& \textbf{67.06}/\textbf{79.60}
& 65.00/74.62 \\

& Micro
& 70.68/81.24
& --
& 70.07/82.26
& 71.96/\textbf{84.81}
& \textbf{73.76}/83.11 \\

\bottomrule
\end{tabular}%
}
\caption{\textbf{Retrieval-stage setting effects} measured by nDCGAll@10 and RecallAll@10. Ctx denotes document-context scoring, B+Ctx combines the ranked
lists from base and document-context retrieval, Rerank denotes
reranking, and Q-Split denotes query splitting.
Bold indicates the best value for each metric.}
\label{tab:setting_effects}
\end{table}

\setlength{\columnsep}{0.65em}
\begin{wrapfigure}[16]{r}{0.52\linewidth}
\vspace{-2em}
\centering
\scriptsize
\begin{tikzpicture}[x=0.95cm,y=0.64cm]

\newcommand{\stresscell}[3]{%
  \pgfmathsetmacro{\val}{#3}
  \pgfmathsetmacro{\shade}{12 + 38*(\val-18)/(97.4-18)}%
  \fill[teal!\shade!white] (#1,#2) rectangle ++(1,1);
  \draw[white,line width=0.35pt] (#1,#2) rectangle ++(1,1);
  \node[font=\scriptsize] at (#1+0.5,#2+0.5) {#3};
}

\foreach \x/\name in {
  0/Page-Single,
  1/Page-Multi,
  2/Doc-Adjacent,
  3/Doc-Near,
  4/Doc-Far,
  5/Cross-Doc
} {
  \node[font=\scriptsize,rotate=35,anchor=west] at (\x+0.08,4.16) {\name};
}

\foreach \y/\name in {
  3/Wikipedia,
  2/arXiv,
  1/PubMed,
  0/SEC
} {
  \node[font=\scriptsize,anchor=east] at (-0.15,\y+0.50) {\name};
}

\stresscell{0}{3}{92.5}
\stresscell{1}{3}{93.0}
\stresscell{2}{3}{67.2}
\stresscell{3}{3}{64.4}
\stresscell{4}{3}{67.1}
\stresscell{5}{3}{50.9}

\stresscell{0}{2}{95.8}
\stresscell{1}{2}{97.1}
\stresscell{2}{2}{79.5}
\stresscell{3}{2}{77.3}
\stresscell{4}{2}{75.5}
\stresscell{5}{2}{18.0}

\stresscell{0}{1}{96.3}
\stresscell{1}{1}{97.4}
\stresscell{2}{1}{82.9}
\stresscell{3}{1}{79.6}
\stresscell{4}{1}{73.7}
\stresscell{5}{1}{28.0}

\stresscell{0}{0}{89.0}
\stresscell{1}{0}{90.2}
\stresscell{2}{0}{62.1}
\stresscell{3}{0}{59.7}
\stresscell{4}{0}{57.3}
\stresscell{5}{0}{26.6}

\draw[black!25] (0,0) rectangle (6,4);

\end{tikzpicture}
\caption{Source and evidence-structure-specific RecallAll@5 for the best evaluated configuration. Each cell shows the RecallAll@5 for a given source and evidence structure. Darker cells indicate higher RecallAll@5.}
\label{fig:evidence_complete_stress_map}
\vspace{-1.0em}
\end{wrapfigure}

\paragraph{Source and evidence-structure stress.} Figure~\ref{fig:evidence_complete_stress_map} shows that aggregate retrieval performance hides the main stress cases. Single-page queries are close to saturated for the strongest configuration across all sources, but distributed-evidence queries are much harder. Within-document multi-page retrieval remains challenging, especially for Wikipedia and SEC proxy statements, where long documents contain repeated structures and visually plausible distractors. In Cross-Doc retrieval, RecallAll@5 of ColNomic's best variant drops to 49.8 on Wikipedia, 26.6 on SEC, 14.1 on arXiv, and 20.0 on PubMed. Across all evaluated settings, the best Cross-Doc RecallAll@5 reaches 50.9 on Wikipedia, 26.6 on SEC, 18.0 on arXiv, and 28.0 on PubMed, showing that query splitting can be more effective than reranking for cross-document regimes.

\setlength{\columnsep}{0.65em}
\begin{wraptable}[12]{r}{0.55\linewidth}
\vspace{-0.8em}
\centering
\small
\setlength{\tabcolsep}{5pt}
\renewcommand{\arraystretch}{0.96}
\begin{tabular}{@{}lrr@{}}
\toprule
Evidence structure & \#Queries & Macro RAll@5 \\
\midrule
Page-Single      & 86{,}196  & 93.4 \\
Page-Multi       & 120{,}624 & 94.4 \\
Doc-Adjacent     & 11{,}972  & 71.4 \\
Doc-Near         & 3{,}392   & 69.0 \\
Doc-Far          & 2{,}563   & 66.3 \\
Cross-Doc   & 47{,}013  & 27.6 \\
\bottomrule
\end{tabular}
\caption{Evidence-structure aggregate view for ColNomic B+Ctx+Rank. RAll@5 denotes RecallAll@5 macro-averaged over the four data sources.}
\label{tab:evidence_structure_aggregate}
\vspace{-1.0em}
\end{wraptable}

\paragraph{Evidence-structure aggregate.}
Table~\ref{tab:evidence_structure_aggregate} summarizes the same trend at the evidence structure level. The benchmark contains a large base of single-page queries, but the source-macro RecallAll@5 drops sharply once evidence must be recovered from multiple pages. The hardest transition is not from Page-Single to Page-Multi, but from within-document evidence to cross-document evidence. This supports the central motivation of VL-DocIR, \ie, long-document retrieval cannot be reduced to single-page ranking over a larger corpus.

\paragraph{Scope of the experimental claims.}
The experiments support retrieval-stage claims about current visual retrievers on VL-DocIR, \ie, multi-vector retrieval is stronger than single-vector retrieval, document-context scoring helps mainly when evidence remains within one document, and cross-document evidence coverage remains weak. They do not show that vision-only retrieval is superior to OCR-only, text-only, or hybrid retrieval, and they do not evaluate end-to-end question answering. A high RecallAll@10 means that all required pages are retrieved, which does not guarantee that a downstream task will answer correctly.

\subsection{Qualitative Results and Analysis}
\begin{figure}[htbp]
  \centering
  \includegraphics[
        trim=4cm 3cm 3cm 4cm,
        clip,
      width=1\textwidth
  ]{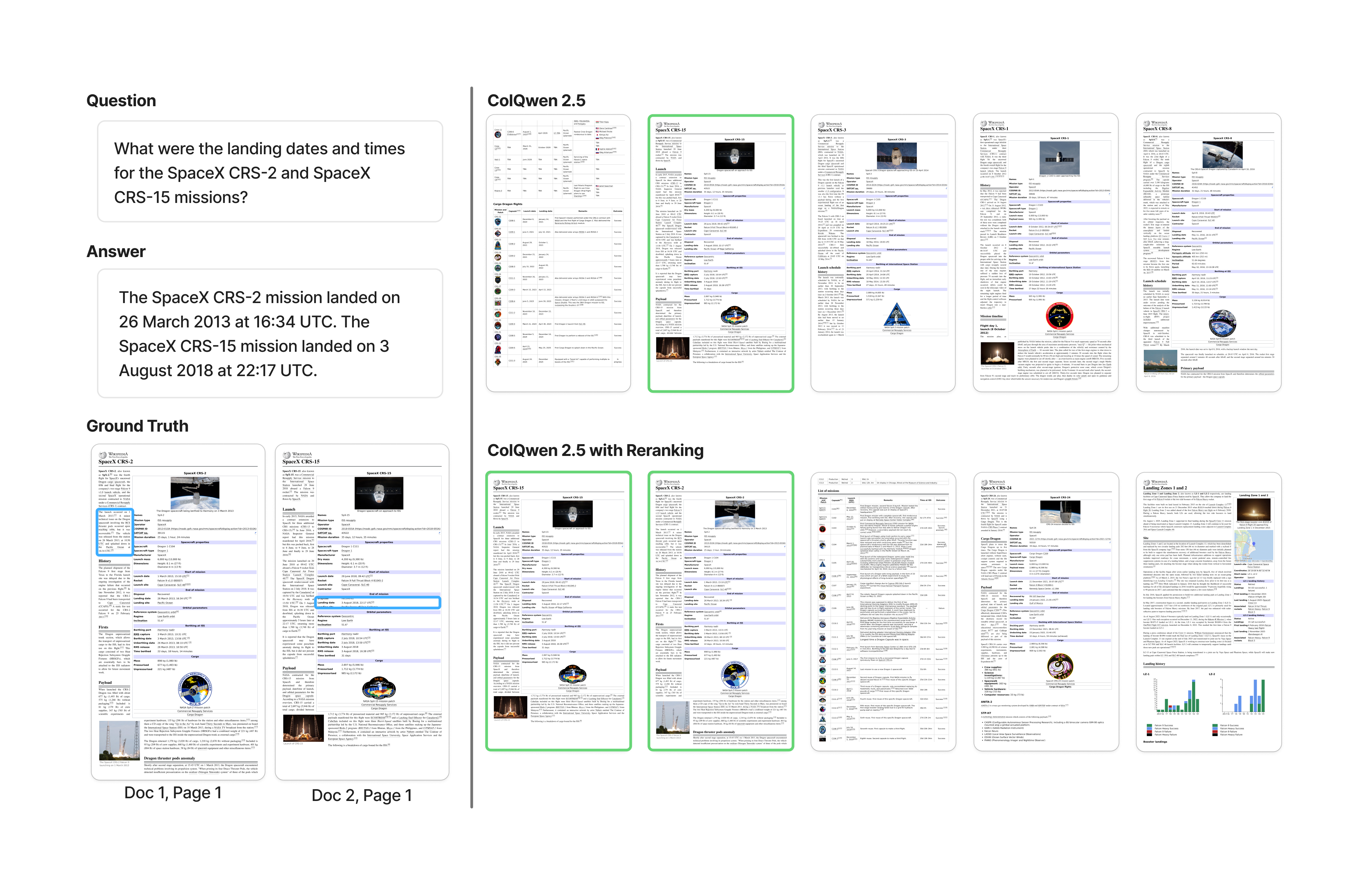}
  \caption{Cross-Doc query top-5 retrieval results of ColQwen2.5 and ColQwen2.5 with reranking. The blue rectangles indicate the evidence bounding-boxes. The green rectangles highlight correctly retrieved pages.}
  \label{fig:cross_document_question_retrieval_rerank_example}
\end{figure}

Fig.~\ref{fig:cross_document_question_retrieval_rerank_example} presents the top-5 document pages retrieved by ColQwen2.5, both with and without reranking, for a Cross-Doc query. This example demonstrates that reranking can substantially improve retrieval performance compared with the base model. Specifically, the base ColQwen2.5 model retrieves only one of the two relevant evidence pages, whereas the reranked results correctly identify both evidence pages. Although the pages retrieved by the base model are thematically plausible, as they also contain information about space missions, they refer to incorrect mission numbers. This indicates that the base retriever can capture high-level semantic relevance but may fail to distinguish fine-grained details. In contrast, the reranking stage better resolves such subtle distinctions, leading to more accurate retrieval of the required document evidence.

\section{Conclusion}
\label{sec:conclusion}

VL-DocIR introduces an evidence-complete page-retrieval benchmark for visually rich long documents, with single-page, multi-page, and cross-document evidence configurations. Unlike standard page-level retrieval evaluations, VL-DocIR requires systems to recover the complete set of evidence pages for each question. Our experiments show that multi-vector page-image retrieval is substantially stronger than single-vector retrieval, document-context scoring is most useful when evidence remains within one document, and cross-document evidence retrieval remains the most challenging regime. The strongest evaluated configuration reaches 75.6 nDCGAll@10 and 86.2 RecallAll@10, while source- and evidence-structure breakdowns show that aggregate scores still mask large cross-document failures.

\paragraph{Limitations.}
VL-DocIR evaluates evidence-page retrieval with page-level rather than exhaustive element level grounding. The current corpus focuses on rendered HTML-born documents, leaving scanned and handwritten documents outside scope. The query set is VLM-generated and automatically cleaned, so generator style and filtering criteria may still influence the benchmark distribution. 

\begin{ack}
This work was supported in part by Helmholtz Association of German Research Centers, in part by the Ministry of Science, Research and the Arts of Baden-Württemberg (MWK) through the Cooperative Graduate School Accessibility through AI-based Assistive Technology (KATE) under Grant BW6-03, and in part by the National Natural Science Foundation of China under Grant No. 62503166, in part by the Hunan Provincial Research and Development Project under Grant number 2026QK3018, and in part by the Yuelushan Industrial Innovation Center. This work was partially performed on the HoreKa supercomputer funded by the MWK and by the Federal Ministry of Education and Research, partially on the HAICORE@KIT partition supported by the Helmholtz Association Initiative and Networking Fund, and partially on bwForCluster Helix supported by the state of Baden-Württemberg through bwHPC and the German Research Foundation (DFG) through grant INST 35/1597-1 FUGG. 
\end{ack}


{
\small
\bibliographystyle{unsrt}
\bibliography{main}

@STRING{aaai	= {AAAI} }

@STRING{eccv	= {ECCV} }

@STRING{wacv	= {WACV} }

@STRING{iclr	= {ICLR} }

@inproceedings{
faysse2024colpali,
title={ColPali: Efficient Document Retrieval with Vision Language Models},
author={Manuel Faysse and Hugues Sibille and Tony Wu and Bilel Omrani and Gautier Viaud and CELINE HUDELOT and Pierre Colombo},
booktitle={The Thirteenth International Conference on Learning Representations},
year={2025},
url={https://openreview.net/forum?id=ogjBpZ8uSi}
}

@misc{nomicembedmultimodal2025,
  title={Nomic Embed Multimodal: Interleaved Text, Image, and Screenshots for Visual Document Retrieval},
  author={Nomic Team},
  year={2025},
  publisher={Nomic AI},
  url={https://nomic.ai/blog/posts/nomic-embed-multimodal},
  howpublished={\url{https://nomic.ai/blog/posts/nomic-embed-multimodal} (last accessed: 21. April 2026)},
}

@inproceedings{guenther2025jinaembeddingsv4universalembeddingsmultimodal,
  title={jina-embeddings-v4: Universal Embeddings for Multimodal Multilingual Retrieval},
  author={G{\"u}nther, Michael and Sturua, Saba and Akram, Mohammad Kalim and Mohr, Isabelle and Ungureanu, Andrei and Wang, Bo and Eslami, Sedigheh and Martens, Scott and Werk, Maximilian and Wang, Nan and Xiao, Han},
  booktitle={Proceedings of the 5th Workshop on Multilingual Representation Learning (MRL 2025)},
  pages={531--550},
  year={2025},
  address={Suzhuo, China},
  publisher={Association for Computational Linguistics},
  url={https://aclanthology.org/2025.mrl-main.36/},
  doi={10.18653/v1/2025.mrl-main.36}
}

@article{loison2026vidore,
  title={ViDoRe V3: A Comprehensive Evaluation of Retrieval Augmented Generation in Complex Real-World Scenarios},
  author={Loison, Ant{\'o}nio and Mac{\'e}, Quentin and Edy, Antoine and Xing, Victor and Balough, Tom and Moreira, Gabriel and Liu, Bo and Faysse, Manuel and Hudelot, C{\'e}line and Viaud, Gautier},
  journal={arXiv preprint arXiv:2601.08620},
  year={2026},
  eprint={2601.08620},
  archivePrefix={arXiv},
  primaryClass={cs.IR},
  url={https://arxiv.org/abs/2601.08620}
}

@article{teiletche2025modernvbert,
  title={ModernVBERT: Towards Smaller Visual Document Retrievers},
  author={Teiletche, Paul and Mac{\'e}, Quentin and Conti, Max and Loison, Antonio and Viaud, Gautier and Colombo, Pierre and Faysse, Manuel},
  journal={arXiv preprint arXiv:2510.01149},
  year={2025},
  eprint={2510.01149},
  archivePrefix={arXiv},
  primaryClass={cs.IR},
  url={https://arxiv.org/abs/2510.01149}
}

@article{zhang2024gme,
  title={GME: Improving Universal Multimodal Retrieval by Multimodal LLMs},
  author={Zhang, Xin and Zhang, Yanzhao and Xie, Wen and Li, Mingxin and Dai, Ziqi and Long, Dingkun and Xie, Pengjun and Zhang, Meishan and Li, Wenjie and Zhang, Min},
  journal={arXiv preprint arXiv:2412.16855},
  year={2024},
  eprint={2412.16855},
  archivePrefix={arXiv},
  primaryClass={cs.CL},
  url={https://arxiv.org/abs/2412.16855}
}

@misc{huang2025tomoroqwen3,
  author = {Huang, Xin and Tan, Kye Min},
  title = {Beyond Text: Unlocking True Multimodal, End-to-end RAG with Tomoro ColQwen3},
  year = {2025},
  url = {https://tomoro.ai/insights/beyond-text-unlocking-true-multimodal-end-to-end-rag-with-tomoro-colqwen3},
  howpublished = {\url{https://tomoro.ai/insights/beyond-text-unlocking-true-multimodal-end-to-end-rag-with-tomoro-colqwen3} (last accessed: 21. April 2026)},
  publisher = {Tomoro.ai}
}

@article{jarvelin2002,
author = {J\"{a}rvelin, Kalervo and Kek\"{a}l\"{a}inen, Jaana},
title = {Cumulated gain-based evaluation of IR techniques},
year = {2002},
issue_date = {October 2002},
publisher = {Association for Computing Machinery},
address = {New York, NY, USA},
volume = {20},
number = {4},
issn = {1046-8188},
journal = {ACM Trans. Inf. Syst.},
month = oct,
pages = {422--446},
numpages = {25},
url = {https://doi.org/10.1145/582415.582418},
doi = {10.1145/582415.582418}
}

@misc{jina2025jinarerankerm0,
  title = {{jina-reranker-m0}},
  author = {JinaAI},
  year = {2025},
  url = {https://huggingface.co/jinaai/jina-reranker-m0},
  howpublished = {\url{https://huggingface.co/jinaai/jina-reranker-m0} (last accessed: 21. April 2026)},
  note = {Hugging Face Model Card}
}

@inproceedings{
    xiao2026metaembed,
    title={MetaEmbed: Scaling Multimodal Retrieval at Test-Time with Flexible Late Interaction},
    author={Zilin Xiao and Qi Ma and Mengting Gu and Chun-cheng Jason Chen and Xintao Chen and Vicente Ordonez and Vijai Mohan},
    booktitle={The Fourteenth International Conference on Learning Representations},
    year={2026},
    url={https://openreview.net/forum?id=yKDqg9HwZX}
}

@inproceedings{uzan2026gqr,
  title     = {Guided Query Refinement: Multimodal Hybrid Retrieval with Test-Time Optimization},
  author    = {Uzan, Omri and Yehudai, Asaf and Pony, Roi and Shnarch, Eyal and Gera, Ariel},
  booktitle = {International Conference on Learning Representations (ICLR)},
  year      = {2026},
  url       = {https://arxiv.org/abs/2510.05038}
}

@inproceedings{mathew2021docvqa,
  title={DocVQA: A Dataset for VQA on Document Images},
  author={Mathew, Minesh and Karatzas, Dimosthenis and Jawahar, C. V.},
  booktitle={Proceedings of the IEEE/CVF Winter Conference on Applications of Computer Vision},
  pages={2200--2209},
  year={2021},
  doi={10.1109/WACV48630.2021.00225},
  url={https://openaccess.thecvf.com/content/WACV2021/papers/Mathew_DocVQA_A_Dataset_for_VQA_on_Document_Images_WACV_2021_paper.pdf}
}

@inproceedings{mathew2021infographicvqa,
  title={InfographicVQA},
  author={Mathew, Minesh and Bagal, Viraj and Tito, Rub{\'e}n P{\'e}rez and Karatzas, Dimosthenis and Valveny, Ernest and Jawahar, C. V.},
  booktitle={Proceedings of the IEEE/CVF Winter Conference on Applications of Computer Vision},
  pages={1697--1706},
  year={2022},
  url={https://openaccess.thecvf.com/content/WACV2022/html/Mathew_InfographicVQA_WACV_2022_paper.html}
}

@inproceedings{vanlandeghem2023dude,
    title={ICDAR 2023 Competition on Document UnderstanDing of Everything (DUDE)},
    author={Van Landeghem, Jordy and Borchmann, {\L}ukasz and Tito, Rub{\'e}n and Pietruszka, Micha{\l} and Jurkiewicz, Dawid and Powalski, Rafa{\l} and J{\'o}ziak, Pawe{\l} and Biswas, Sanket and Coustaty, Micka{\"e}l and Stanis{\l}awek, Tomasz},
    booktitle={Document Analysis and Recognition -- ICDAR 2023},
    pages={420--434},
    year={2023},
    publisher={Springer Nature Switzerland},
    doi={10.1007/978-3-031-41679-8_24},
    url={https://link.springer.com/chapter/10.1007/978-3-031-41679-8_24}
}

@article{tito2023mpdocvqa,
author = {Tito, Rub\`{e}n and Karatzas, Dimosthenis and Valveny, Ernest},
title = {Hierarchical multimodal transformers for Multipage DocVQA},
year = {2023},
issue_date = {Dec 2023},
publisher = {Elsevier Science Inc.},
address = {USA},
volume = {144},
number = {C},
issn = {0031-3203},
url = {https://doi.org/10.1016/j.patcog.2023.109834},
doi = {10.1016/j.patcog.2023.109834},
journal = {Pattern Recogn.},
month = dec,
numpages = {11}
}

@inproceedings{
ma2024mmlongbenchdoc,
title={{MMLONGBENCH}-{DOC}: Benchmarking Long-context Document Understanding with Visualizations},
author={Yubo Ma and Yuhang Zang and Liangyu Chen and Meiqi Chen and Yizhu Jiao and Xinze Li and Xinyuan Lu and Ziyu Liu and Yan Ma and Xiaoyi Dong and Pan Zhang and Liangming Pan and Yu-Gang Jiang and Jiaqi Wang and Yixin Cao and Aixin Sun},
booktitle={The Thirty-eight Conference on Neural Information Processing Systems Datasets and Benchmarks Track},
year={2024},
url={https://openreview.net/forum?id=loJM1acwzf}
}

@inproceedings{deng2025longdocurl,
    title = "{L}ong{D}oc{URL}: a Comprehensive Multimodal Long Document Benchmark Integrating Understanding, Reasoning, and Locating",
    author = "Deng, Chao  and
      Yuan, Jiale  and
      Bu, Pi  and
      Wang, Peijie  and
      Li, Zhong-Zhi  and
      Xu, Jian  and
      Li, Xiao-Hui  and
      Gao, Yuan  and
      Song, Jun  and
      Zheng, Bo  and
      Liu, Cheng-Lin",
    editor = "Che, Wanxiang  and
      Nabende, Joyce  and
      Shutova, Ekaterina  and
      Pilehvar, Mohammad Taher",
    booktitle = "Proceedings of the 63rd Annual Meeting of the Association for Computational Linguistics (Volume 1: Long Papers)",
    month = jul,
    year = "2025",
    address = "Vienna, Austria",
    publisher = "Association for Computational Linguistics",
    url = "https://aclanthology.org/2025.acl-long.57/",
    doi = "10.18653/v1/2025.acl-long.57",
    pages = "1135--1159",
    ISBN = "979-8-89176-251-0",
}

@inproceedings{masry2022chartqa,
    title = "{C}hart{QA}: A Benchmark for Question Answering about Charts with Visual and Logical Reasoning",
    author = "Masry, Ahmed  and
      Long, Do Xuan  and
      Tan, Jia Qing  and
      Joty, Shafiq  and
      Hoque, Enamul",
    editor = "Muresan, Smaranda  and
      Nakov, Preslav  and
      Villavicencio, Aline",
    booktitle = "Findings of the Association for Computational Linguistics: ACL 2022",
    month = may,
    year = "2022",
    address = "Dublin, Ireland",
    publisher = "Association for Computational Linguistics",
    url = "https://aclanthology.org/2022.findings-acl.177/",
    doi = "10.18653/v1/2022.findings-acl.177",
    pages = "2263--2279",
}

@InProceedings{methani2020plotqa,
author = {Methani, Nitesh and Ganguly, Pritha and Khapra, Mitesh M. and Kumar, Pratyush},
title = {PlotQA: Reasoning over Scientific Plots},
booktitle = {The IEEE Winter Conference on Applications of Computer Vision (WACV)},
month = {March},
year = {2020},
pages = {1527--1536},
doi = {10.1109/WACV45572.2020.9093523},
url = {https://openaccess.thecvf.com/content_WACV_2020/html/Methani_PlotQA_Reasoning_over_Scientific_Plots_WACV_2020_paper.html}
}

@inproceedings{
    thakur2021beir,
    title={{BEIR}: A Heterogeneous Benchmark for Zero-shot Evaluation of Information Retrieval Models},
    author={Nandan Thakur and Nils Reimers and Andreas R{\"u}ckl{\'e} and Abhishek Srivastava and Iryna Gurevych},
    booktitle={Thirty-fifth Conference on Neural Information Processing Systems Datasets and Benchmarks Track (Round 2)},
    year={2021},
    url={https://openreview.net/forum?id=wCu6T5xFjeJ}
}

@inproceedings{muennighoff2022mteb,
    title = "{MTEB}: Massive Text Embedding Benchmark",
    author = "Muennighoff, Niklas  and
      Tazi, Nouamane  and
      Magne, Loic  and
      Reimers, Nils",
    editor = "Vlachos, Andreas  and
      Augenstein, Isabelle",
    booktitle = "Proceedings of the 17th Conference of the European Chapter of the Association for Computational Linguistics",
    month = may,
    year = "2023",
    address = "Dubrovnik, Croatia",
    publisher = "Association for Computational Linguistics",
    url = "https://aclanthology.org/2023.eacl-main.148/",
    doi = "10.18653/v1/2023.eacl-main.148",
    pages = "2014--2037"
}

@inproceedings{
jiang2025vlm2vec,
title={{VLM}2Vec: Training Vision-Language Models for Massive Multimodal Embedding Tasks},
author={Ziyan Jiang and Rui Meng and Xinyi Yang and Semih Yavuz and Yingbo Zhou and Wenhu Chen},
booktitle={The Thirteenth International Conference on Learning Representations},
year={2025},
url={https://openreview.net/forum?id=TE0KOzWYAF}
}

@inproceedings{yang2018hotpotqa,
    title = "{H}otpot{QA}: A Dataset for Diverse, Explainable Multi-hop Question Answering",
    author = "Yang, Zhilin  and
      Qi, Peng  and
      Zhang, Saizheng  and
      Bengio, Yoshua  and
      Cohen, William  and
      Salakhutdinov, Ruslan  and
      Manning, Christopher D.",
    editor = "Riloff, Ellen  and
      Chiang, David  and
      Hockenmaier, Julia  and
      Tsujii, Jun{'}ichi",
    booktitle = "Proceedings of the 2018 Conference on Empirical Methods in Natural Language Processing",
    month = oct # "-" # nov,
    year = "2018",
    address = "Brussels, Belgium",
    publisher = "Association for Computational Linguistics",
    url = "https://aclanthology.org/D18-1259/",
    doi = "10.18653/v1/D18-1259",
    pages = "2369--2380",
}

@inproceedings{bai2023longbench,
    title = "{L}ong{B}ench: A Bilingual, Multitask Benchmark for Long Context Understanding",
    author = "Bai, Yushi  and
      Lv, Xin  and
      Zhang, Jiajie  and
      Lyu, Hongchang  and
      Tang, Jiankai  and
      Huang, Zhidian  and
      Du, Zhengxiao  and
      Liu, Xiao  and
      Zeng, Aohan  and
      Hou, Lei  and
      Dong, Yuxiao  and
      Tang, Jie  and
      Li, Juanzi",
    editor = "Ku, Lun-Wei  and
      Martins, Andre  and
      Srikumar, Vivek",
    booktitle = "Proceedings of the 62nd Annual Meeting of the Association for Computational Linguistics (Volume 1: Long Papers)",
    month = aug,
    year = "2024",
    address = "Bangkok, Thailand",
    publisher = "Association for Computational Linguistics",
    url = "https://aclanthology.org/2024.acl-long.172/",
    doi = "10.18653/v1/2024.acl-long.172",
    pages = "3119--3137",
}

@article{mace2025vidorev2,
  author = {Mac{\'e}, Quentin and Loison, Ant{\'o}nio and Faysse, Manuel},
  journal = {arXiv preprint arXiv:2505.17166},
  title = {ViDoRe Benchmark V2: Raising the Bar for Visual Retrieval},
  year = {2025},
  eprint = {2505.17166},
  archivePrefix = {arXiv},
  primaryClass = {cs.IR},
  url = {https://arxiv.org/abs/2505.17166},
}

@inproceedings{ma2024dse,
    title = "Unifying Multimodal Retrieval via Document Screenshot Embedding",
    author = "Ma, Xueguang  and
      Lin, Sheng-Chieh  and
      Li, Minghan  and
      Chen, Wenhu  and
      Lin, Jimmy",
    editor = "Al-Onaizan, Yaser  and
      Bansal, Mohit  and
      Chen, Yun-Nung",
    booktitle = "Proceedings of the 2024 Conference on Empirical Methods in Natural Language Processing",
    month = nov,
    year = "2024",
    address = "Miami, Florida, USA",
    publisher = "Association for Computational Linguistics",
    url = "https://aclanthology.org/2024.emnlp-main.373/",
    doi = "10.18653/v1/2024.emnlp-main.373",
    pages = "6492--6505",
}

@inproceedings{
yu2025visrag,
title={Vis{RAG}: Vision-based Retrieval-augmented Generation on Multi-modality Documents},
author={Shi Yu and Chaoyue Tang and Bokai Xu and Junbo Cui and Junhao Ran and Yukun Yan and Zhenghao Liu and Shuo Wang and Xu Han and Zhiyuan Liu and Maosong Sun},
booktitle={The Thirteenth International Conference on Learning Representations},
year={2025},
url={https://openreview.net/forum?id=zG459X3Xge}
}

@article{bai2025qwen3,
  title={Qwen3-VL Technical Report},
  author={Bai, Shuai and Cai, Yuxuan and Chen, Ruizhe and Chen, Keqin and Chen, Xionghui and Cheng, Zesen and Deng, Lianghao and Ding, Wei and Gao, Chang and Ge, Chunjiang and others},
  journal={arXiv preprint arXiv:2511.21631},
  year={2025},
  eprint={2511.21631},
  archivePrefix={arXiv},
  primaryClass={cs.CV},
  url={https://arxiv.org/abs/2511.21631}
}

@article{tfidf,
author = {Salton, G. and Wong, A. and Yang, C. S.},
title = {A vector space model for automatic indexing},
year = {1975},
issue_date = {Nov. 1975},
publisher = {Association for Computing Machinery},
address = {New York, NY, USA},
volume = {18},
number = {11},
issn = {0001-0782},
url = {https://doi.org/10.1145/361219.361220},
doi = {10.1145/361219.361220},
journal = {Commun. ACM},
month = nov,
pages = {613--620},
numpages = {8}
}

@misc{openai2025gpt5nano,
  title = {{GPT-5 nano}},
  author = {OpenAI},
  year = {2025},
  url = {https://developers.openai.com/api/docs/models/gpt-5-nano},
  howpublished = {\url{https://developers.openai.com/api/docs/models/gpt-5-nano} (last accessed: 6 May 2026)},
}

@misc{openai2024gpt41mini,
  title = {{GPT-4.1 mini}},
  author = {OpenAI},
  year = {2025},
  url = {https://developers.openai.com/api/docs/models/gpt-4.1-mini},
  howpublished = {\url{https://developers.openai.com/api/docs/models/gpt-4.1-mini} (last accessed: 6 May 2026)},
}

@misc{chen2025visrbench,
      title={VisR-Bench: An Empirical Study on Visual Retrieval-Augmented Generation for Multilingual Long Document Understanding}, 
      author={Jian Chen and Ming Li and Jihyung Kil and Chenguang Wang and Tong Yu and Ryan Rossi and Tianyi Zhou and Changyou Chen and Ruiyi Zhang},
      year={2025},
      eprint={2508.07493},
      archivePrefix={arXiv},
      primaryClass={cs.CV},
      url={https://arxiv.org/abs/2508.07493}, 
}

@misc{dong2025mmdocir,
      title={MMDocIR: Benchmarking Multi-Modal Retrieval for Long Documents}, 
      author={Kuicai Dong and Yujing Chang and Xin Deik Goh and Dexun Li and Ruiming Tang and Yong Liu},
      year={2025},
      eprint={2501.08828},
      archivePrefix={arXiv},
      primaryClass={cs.IR},
      url={https://arxiv.org/abs/2501.08828}, 
}

@inproceedings{cho2025m3docvqa,
  title={M3DocVQA: Multi-modal Multi-page Multi-document Understanding},
  author={Cho, Jaemin and Mahata, Debanjan and Irsoy, Ozan and He, Yujie and Bansal, Mohit},
  booktitle={Proceedings of the IEEE/CVF International Conference on Computer Vision Workshops},
  pages={6237--6247},
  year={2025},
  url={https://openaccess.thecvf.com/content/ICCV2025W/Findings/html/Cho_M3DocVQA_Multi-modal_Multi-page_Multi-document_Understanding_ICCVW_2025_paper.html}
}

@inproceedings{suri2025visdom,
  title={VisDoM: Multi-Document QA with Visually Rich Elements Using Multimodal Retrieval-Augmented Generation},
  author={Suri, Manan and Mathur, Puneet and Dernoncourt, Franck and Goswami, Kanika and Rossi, Ryan A. and Manocha, Dinesh},
  booktitle={Proceedings of the 2025 Conference of the Nations of the Americas Chapter of the Association for Computational Linguistics: Human Language Technologies (Volume 1: Long Papers)},
  month=apr,
  year={2025},
  address={Albuquerque, New Mexico},
  publisher={Association for Computational Linguistics},
  url={https://aclanthology.org/2025.naacl-long.310/},
  doi={10.18653/v1/2025.naacl-long.310},
  pages={6088--6109}
}

@misc{wang2025vidorag,
  title={ViDoRAG: Visual Document Retrieval-Augmented Generation via Dynamic Iterative Reasoning Agents},
  author={Wang, Qiuchen and Ding, Ruixue and Chen, Zehui and Wu, Weiqi and Wang, Shihang and Xie, Pengjun and Zhao, Feng},
  year={2025},
  eprint={2502.18017},
  archivePrefix={arXiv},
  primaryClass={cs.CV},
  url={https://arxiv.org/abs/2502.18017}
}

@misc{yan2026colparse,
  title={Beyond the Grid: Layout-Informed Multi-Vector Retrieval with Parsed Visual Document Representations},
  author={Yan, Yibo and Ou, Mingdong and Cao, Yi and Zou, Xin and Liu, Shuliang and Huo, Jiahao and Huang, Yu and Kwok, James and Hu, Xuming},
  year={2026},
  eprint={2603.01666},
  archivePrefix={arXiv},
  primaryClass={cs.CL},
  url={https://arxiv.org/abs/2603.01666}
}

@misc{yan2026colchunk,
  title={Visual Late Chunking: An Empirical Study of Contextual Chunking for Efficient Visual Document Retrieval},
  author={Yan, Yibo and Ou, Mingdong and Cao, Yi and Huo, Jiahao and Zou, Xin and Liu, Shuliang and Kwok, James and Hu, Xuming},
  year={2026},
  eprint={2604.10167},
  archivePrefix={arXiv},
  primaryClass={cs.CV},
  url={https://arxiv.org/abs/2604.10167}
}

@misc{abdallah2026mmbright,
  title={MM-BRIGHT: A Multi-Task Multimodal Benchmark for Reasoning-Intensive Retrieval},
  author={Abdallah, Abdelrahman and Mounis, Mohamed Darwish and Abdalla, Mahmoud and Kasem, Mahmoud SalahEldin and Senussi, Mostafa Farouk and Mahmoud, Mohamed and Ali, Mohammed and Jatowt, Adam and Kang, Hyun-Soo},
  year={2026},
  eprint={2601.09562},
  archivePrefix={arXiv},
  primaryClass={cs.IR},
  url={https://arxiv.org/abs/2601.09562}
}

@misc{kasem2026marvel,
  title={MARVEL: Multimodal Adaptive Reasoning-intensiVe Expand-rerank and retrievaL},
  author={Kasem, Mahmoud SalahEldin and Mahmoud, Mohamed and Senussi, Mostafa Farouk and Abdalla, Mahmoud and Abdallah, Abdelrahman and Kang, Hyun-Soo},
  year={2026},
  eprint={2604.07079},
  archivePrefix={arXiv},
  primaryClass={cs.IR},
  url={https://arxiv.org/abs/2604.07079}
}

@inproceedings{singh2019textvqa,
  title={Towards VQA Models That Can Read},
  author={Singh, Amanpreet and Natarajan, Vivek and Shah, Meet and Jiang, Yu and Chen, Xinlei and Batra, Dhruv and Parikh, Devi and Rohrbach, Marcus},
  booktitle={Proceedings of the IEEE/CVF Conference on Computer Vision and Pattern Recognition},
  pages={8317--8326},
  year={2019},
  url={https://openaccess.thecvf.com/content_CVPR_2019/html/Singh_Towards_VQA_Models_That_Can_Read_CVPR_2019_paper.html}
}

@inproceedings{biten2019scene,
  title={Scene Text Visual Question Answering},
  author={Biten, Ali Furkan and Tito, Ruben and Mafla, Andres and Gomez, Lluis and Rusinol, Marcal and Mathew, Minesh and Jawahar, C. V. and Valveny, Ernest and Karatzas, Dimosthenis},
  booktitle={Proceedings of the IEEE/CVF International Conference on Computer Vision},
  pages={4291--4301},
  year={2019},
  doi={10.1109/ICCV.2019.00439},
  url={https://openaccess.thecvf.com/content_ICCV_2019/html/Biten_Scene_Text_Visual_Question_Answering_ICCV_2019_paper.html}
}

@inproceedings{mishra2019ocrvqa,
  title={OCR-VQA: Visual Question Answering by Reading Text in Images},
  author={Mishra, Anand and Shekhar, Shashank and Singh, Ajeet Kumar and Chakraborty, Anirban},
  booktitle={Proceedings of the International Conference on Document Analysis and Recognition},
  pages={947--952},
  year={2019},
  doi={10.1109/ICDAR.2019.00156},
  url={https://doi.org/10.1109/ICDAR.2019.00156}
}

@inproceedings{zhu2021tatqa,
  title={TAT-QA: A Question Answering Benchmark on a Hybrid of Tabular and Textual Content in Finance},
  author={Zhu, Fengbin and Lei, Wenqiang and Huang, Youcheng and Wang, Chao and Zhang, Shuo and Lv, Jiancheng and Feng, Fuli and Chua, Tat-Seng},
  booktitle={Proceedings of the 59th Annual Meeting of the Association for Computational Linguistics},
  pages={3277--3287},
  year={2021},
  doi={10.18653/v1/2021.acl-long.254},
  url={https://aclanthology.org/2021.acl-long.254/}
}

@inproceedings{talmor2021multimodalqa,
  title={MultiModalQA: Complex Question Answering over Text, Tables and Images},
  author={Talmor, Alon and Yoran, Ori and Catav, Amnon and Lahav, Dan and Wang, Yizhong and Asai, Akari and Ilharco, Gabriel and Hajishirzi, Hannaneh and Berant, Jonathan},
  booktitle={International Conference on Learning Representations},
  year={2021},
  url={https://openreview.net/forum?id=ee6W5UgQLa}
}

@inproceedings{chen2021websrc,
  title={WebSRC: A Dataset for Web-Based Structural Reading Comprehension},
  author={Chen, Xingyu and Zhao, Zihan and Chen, Lu and Ji, JiaBao and Zhang, Danyang and Luo, Ao and Xiong, Yuxuan and Yu, Kai},
  booktitle={Proceedings of the 2021 Conference on Empirical Methods in Natural Language Processing},
  pages={4173--4185},
  year={2021},
  doi={10.18653/v1/2021.emnlp-main.343},
  url={https://aclanthology.org/2021.emnlp-main.343/}
}

@inproceedings{tanaka2023slidevqa,
  title={SlideVQA: A Dataset for Document Visual Question Answering on Multiple Images},
  author={Tanaka, Ryota and Nishida, Kyosuke and Nishida, Kosuke and Hasegawa, Taku and Saito, Itsumi and Saito, Kuniko},
  booktitle={Proceedings of the AAAI Conference on Artificial Intelligence},
  volume={37},
  pages={13636--13645},
  year={2023},
  doi={10.1609/aaai.v37i11.26598},
  url={https://ojs.aaai.org/index.php/AAAI/article/view/26598}
}

@misc{wang2025mmlongbench,
  title={MMLongBench: Benchmarking Long-Context Vision-Language Models Effectively and Thoroughly},
  author={Wang, Zhaowei and Yu, Wenhao and Ren, Xiyu and Zhang, Jipeng and Zhao, Yu and Saxena, Rohit and Cheng, Liang and Wong, Ginny and See, Simon and Minervini, Pasquale and Song, Yangqiu and Steedman, Mark},
  year={2025},
  eprint={2505.10610},
  archivePrefix={arXiv},
  primaryClass={cs.CV},
  url={https://arxiv.org/abs/2505.10610}
}

@inproceedings{xu2020layoutlm,
  title={LayoutLM: Pre-training of Text and Layout for Document Image Understanding},
  author={Xu, Yiheng and Li, Minghao and Cui, Lei and Huang, Shaohan and Wei, Furu and Zhou, Ming},
  booktitle={Proceedings of the 26th ACM SIGKDD International Conference on Knowledge Discovery and Data Mining},
  pages={1192--1200},
  year={2020},
  doi={10.1145/3394486.3403172},
  url={https://doi.org/10.1145/3394486.3403172}
}

@inproceedings{xu2021layoutlmv2,
  title={LayoutLMv2: Multi-modal Pre-training for Visually-Rich Document Understanding},
  author={Xu, Yang and Xu, Yiheng and Lv, Tengchao and Cui, Lei and Wei, Furu and Wang, Guoxin and Lu, Yijuan and Florencio, Dinei and Zhang, Cha and Che, Wanxiang and Zhang, Min and Zhou, Lidong},
  booktitle={Proceedings of the 59th Annual Meeting of the Association for Computational Linguistics and the 11th International Joint Conference on Natural Language Processing (Volume 1: Long Papers)},
  pages={2579--2591},
  year={2021},
  address={Online},
  publisher={Association for Computational Linguistics},
  url={https://aclanthology.org/2021.acl-long.201/},
  doi={10.18653/v1/2021.acl-long.201}
}

@inproceedings{huang2022layoutlmv3,
  title={LayoutLMv3: Pre-training for Document AI with Unified Text and Image Masking},
  author={Huang, Yupan and Lv, Tengchao and Cui, Lei and Lu, Yutong and Wei, Furu},
  booktitle={Proceedings of the 30th ACM International Conference on Multimedia},
  pages={4083--4091},
  year={2022},
  doi={10.1145/3503161.3548112},
  url={https://doi.org/10.1145/3503161.3548112}
}

@inproceedings{appalaraju2021docformer,
  title={DocFormer: End-to-End Transformer for Document Understanding},
  author={Appalaraju, Srikar and Jasani, Bhavan and Kota, Bhargava Urala and Xie, Yusheng and Manmatha, R.},
  booktitle={Proceedings of the IEEE/CVF International Conference on Computer Vision},
  pages={993--1003},
  year={2021},
  doi={10.1109/ICCV48922.2021.00103},
  url={https://openaccess.thecvf.com/content/ICCV2021/html/Appalaraju_DocFormer_End-to-End_Transformer_for_Document_Understanding_ICCV_2021_paper.html}
}

@inproceedings{kim2022ocrfree,
  title={OCR-Free Document Understanding Transformer},
  author={Kim, Geewook and Hong, Teakgyu and Yim, Moonbin and Nam, JeongYeon and Park, Jinyoung and Yim, Jinyeong and Hwang, Wonseok and Yun, Sangdoo and Han, Dongyoon and Park, Seunghyun},
  booktitle={Computer Vision -- ECCV 2022},
  pages={498--517},
  year={2022},
  publisher={Springer},
  doi={10.1007/978-3-031-19815-1_29},
  url={https://link.springer.com/chapter/10.1007/978-3-031-19815-1_29}
}

@inproceedings{lee2023pix2struct,
  title={Pix2Struct: Screenshot Parsing as Pretraining for Visual Language Understanding},
  author={Lee, Kenton and Joshi, Mandar and Turc, Iulia and Hu, Hexiang and Liu, Fangyu and Eisenschlos, Julian and Khandelwal, Urvashi and Shaw, Peter and Chang, Ming-Wei and Toutanova, Kristina},
  booktitle={Proceedings of the 40th International Conference on Machine Learning},
  pages={18893--18912},
  year={2023},
  organization={PMLR},
  url={https://proceedings.mlr.press/v202/lee23g.html}
}

@inproceedings{blecher2024nougat,
  title={Nougat: Neural Optical Understanding for Academic Documents},
  author={Blecher, Lukas and Cucurull, Guillem and Scialom, Thomas and Stojnic, Robert},
  booktitle={International Conference on Learning Representations},
  year={2024},
  url={https://openreview.net/forum?id=fUtxNAKpdV}
}

@inproceedings{wang2024docllm,
  title={DocLLM: A Layout-Aware Generative Language Model for Multimodal Document Understanding},
  author={Wang, Dongsheng and Raman, Natraj and Sibue, Mathieu and Ma, Zhiqiang and Babkin, Petr and Kaur, Simerjot and Pei, Yulong and Nourbakhsh, Armineh and Liu, Xiaomo},
  booktitle={Proceedings of the 62nd Annual Meeting of the Association for Computational Linguistics},
  year={2024},
  url={https://aclanthology.org/2024.acl-long.463/}
}

@inproceedings{radford2021learning,
  title={Learning Transferable Visual Models From Natural Language Supervision},
  author={Radford, Alec and Kim, Jong Wook and Hallacy, Chris and Ramesh, Aditya and Goh, Gabriel and Agarwal, Sandhini and Sastry, Girish and Askell, Amanda and Mishkin, Pamela and Clark, Jack and Krueger, Gretchen and Sutskever, Ilya},
  booktitle={Proceedings of the 38th International Conference on Machine Learning},
  pages={8748--8763},
  year={2021},
  organization={PMLR},
  url={https://proceedings.mlr.press/v139/radford21a.html}
}

@inproceedings{li2023blip2,
  title={BLIP-2: Bootstrapping Language-Image Pre-training with Frozen Image Encoders and Large Language Models},
  author={Li, Junnan and Li, Dongxu and Savarese, Silvio and Hoi, Steven},
  booktitle={Proceedings of the 40th International Conference on Machine Learning},
  pages={19730--19742},
  year={2023},
  organization={PMLR},
  url={https://proceedings.mlr.press/v202/li23q.html}
}

@inproceedings{liu2023visual,
  title={Visual Instruction Tuning},
  author={Liu, Haotian and Li, Chunyuan and Wu, Qingyang and Lee, Yong Jae},
  booktitle={Advances in Neural Information Processing Systems},
  volume={36},
  year={2023},
  url={https://proceedings.neurips.cc/paper_files/paper/2023/hash/6dcf277ea32ce3288914faf369fe6de0-Abstract-Conference.html}
}

@misc{bai2023qwenvl,
  title={Qwen-VL: A Versatile Vision-Language Model for Understanding, Localization, Text Reading, and Beyond},
  author={Bai, Jinze and Bai, Shuai and Yang, Shusheng and Wang, Shijie and Tan, Sinan and Wang, Peng and Lin, Junyang and Zhou, Chang and Zhou, Jingren},
  year={2023},
  eprint={2308.12966},
  archivePrefix={arXiv},
  primaryClass={cs.CV},
  url={https://arxiv.org/abs/2308.12966}
}

@misc{bai2025qwen25vl,
  title={Qwen2.5-VL Technical Report},
  author={Bai, Shuai and Chen, Keqin and Liu, Xuejing and Wang, Jialin and Ge, Wenbin and Song, Sibo and Dang, Kai and Wang, Peng and Wang, Shijie and Tang, Jun and others},
  year={2025},
  eprint={2502.13923},
  archivePrefix={arXiv},
  primaryClass={cs.CV},
  url={https://arxiv.org/abs/2502.13923}
}

@inproceedings{karpukhin2020dense,
  title={Dense Passage Retrieval for Open-Domain Question Answering},
  author={Karpukhin, Vladimir and Oguz, Barlas and Min, Sewon and Lewis, Patrick and Wu, Ledell and Edunov, Sergey and Chen, Danqi and Yih, Wen-tau},
  booktitle={Proceedings of the 2020 Conference on Empirical Methods in Natural Language Processing},
  pages={6769--6781},
  year={2020},
  doi={10.18653/v1/2020.emnlp-main.550},
  url={https://aclanthology.org/2020.emnlp-main.550/}
}

@inproceedings{xiong2021approximate,
  title={Approximate Nearest Neighbor Negative Contrastive Learning for Dense Text Retrieval},
  author={Xiong, Lee and Xiong, Chenyan and Li, Ye and Tang, Kwok-Fung and Liu, Jialin and Bennett, Paul and Ahmed, Junaid and Overwijk, Arnold},
  booktitle={International Conference on Learning Representations},
  year={2021},
  url={https://openreview.net/forum?id=zeFrfgyZln}
}

@inproceedings{formal2021splade,
  title={SPLADE: Sparse Lexical and Expansion Model for First Stage Ranking},
  author={Formal, Thibault and Piwowarski, Benjamin and Clinchant, Stephane},
  booktitle={Proceedings of the 44th International ACM SIGIR Conference on Research and Development in Information Retrieval},
  pages={2288--2292},
  year={2021},
  doi={10.1145/3404835.3463098},
  url={https://doi.org/10.1145/3404835.3463098}
}

@inproceedings{khattab2020colbert,
  title={ColBERT: Efficient and Effective Passage Search via Contextualized Late Interaction over BERT},
  author={Khattab, Omar and Zaharia, Matei},
  booktitle={Proceedings of the 43rd International ACM SIGIR Conference on Research and Development in Information Retrieval},
  pages={39--48},
  year={2020},
  doi={10.1145/3397271.3401075},
  url={https://doi.org/10.1145/3397271.3401075}
}

@inproceedings{santhanam2022colbertv2,
  title={ColBERTv2: Effective and Efficient Retrieval via Lightweight Late Interaction},
  author={Santhanam, Keshav and Khattab, Omar and Saad-Falcon, Jon and Potts, Christopher and Zaharia, Matei},
  booktitle={Proceedings of the 2022 Conference of the North American Chapter of the Association for Computational Linguistics},
  pages={3715--3734},
  year={2022},
  doi={10.18653/v1/2022.naacl-main.272},
  url={https://aclanthology.org/2022.naacl-main.272/}
}

@article{izacard2022unsupervised,
  title={Unsupervised Dense Information Retrieval with Contrastive Learning},
  author={Izacard, Gautier and Caron, Mathilde and Hosseini, Lucas and Riedel, Sebastian and Bojanowski, Piotr and Joulin, Armand and Grave, Edouard},
  journal={Transactions on Machine Learning Research},
  year={2022},
  url={https://openreview.net/forum?id=jKN1pXi7b0}
}

@inproceedings{lewis2020retrieval,
  title={Retrieval-Augmented Generation for Knowledge-Intensive NLP Tasks},
  author={Lewis, Patrick and Perez, Ethan and Piktus, Aleksandra and Petroni, Fabio and Karpukhin, Vladimir and Goyal, Naman and Kuttler, Heinrich and Lewis, Mike and Yih, Wen-tau and Rocktaschel, Tim and others},
  booktitle={Advances in Neural Information Processing Systems},
  volume={33},
  pages={9459--9474},
  year={2020},
  url={https://proceedings.neurips.cc/paper/2020/hash/6b493230205f780e1bc26945df7481e5-Abstract.html}
}

@misc{jiang2024e5v,
  title={E5-V: Universal Embeddings with Multimodal Large Language Models},
  author={Jiang, Ting and Song, Minghui and Zhang, Zihan and Huang, Haizhen and Deng, Weiwei and Sun, Feng and Zhang, Qi and Wang, Deqing and Zhuang, Fuzhen},
  year={2024},
  eprint={2407.12580},
  archivePrefix={arXiv},
  primaryClass={cs.CL},
  url={https://arxiv.org/abs/2407.12580}
}

@misc{lin2024mmembed,
  title={MM-Embed: Universal Multimodal Retrieval with Multimodal LLMs},
  author={Lin, Sheng-Chieh and Lee, Chankyu and Shoeybi, Mohammad and Lin, Jimmy and Catanzaro, Bryan and Ping, Wei},
  year={2024},
  eprint={2411.02571},
  archivePrefix={arXiv},
  primaryClass={cs.CL},
  url={https://arxiv.org/abs/2411.02571}
}

@article{tesseract,
author = {Kay, Anthony},
title = {Tesseract: an open-source optical character recognition engine},
year = {2007},
issue_date = {July 2007},
publisher = {Belltown Media},
address = {Houston, TX},
volume = {2007},
number = {159},
issn = {1075-3583},
journal = {Linux J.},
month = jul,
pages = {2}
}

@article{bm25,
author = {Robertson, Stephen and Zaragoza, Hugo},
title = {The Probabilistic Relevance Framework: BM25 and Beyond},
year = {2009},
issue_date = {April 2009},
publisher = {Now Publishers Inc.},
address = {Hanover, MA, USA},
volume = {3},
number = {4},
issn = {1554-0669},
url = {https://doi.org/10.1561/1500000019},
doi = {10.1561/1500000019},
journal = {Found. Trends Inf. Retr.},
month = apr,
pages = {333–389},
numpages = {57}
}

@misc{bge-m3,
      title={BGE M3-Embedding: Multi-Lingual, Multi-Functionality, Multi-Granularity Text Embeddings Through Self-Knowledge Distillation}, 
      author={Jianlv Chen and Shitao Xiao and Peitian Zhang and Kun Luo and Defu Lian and Zheng Liu},
      year={2024},
      eprint={2402.03216},
      archivePrefix={arXiv},
      primaryClass={cs.CL}
}

@inproceedings{RRF,
author = {Cormack, Gordon V. and Clarke, Charles L A and Buettcher, Stefan},
title = {Reciprocal rank fusion outperforms condorcet and individual rank learning methods},
year = {2009},
isbn = {9781605584836},
publisher = {Association for Computing Machinery},
address = {New York, NY, USA},
url = {https://doi.org/10.1145/1571941.1572114},
doi = {10.1145/1571941.1572114},
booktitle = {Proceedings of the 32nd International ACM SIGIR Conference on Research and Development in Information Retrieval},
pages = {758–759},
numpages = {2},
location = {Boston, MA, USA},
series = {SIGIR '09}
}
}


\appendix

\definecolor{colorColPali}{RGB}{228,26,28}
\definecolor{colorColQwen25}{RGB}{55,126,184}
\definecolor{colorColNomic}{RGB}{77,175,74}
\definecolor{colorColSmol}{RGB}{152,78,163}
\definecolor{colorColModernVBERT}{RGB}{255,127,0}
\definecolor{colorColQwen3}{RGB}{166,86,40}


\newpage
\section{Benchmark Details}
We present in the following the benchmark creation details and further statistics for the document dataset and benchmark query dataset.

\subsection{Document Collection}
\noindent \textbf{Wikipedia:} For the collection of Wikipedia articles, we build an intelligent spider that recursively searches for new Wikipedia articles.
As a starting point, we manually select 78 seed articles across domains such as geography, sports, biology, chemistry, arts, history, and economics. Additionally, we craft for each seed article a query instruction for further articles that are related to the seed article but likely do not represent a duplication of article content. For example, the query instruction for the Wikipedia article about "Mount Everest" is "\textit{Search for articles which are about mountains. The title shall contain "Mount" or the name of a peak.}".
Starting from the seed articles, the intelligent spider recursively searches in the Wikipedia articles for all referenced articles. The spider employs an LLM to decide which articles to select. The prompt for the LLM includes the query instruction together with the referenced articles' titles.

\begin{tcolorbox}[
  colback=gray!5,
  colframe=gray!50,
  title=Wikipedia Spider Article Selection Prompt,
  fonttitle=\bfseries,
  breakable
]
\ttfamily
\textbf{System:}
You are a precise classifier. Your OUTPUT MUST be valid JSON and nothing else.
Do NOT provide chain-of-thought or step-by-step reasoning. Short reasons only (<=10 words). \\
 \\
\textbf{User:}
Description of desired wikipedia articles:\\
"<QUERY\_INSTRUCTION>" \\
\\
Max results: <INT> \\
\\
Candidate articles (index: title):\\
<LIST\_OF\_ARTICLE\_TITLE\_NAMES\_WITH\_INDEX\_NUMBERS> \\
\\
Return JSON with this exact shape:\\
\{ "matches": [ \{ "index": <int>, "score": <float between 0 and 1>, "reason": "<short reason>" \}, ... ] \} \\
\\
Only include articles you consider relevant (omit irrelevant ones). Sort matches by score descending. If none match, return \{"matches": []\}.
\end{tcolorbox}

\noindent \textbf{arXiv:} In contrast to the Wikipedia spider, the arXiv article indexer does not employ an LLM for determining which articles to visit next. Instead, the indexer simply searches for all available links to other arXiv articles. The reason for this decision is that scientific article titles are often ambiguous and do not fit into a clearly specified category, such as e.g. ''Mount Everest'' in the case of Wikipedia articles. Initially, we use a similar approach to the Wikipedia article indexing with 41 manually searched seed articles. The seed articles consist of recent survey articles. We select them because they contain references to closely related articles, are suitable for Cross-Doc query generation, and also contain more likely HTML versions due to their recency.

Since this approach leads to less than 500 arXiv articles, we extend the initial seed articles by 20,000 articles from the arXiv categories astrophysics, condensed matter, quantitative biology, and electrical engineering resulting in a total of 36,132 articles. We decide to create a larger initial index for arXiv to account for removal due to license restrictions and invalidly converted arXiv articles.

\noindent \textbf{PubMed:} In contrast to arXiv, PubMed articles contain fewer references with URLs to open-access articles. Therefore, we decide to retrieve a list of 9,488 open-access articles from 2025 to 2026 directly from PubMed Central instead of a recursive search. Furthermore, we collect for each article licensing information to later filter out restricted articles.

\noindent \textbf{SEC Proxy Statements:} We select about 2k large publicly listed companies' most recent proxy statements, especially due to their high document length. Since proxy statements contain overlapping information between multiple years, leading to ambiguous retrieval problems, we decide to use only one proxy statement per company.

\subsection{Document Items}
\begin{figure}[htbp]
  \centering
  \includegraphics[
      width=0.85\textwidth,
      trim=0 2000 0 0,
      clip
  ]{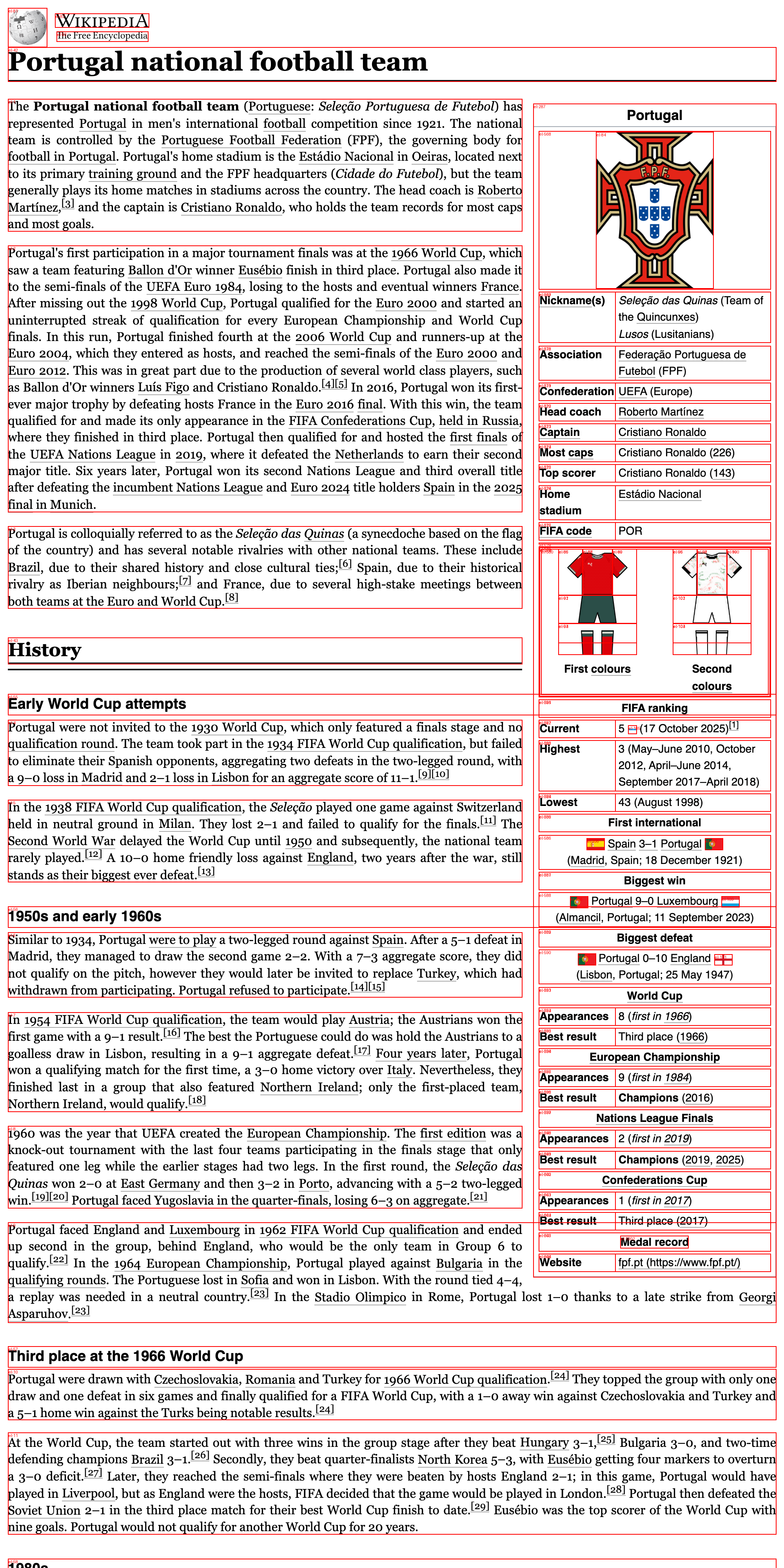}
  \caption[Bounding-Box Annotation Visualization]{Bounding-box annotation visualization for the Wikipedia article about the Portugal national football team.}
  \label{fig:bounding_box_annotation_visualization}
\end{figure}

The main components of a document item are the equally-sized page images, the underlying HTML with injected identifiers, and pixel-exact measured bounding-boxes. Figure \ref{fig:bounding_box_annotation_visualization} visualizes the bounding-boxes for an example document page cutout. Apart from this, we also collect abstracts, licensing, and citations of the document.

\subsection{Document Dataset Statistics}
\begin{figure}[htbp]
    \begin{tikzpicture}
    \begin{axis}[
        ybar,
        width=14cm,
        height=6cm,
        bar width=12pt,
        ymin=0,
        ylabel={Number of Documents per Domain},
        symbolic x coords={
            Astrophysics,Science,Sports,{Medical Reports},{Condensed Matter Physics},Geography,{Artificial Intelligence},{Financial Reports},History,{Quantitative Biology},Entertainment,{Electrical Engineering},Bioinformatics,Technology,{Public Health Reports},Arts,{General Computer Science},Chemistry,Politics,{Environmental Reports},{Statistics},{Surgical Case Reports},{Economics}
        },
        xtick=data,
        x tick label style={rotate=45, anchor=east},
        tick label style={font=\scriptsize},
        label style={font=\small},
        nodes near coords,
        nodes near coords align={vertical},
        every node near coord/.append style={
            font=\scriptsize,
            rotate=45,
            anchor=west
        },
        enlarge x limits=0.05,
        ymajorgrids=true,
        grid style={opacity=0.25},
    ]
    \addplot coordinates {
        (Astrophysics,3497)
        (Science,3053)
        (Sports,3007)
        ({Medical Reports},2820)
        ({Condensed Matter Physics},2704)
        (Geography,2659)
        ({Artificial Intelligence},2273)
        ({Financial Reports},1992)
        (History,1099)
        ({Quantitative Biology},1042)
        (Entertainment,747)
        ({Electrical Engineering},630)
        (Bioinformatics,600)
        (Technology,551)
        ({Public Health Reports},545)
        (Arts,527)
        ({General Computer Science},485)
        (Chemistry,396)
        (Politics,293)
        ({Environmental Reports},258)
        ({Statistics},214)
        ({Surgical Case Reports},197)
        ({Economics},52)
    };
    \end{axis}
    \end{tikzpicture}
    \caption{Domain distribution in the document dataset.}
    \label{fig:docment_dataset_domain_distribution}
\end{figure}
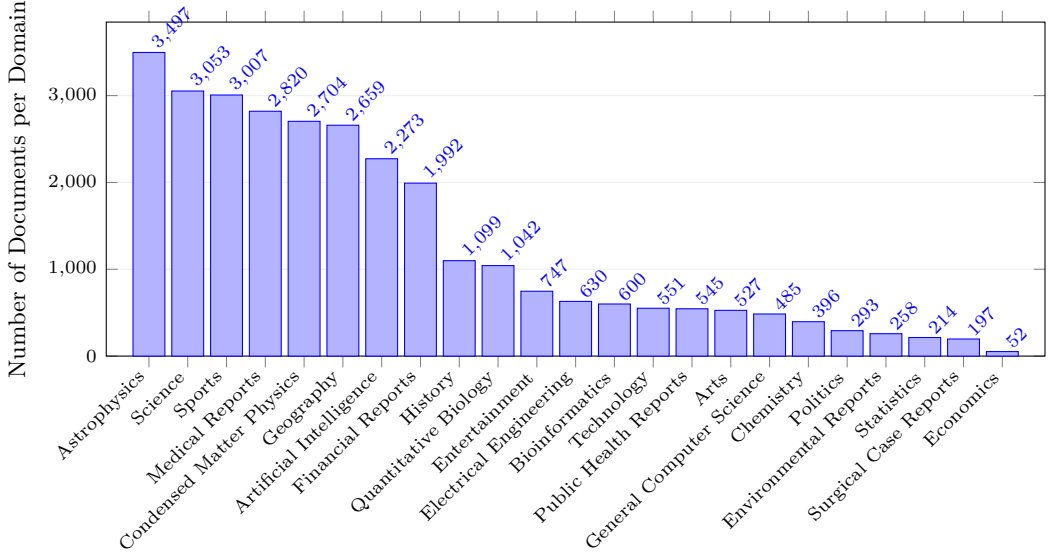

In this subsection, we provide further insights into the structure of our document dataset. Figure \ref{fig:docment_dataset_domain_distribution} illustrates the domain distribution in the document dataset. Our VL-DocIR benchmark covers 23 domains: \textit{Arts}, \textit{Entertainment}, \textit{Science}, \textit{Technology}, \textit{Geography}, \textit{Sports}, \textit{History}, \textit{Politics}, \textit{Economics}, \textit{Astrophysics}, \textit{Condensed Matter Physics}, \textit{Quantitative Biology}, \textit{Electrical Engineering}, \textit{General Computer Science}, \textit{Artificial Intelligence}, \textit{Statistics}, \textit{Surgical Case Reports}, \textit{Public Health Reports}, \textit{Environmental Reports}, \textit{Medical Reports}, \textit{Chemistry}, \textit{Bioinformatics}, and \textit{Financial Reports}. The most occurring domain is \textit{Astrophysics} with 3,497 documents followed by the Wikipedia article domains \textit{Science} and \textit{Sports} with 3,053 and 3,007 documents. In general the document domain distribution is balanced and we have at least 50 documents per domain.

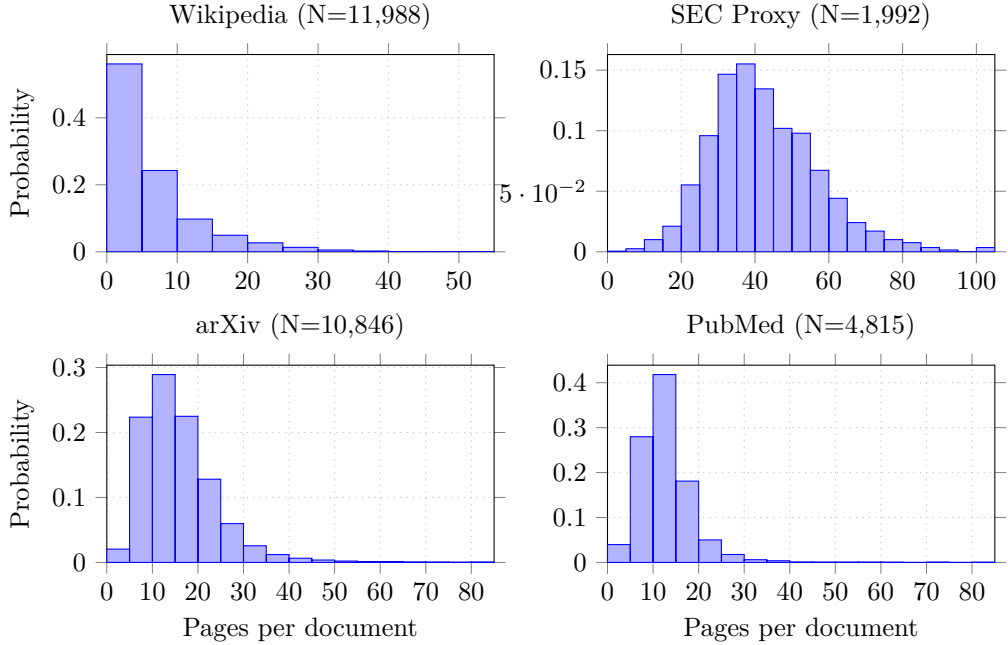
\begin{figure}[htb]
\centering
\begin{tikzpicture}
\begin{groupplot}[
  group style={
    group size=2 by 2,
    horizontal sep=1.5cm,
    vertical sep=1.5cm,
    xlabels at=edge bottom,
    ylabels at=edge left,
  },
  width=0.48\textwidth,
  height=0.3\textwidth,
  ybar interval,
  ymin=0,
  xmin=0,
  enlarge x limits=false,
  enlarge y limits={upper, value=0.05},
  ylabel={Probability},
  xlabel={Pages per document},
  xticklabel pos=lower,
  xticklabel style={anchor=north},
  grid=major,
  grid style={dotted},
  tick align=outside,
]

\nextgroupplot[
  title={Wikipedia (N=11{,}988)},
  xmin=0, xmax=55,
  xtick={0,10,20,30,40,50},
  extra x tick style={major tick length=0pt},
  x tick label as interval=false,
]
\addplot+[fill, draw] table[col sep=comma, x=left, y=pct] {Data/wiki_interval5_overflow.csv};

\nextgroupplot[
  title={SEC Proxy (N=1{,}992)},
  xmin=0, xmax=105,
  xtick={0,20,40,60,80,100},
  extra x tick style={major tick length=0pt},
  x tick label as interval=false,
]
\addplot+[fill, draw] table[col sep=comma, x=left, y=pct] {Data/sec_interval5_overflow.csv};

\nextgroupplot[
  title={arXiv (N=10{,}846)},
  xmin=0, xmax=85,
  xtick={0,10,20,30,40,50,60,70,80},
  extra x tick style={major tick length=0pt},
  x tick label as interval=false,
]
\addplot+[fill, draw] table[col sep=comma, x=left, y=pct] {Data/arxiv_interval5_overflow.csv};

\nextgroupplot[
  title={PubMed (N=4{,}815)},
  xmin=0, xmax=85,
  xtick={0,10,20,30,40,50,60,70,80},
  extra x tick style={major tick length=0pt},
  x tick label as interval=false,
]
\addplot+[fill, draw] table[col sep=comma, x=left, y=pct] {Data/pubmed_interval5_overflow.csv};

\end{groupplot}
\end{tikzpicture}

\caption[Page-Count Distribution]{Page-count distributions with 5-page bins for every data source. The plots show that Wikipedia articles tend to be shorter documents, whereas SEC Proxy Statements are much longer. The page count distribution of arXiv and PubMed articles is similar, with the most articles having between 10 and 15 pages.}
\label{fig:page_count_dist_5bin}
\end{figure}

Figure \ref{fig:page_count_dist_5bin} visualizes the distribution of pages per document for the four data sources. The distribution of Wikipedia highlights that most articles have a length between 1 and 10. Both scientific arXiv and PubMed article lengths are within 5 and 25. The distribution of SEC Proxy Statement page lengths deviates from the others, with the majority of documents being between 20 and 60 pages long.

\subsection{Query Generation}
\label{subsec:appendix_question_generation}
For each query generation configuration entry, the query generation consists of two steps. First, the page images are described using a VLM. In the second step, we prompt a VLM to generate 10 queries for the given input including the page summaries, underlying HTML with the injected element identifiers, and the raw page images. We use two different combinations of VLMs. The first combination is to use for both steps Qwen3-VL-32B \cite{bai2025qwen3}, and the second combination uses for the first step the OpenAI GPT-5-nano model \cite{openai2025gpt5nano} and for the second step GPT-4.1-mini \cite{openai2024gpt41mini}. The cause for splitting into two different model combinations is that initial qualitative tests have revealed a better query quality for the OpenAI combination, however, being accompanied by higher costs. Therefore, we process for each data source and for each query evidence structure type 2,000 configuration entries with Qwen3-VL-32B and 500 configuration entries with the OpenAI models combination.

\noindent \textbf{Page Image Analysis:} For the page image analysis step, we use different prompts for the Qwen3-VL-32B and GPT-5-nano since GPT-5-nano is a reasoning model that requires fewer instructions than Qwen3-VL-32B. The multi-page and Cross-Doc query creators apply the page image analysis to each page individually. Since the query generation step utilizes also the HTML, we focus the prompt on describing visual content in the images.
\begin{tcolorbox}[
  colback=gray!5,
  colframe=gray!50,
  title=Page Image Analysis Prompt for Qwen3-VL-32B (temperature: 0.8),
  fonttitle=\bfseries,
  breakable
]
\ttfamily
\small
\textbf{User:}
Describe in detail all visual elements (images, diagrams, graphics, photos, maps, charts, or tables) appearing on the document page shown in the attached image.\\
For each visual element, specify:\\
- A detailed description of what it depicts, including visible objects, people, text labels, numerical data, and symbols. Include i.e. facts, statistics, and concepts visible in pictures. Pay attention to details and name all of them.\\
\\
Avoid summarizing or explaining meaning, just describe what is visibly present. \\
\end{tcolorbox}

\begin{tcolorbox}[
  colback=gray!5,
  colframe=gray!50,
  title=Page Image Analysis Prompt for GPT-5-nano,
  fonttitle=\bfseries,
  breakable
]
\ttfamily
\small
\textbf{User:}
Describe in detail what the images/diagrams/graphics/photos/maps in the attached image include
\end{tcolorbox}

\noindent \textbf{General Query Generation:} For the query generation step, the prompts differ depending on the data source, query evidence structure type, and VLM. However, they share in general all the same prompt structure consisting of the task description, output format description, and the input data. The goal of each VLM prompt is to get a list of queries together with the answers and HTML element identifiers as evidence.

\begin{tcolorbox}[
  colback=gray!5,
  colframe=gray!50,
  title=Query Generation Prompt Structure,
  fonttitle=\bfseries,
  breakable
]
\ttfamily
\textbf{User:} \\
<TASK\_DESCRIPTION (high-level, criteria, good and bad examples)> \\
<OUTPUT\_DESCRIPTION (thinking steps, JSON output format)> \\
<INPUT\_DATA (article titles, page image descriptions, HTML with element identifiers)>
\end{tcolorbox}

The task description always starts with a high-level instruction explaining the task to the VLM. Following, we provide a detailed description of the criteria for the queries to generate together with good and bad query examples. Furthermore, we formulate restrictions for the thinking process. 
The output description contains a definition of necessary thinking steps for the VLM, as well as the required JSON output format. We instruct the VLM to use a split string to separate thinking from JSON, since we do not use a reasoning model to reduce compute time. 
The input data prompt part comprises article titles, the previously generated page image descriptions, and the HTML with element identifiers.

In case of missing splitters, invalid JSON format, or invalid HTML element identifier, we discard the generation output. For valid output, we extract the list of queries, answers, and HTML element identifiers. Thereon, we determine via the HTML element identifiers the exact page-relative bounding boxes and store the results.

\begin{tcolorbox}[
  colback=gray!5,
  colframe=gray!50,
  title=Single-Page Query Generation Prompt for Wikipedia (Qwen3-VL-32B),
  fonttitle=\bfseries,
  breakable
]
\ttfamily
\small
\textbf{User:}\\
\#\#\# Task:\\
You are given:\\
1. A **document page image**.\\
2. The corresponding **HTML snippet**, which provides element IDs and layout information.\\
3. A description of the document page image\\
\\
Your task is to generate **10 questions** about the textual and visual content of the document on the given page.\\
Good questions are:\\
- clearly and precisely formulated\\
- unambiguous,\\
- fact-based,\\
- closed: include all required information in the question to answer it without the knowledge of the document page.\\
- about knowledge, facts, trends, statistics, ... contained in graphics, charts, photos, tables, diagrams, maps, pictures\\
- do not explicitly reference the page layout. Do NOT use terms such as 'in the table', 'in the figure', or 'according to the image'! Instead, try to extract from the tables and figures facts that are answerable in general.\\
- do not explicitly reference the document/article (do NOT use 'in the article', 'in the document', 'in the report', ...). Instead, use the name of the article, entities, events, ... to specify the context of each question.\\
- independent from each other, which means that the necessary context to answer them is included in each question\\
\\
Good question examples:\\
- Good: What is the bond angle between the two hydrogen atoms in a water molecule?\\
- Good: What are the neighboring countries of France? \\
- Good: What was the total annual revenue of Apple Inc. in the fiscal year 2025?\\
- Good: Which aircraft did Fokker assemble for the Royal Netherlands Air Force that arrived for the Royal International Air Tattoo in England in 2014?\\
\\
Bad question examples:\\
- Bad: What are the bony structures visible in the CT images of Figure 4, and how are they represented?\\
- Bad: What is the name of the galaxy cluster shown in the photograph at the top of the page?\\
- Bad: Which persons can be seen in the picture?\\
\\
Furthermore, make no assumption that a question answerer knows about which document you are asking the questions.\\
Imagine that the person who answers the questions does not know which document the questions are about. Use precise questions with named methods, concepts, experiments, ... .\\
\begin{verbatim}<MULTI_EVIDENCE_USAGE_INSTRUCTION (optional)>\end{verbatim}

\begin{verbatim}The document is a Wikipedia article about <ARTICLE_TITLE>.\end{verbatim}

\#\#\# Output\\
The answer consists of multiple parts: \\
Start with:\\
    1) Reflect on the document and extract general knowledge of the document\\
    2) Draft a list of questions and answers (draft more than 10 questions so that you can select thereafter the best ones)\\
    3) Analyze each question with the criteria above\\
    4) Select the questions that best match the question requirement from above\\
    \\
    Then print exactly \_\_\_SPLITTER\_\_\_ (after \_\_\_SPLITTER\_\_\_ comes only json)\\
\\
    Create the final JSON result: Return a **valid JSON array** following this exact schema:\\
\begin{verbatim}
[
  {
    "question": "<string>",
    "answer": "<string>",
    "evidence": ["<element_id_1>", "<element_id_2>", ...]
  },
  ...
]
\end{verbatim}

\begin{verbatim}---\end{verbatim}
\#\#\# Input:\\
This is the cleaned HTML snippet associated with the attached image:
\begin{verbatim}"""<HTML_WITH_ELEMENT_IDS>"""\end{verbatim}

There is also a detailed description of the document page image:
\begin{verbatim}"""<PAGE_IMAGE_DESCRIPTION>"""\end{verbatim}
\end{tcolorbox}

\noindent \textbf{Multi-Page Queries:} The prompts for multi-page query generation are similar to the single-page query generation prompts. The main difference is that there are two pages given as input, and the instruction to create only queries that are unanswerable with a single page of evidence. The further query prompts can be seen in our published benchmark creation program. Between Doc-Adjacent, Doc-Near, and Doc-Far queries, there is not difference in the generation process and we assign the query evidence structure afterwards.

\noindent \textbf{Cross-Doc Queries:} For Cross-Doc queries, the prompt differs more strongly compared to single-page and multi-page queries. Since we have two documents, the evidence localization requires determining not only the HTML element identifier, which are integers starting with 0, but also the document identifier. In the initial prompt evaluation, we find out that when naming the documents with the terms "Document 1" and "Document 2", the VLMs tend to use those also inside the queries itself. Since this is not suitable for an open-domain queries dataset, we use identifiers "12345" and "54321" for the document identifiers to reduce their usage in queries. Furthermore, we reference them by using "first" and "second" instead of a number. As in the case of the multi-page query generation prompt, we instruct the model to use evidence of both documents to create queries.

\subsection{Query Cleaning Evaluation Dataset}
\label{subsec:appendix_question_cleaning_evaluation_dataset}
In this subsection, we provide further details regarding the evaluation of our query removal methods. We evaluate each method on our query removal evaluation dataset, which consists of 1,438 manually labeled queries.

\noindent \textbf{LLM-Rating Method:} The query LLM rating method leverages a Qwen3-VL-32B as an evaluator. Instead of directly determining the removal class, we let the VLM rate the query together with the answer by the following four criteria with scores within ${0, 0.25, 0.5, 0.75, 1}$:
\begin{itemize}
    \item Clarity: not ambiguously formulated
    \item Answer correctness: in case that we do not know the answer, we use plausibility as a measure
    \item Closedness: no reference of articles / studies / documents without naming them (e.g. not ''in the document'')
    \item No layout reference: no terms such as "in the table"
\end{itemize}
For the removal of queries, we test thresholds of 0.75, 0.5, 0.25, and 0.0 for queries to be removed. 
If any of the four criteria scores falls below a threshold, then we consider a query as removed. The classification results can be seen in Table \ref{tab:confusion_matrices_question_llm_evaluation_removal_classification_thresholds}. The false omission rate, where omission means, in this case, non-removal of a query, is for all four threshold nearly the same. The removal with threshold 0.75 achieves the best false omission rate with 13.37\%.

\begin{table}[htb]
\centering
\renewcommand{\arraystretch}{1.2}

\begin{subtable}{0.22\linewidth}
\centering
\begin{tabular}{c|cc}
 & P & N \\
\hline
P & \cellcolor{green!25} 279 & \cellcolor{red!15} 161 \\
N & \cellcolor{red!15} 65  & \cellcolor{green!25} 933 \\
\end{tabular}
\caption{Threshold = 0.0 \\ FOR = 0.1472}
\end{subtable}
\hfill
\begin{subtable}{0.22\linewidth}
\centering
\begin{tabular}{c|cc}
 & P & N \\
\hline
P & \cellcolor{green!25} 281 & \cellcolor{red!15} 159 \\
N & \cellcolor{red!15} 71  & \cellcolor{green!25} 927 \\
\end{tabular}
\caption{Threshold = 0.25 \\ FOR = 0.1464}
\end{subtable}
\hfill
\begin{subtable}{0.22\linewidth}
\centering
\begin{tabular}{c|cc}
 & P & N \\
\hline
P & \cellcolor{green!25} 281 & \cellcolor{red!15} 159 \\
N & \cellcolor{red!15} 77  & \cellcolor{green!25} 921 \\
\end{tabular}
\caption{Threshold = 0.5 \\ FOR = 0.1472}
\end{subtable}
\hfill
\begin{subtable}{0.22\linewidth}
\centering
\begin{tabular}{c|cc}
 & P & N \\
\hline
P & \cellcolor{green!25} 294 & \cellcolor{red!15} 146 \\
N & \cellcolor{red!15} 128 & \cellcolor{green!25} 870 \\
\end{tabular}
\caption{Threshold = 0.75 \\ FOR = 0.1437}
\end{subtable}

\caption[Confusion Matrices for Different Question LLM-Evaluation Removal Classification Thresholds]{Confusion matrices for different question LLM-evaluation removal classification thresholds (total samples = 1438). Positive P stands for a question being removed. Negative N stands for a question not being removed. FOR is the false omission rate, where omission means non-removal.}
\label{tab:confusion_matrices_question_llm_evaluation_removal_classification_thresholds}
\end{table}

\noindent \textbf{LLM-Classifier Method:} Similar to the LLM rating removal method, the LLM classifier method makes use of Qwen3-VL-32B to evaluate a query. However, instead of producing scores, the strategy directly prompts the LLM to classify a query into an accept-class or discard-class.

Compared to the LLM rating remover, the performance is with a false omission rate of~17.82\%. A potential cause for this is that the score creation instructs the LLM to think in more detail about the query and be more critical.

\begin{table}[ht]
\centering
\renewcommand{\arraystretch}{1.3}
\begin{tabular}{c|cc}
 & \textbf{Predicted P} & \textbf{Predicted N} \\
\hline
\textbf{Actual P} &
\cellcolor{green!25} 237 &
\cellcolor{red!15} 203 \\
\textbf{Actual N} &
\cellcolor{red!15} 62 &
\cellcolor{green!25} 936 \\
\end{tabular}
\vspace{7pt}
\caption[Confusion Matrix for LLM Query Removal Classification Strategy]{Confusion matrix for LLM query removal classification strategy (total samples = 1438, false omission rate = 0.1782).}
\end{table}

Furthermore, we analyze the impact of combining the LLM classifier method with the query LLM rating method. In the combination, we remove queries if at least one of the two methods predicts a removal. For the query LLM rating method we use the threshold 0.75 for the rating scores because the false omission rate is the best for this value. As the confusion matrix Table \ref{tab:confusion_matrix_combination_llm_classifier_And_llm_rating_methods} depicts, the combination of both methods drops the false omission rate compared to the pure LLM rating method further to 13.60\%.

\begin{table}[ht]
\centering
\renewcommand{\arraystretch}{1.3}
\begin{tabular}{c|cc}
 & \textbf{Predicted P} & \textbf{Predicted N} \\
\hline
\textbf{Actual P} &
\cellcolor{green!25} 305 &
\cellcolor{red!15} 135 \\
\textbf{Actual N} &
\cellcolor{red!15} 140 &
\cellcolor{green!25} 858 \\
\end{tabular}
\vspace{7pt}
\caption[Confusion Matrix for Combination of LLM Classifier and Query LLM Rating Removal Methods]{Confusion matrix for combination of LLM classifier and Query LLM rating removal methods (total samples = 1438, false omission rate = 0.1360).}
\label{tab:confusion_matrix_combination_llm_classifier_And_llm_rating_methods}
\end{table}

\noindent \textbf{Removal Classifiers:} The input to the classifiers are text embeddings of the queries created with Alibaba text-embedding-v4. We choose 800 samples from the query removal dataset for the training set, leaving 638 samples for the evaluation. The results indicate that XGBoost is with a false omission rate of 12.46\% slightly better than logistic regression with 13.07\%. However, in terms of the resulting dataset, the queries count would be likely lower, because logistic regression predicts for the evaluation set 392 query non-removals, whereas XGBoost predicts 353 removals. The detailed classification results can be seen in the following Table \ref{tab:confusion_matrices_question_classifier_removal_classification_thresholds}.

\begin{table}[htb]
\centering
\renewcommand{\arraystretch}{1.2}

\begin{subtable}{0.45\linewidth}
\centering
\begin{tabular}{c|cc}
 & P & N \\
\hline
P & \cellcolor{green!25} 141 & \cellcolor{red!15} 52 \\
N & \cellcolor{red!15} 99  & \cellcolor{green!25} 346 \\
\end{tabular}
\caption{Logistic Regression \\ FOR = 0.1307}
\end{subtable}
\hfill
\begin{subtable}{0.45\linewidth}
\centering
\begin{tabular}{c|cc}
 & P & N \\
\hline
P & \cellcolor{green!25} 150 & \cellcolor{red!15} 44 \\
N & \cellcolor{red!15} 135  & \cellcolor{green!25} 309 \\
\end{tabular}
\caption{XGBoost \\ FOR = 0.1246}
\end{subtable}

\caption[Confusion Matrices for the Logistic Regression and XGBoost Classifiers for Query Removal]{Confusion matrices for the Logistic Regression and XGBoost classifier for query removal. Positive P stands for a query being removed. Negative N stands for a query not being removed. FOR is the false omission rate, where omission means non-removal.}
\label{tab:confusion_matrices_question_classifier_removal_classification_thresholds}
\end{table}

\noindent \textbf{Nearest-Neighbor Method:} As for the classifiers, we use the query text embeddings generated with Alibaba text-embedding-v4 and 800 samples as representatives, with 638 samples being in the evaluation set. For a given sample, the method searches the five closest neighbors in terms of cosine similarity between the query embeddings and classifies the sample by the majority class of the neighbors. The results, however, show as presented in Table \ref{tab:confusion_matrix_neares_neighbor_removal_strategy}, that compared with both classifiers, the nearest neighbors method leads to a worse false omission rate of 19.77\%.

\begin{table}[ht]
\centering
\renewcommand{\arraystretch}{1.3}
\begin{tabular}{c|cc}
 & \textbf{Predicted P} & \textbf{Predicted N} \\
\hline
\textbf{Actual P} &
\cellcolor{green!25} 116 &
\cellcolor{red!15} 85 \\
\textbf{Actual N} &
\cellcolor{red!15} 92 &
\cellcolor{green!25} 345 \\
\end{tabular}
\vspace{7pt}
\caption[Confusion Matrix for Five Nearest Neighbor Question Removal Classification Strategy]{Confusion matrix for five nearest neighbor question removal classification strategy (total samples = 638, FOR = 0.1977).}
\label{tab:confusion_matrix_neares_neighbor_removal_strategy}
\end{table}

\noindent \textbf{Final Query Cleaning Method:} Although the classifiers logistic regression and XGBoost have a slightly lower false omission rate compared to the combination method of LLM classifier and query LLM rating, we decide not to use the former methods. The main cause for this is the relatively small query removal dataset, which could possibly mean higher variance for the classifiers for unknown query types. Moreover, the difference between the two approaches is quite small, and the LLM approach keeps in the evaluation set 69.05\% of the queries and the classifiers 62.38\% for logistic regression and XGBoost 55.33\%.

\subsection{Query Dataset Statistics}
In this subsection, we present additional statistics for our query dataset. Table \ref{fig:domain_and_question_type_distribution} shows the query distribution by evidence structure type and domain type. Across all query evidence structures, the Financial Reports domain contains the most queries with a total of 79,791 queries. The evidence structure type distribution is similar between the different domain types with single-page queries being the most common ones followed by Cross-Doc queries. Since there are for less represented domains, such as economics, only few queries, we recommend for further evaluations on our benchmark to treat results grouped by evidence structure type and domain type with caution.

\newcommand{\rot}[1]{\rotatebox[origin=c]{90}{\parbox{2cm}{\centering #1}}}

\begin{table}[htbp]
\centering
\scriptsize
\begin{tabular}{l|rrrrrr|r}
\toprule
 & \rot{Page-Single} & \rot{Page-Multi} & \rot{Doc-Adjacent} & \rot{Doc-Near} & \rot{Doc-Far} & \rot{Cross-Doc} & \rot{Total} \\
\midrule
Financial Reports & 31,735 & 29,792 & 2,773 & 586 & 1,234 & 13,671 & 79,791 \\
Medical Reports & 7,835 & 17,006 & 1,902 & 523 & 416 & 6,324 & 34,006 \\
Sports & 5,891 & 10,017 & 1,067 & 436 & 69 & 7,026 & 24,506 \\
Astrophysics & 4,359 & 8,420 & 941 & 205 & 167 & 7,829 & 21,921 \\
Science & 6,006 & 7,080 & 632 & 212 & 41 & 2,566 & 16,537 \\
Geography & 6,064 & 6,957 & 442 & 206 & 85 & 2,267 & 16,021 \\
Condensed Matter Physics & 3,669 & 6,462 & 632 & 140 & 79 & 622 & 11,604 \\
Artificial Intelligence & 2,975 & 5,950 & 554 & 162 & 114 & 324 & 10,079 \\
Bioinformatics & 1,892 & 4,184 & 487 & 135 & 51 & 1,166 & 7,915 \\
History & 2,973 & 3,399 & 349 & 87 & 24 & 704 & 7,536 \\
Entertainment & 2,083 & 3,029 & 340 & 134 & 18 & 487 & 6,091 \\
Public Health Reports & 1,531 & 3,037 & 314 & 100 & 75 & 978 & 6,035 \\
Chemistry & 1,438 & 2,979 & 416 & 96 & 20 & 704 & 5,653 \\
Quantitative Biology & 1,298 & 2,768 & 215 & 59 & 51 & 84 & 4,475 \\
Technology & 1,257 & 1,420 & 159 & 32 & 10 & 1,258 & 4,136 \\
Environmental Reports & 725 & 1,690 & 132 & 71 & 36 & 214 & 2,868 \\
Electrical Engineering & 846 & 1,713 & 79 & 39 & 24 & 74 & 2,775 \\
Politics & 1,161 & 1,051 & 123 & 43 & 3 & 165 & 2,546 \\
Arts & 778 & 1,164 & 152 & 21 & 2 & 96 & 2,213 \\
General Computer Science & 659 & 1,248 & 101 & 24 & 18 & 54 & 2,104 \\
Surgical Case Reports & 707 & 724 & 109 & 75 & 15 & 357 & 1,987 \\
Statistics & 221 & 429 & 34 & 5 & 8 & 14 & 711 \\
Economics & 93 & 105 & 19 & 1 & 3 & 29 & 250 \\
\midrule
Total & 86,196 & 120,624 & 11,972 & 3,392 & 2,563 & 47,013 & 271,760 \\
\bottomrule
\end{tabular}
\caption{Distribution of dataset questions by question type and domain type.}
\label{fig:domain_and_question_type_distribution}
\end{table}

Figure \ref{fig:question_evidence_type_co_occurrence} illustrates the distribution of query evidence types and their co-occurrence. The most occurring evidence is pure text with 90,248 queries, followed by table row evidence with 40,955 queries, and the third most occurring combination is image+text with 28,932 queries. In general, this Figure \ref{fig:question_evidence_type_co_occurrence} shows that our benchmark queries' evidence types are well balanced across text, images and tabular data.

\definecolor{setC0}{HTML}{1F77B4}
\definecolor{setC1}{HTML}{FF7F0E}
\definecolor{setC2}{HTML}{2CA02C}
\definecolor{setC3}{HTML}{D62728}
\definecolor{setC4}{HTML}{9467BD}
\definecolor{setC5}{HTML}{8C564B}
\definecolor{setC6}{HTML}{E377C2}
\definecolor{setC7}{HTML}{7F7F7F}

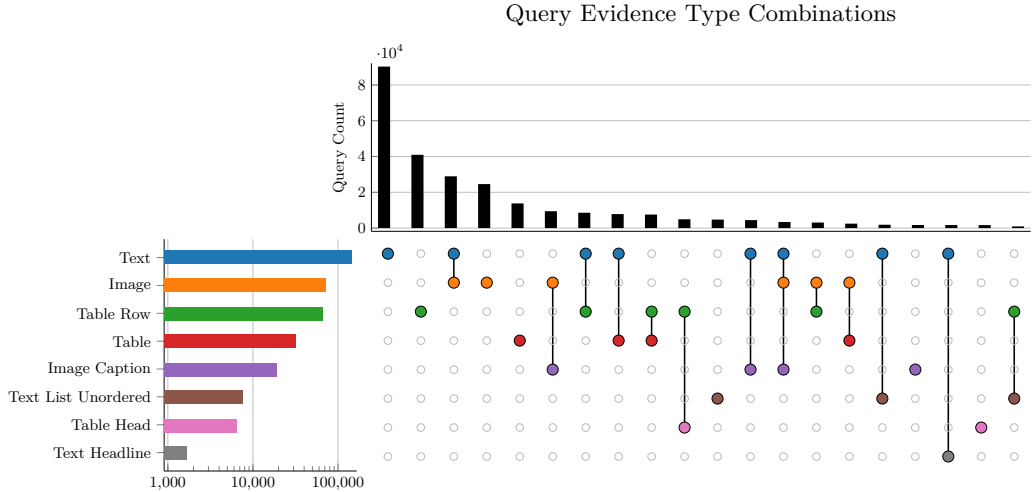
\begin{figure}[htbp]
    \centering

\begin{tikzpicture}[scale=0.65]

\small
\begin{axis}[
  name=topbars,
  width=15.0cm, height=5.0cm,
  ymin=0,
  xtick=\empty,
  ylabel={Query Count},
  ymajorgrids,
  axis x line*=bottom,
  axis y line*=left,
  enlargelimits=0.02,
  bar width=7.0pt,
]
\addplot[ybar, fill=black, draw=none] coordinates {
  (1,90248) (2,40955) (3,28932) (4,24587) (5,13780) (6,9429) (7,8576) (8,7820) (9,7556) (10,4911) (11,4736) (12,4450) (13,3382) (14,3092) (15,2467) (16,1868) (17,1720) (18,1697) (19,1655) (20,934)
};
\end{axis}

\node[font=\small, anchor=south] at ($(topbars.north)+(0,6mm)$) {Query Evidence Type Combinations};

\begin{axis}[
  name=matrix,
  at={(topbars.below south west)},
  anchor=above north west,
  yshift=-2mm,
  width=15.0cm, height=6.3cm,
  xmin=0.5, xmax=20.5,
  ymin=0.5, ymax=8.5,
  xtick=\empty,
  ytick=\empty,
  axis lines=none,
  clip=false,
]
\foreach \x in {1,...,20} {
  \foreach \y in {1,...,8} {
    \addplot[only marks, mark=o, mark size=2.2pt, draw=gray!55, fill=white]
      coordinates {(\x,\y)};
  }
}

\addplot[only marks, mark=*, mark size=3.2pt, draw=black, fill=setC0] coordinates {(1,8)};
\addplot[only marks, mark=*, mark size=3.2pt, draw=black, fill=setC2] coordinates {(2,6)};
\addplot[black, line width=0.9pt] coordinates {(3,7) (3,8)};
\addplot[only marks, mark=*, mark size=3.2pt, draw=black, fill=setC1] coordinates {(3,7)};
\addplot[only marks, mark=*, mark size=3.2pt, draw=black, fill=setC0] coordinates {(3,8)};
\addplot[only marks, mark=*, mark size=3.2pt, draw=black, fill=setC1] coordinates {(4,7)};
\addplot[only marks, mark=*, mark size=3.2pt, draw=black, fill=setC3] coordinates {(5,5)};
\addplot[black, line width=0.9pt] coordinates {(6,4) (6,7)};
\addplot[only marks, mark=*, mark size=3.2pt, draw=black, fill=setC4] coordinates {(6,4)};
\addplot[only marks, mark=*, mark size=3.2pt, draw=black, fill=setC1] coordinates {(6,7)};
\addplot[black, line width=0.9pt] coordinates {(7,6) (7,8)};
\addplot[only marks, mark=*, mark size=3.2pt, draw=black, fill=setC2] coordinates {(7,6)};
\addplot[only marks, mark=*, mark size=3.2pt, draw=black, fill=setC0] coordinates {(7,8)};
\addplot[black, line width=0.9pt] coordinates {(8,5) (8,8)};
\addplot[only marks, mark=*, mark size=3.2pt, draw=black, fill=setC3] coordinates {(8,5)};
\addplot[only marks, mark=*, mark size=3.2pt, draw=black, fill=setC0] coordinates {(8,8)};
\addplot[black, line width=0.9pt] coordinates {(9,5) (9,6)};
\addplot[only marks, mark=*, mark size=3.2pt, draw=black, fill=setC3] coordinates {(9,5)};
\addplot[only marks, mark=*, mark size=3.2pt, draw=black, fill=setC2] coordinates {(9,6)};
\addplot[black, line width=0.9pt] coordinates {(10,2) (10,6)};
\addplot[only marks, mark=*, mark size=3.2pt, draw=black, fill=setC6] coordinates {(10,2)};
\addplot[only marks, mark=*, mark size=3.2pt, draw=black, fill=setC2] coordinates {(10,6)};
\addplot[only marks, mark=*, mark size=3.2pt, draw=black, fill=setC5] coordinates {(11,3)};
\addplot[black, line width=0.9pt] coordinates {(12,4) (12,8)};
\addplot[only marks, mark=*, mark size=3.2pt, draw=black, fill=setC4] coordinates {(12,4)};
\addplot[only marks, mark=*, mark size=3.2pt, draw=black, fill=setC0] coordinates {(12,8)};
\addplot[black, line width=0.9pt] coordinates {(13,4) (13,8)};
\addplot[only marks, mark=*, mark size=3.2pt, draw=black, fill=setC4] coordinates {(13,4)};
\addplot[only marks, mark=*, mark size=3.2pt, draw=black, fill=setC1] coordinates {(13,7)};
\addplot[only marks, mark=*, mark size=3.2pt, draw=black, fill=setC0] coordinates {(13,8)};
\addplot[black, line width=0.9pt] coordinates {(14,6) (14,7)};
\addplot[only marks, mark=*, mark size=3.2pt, draw=black, fill=setC2] coordinates {(14,6)};
\addplot[only marks, mark=*, mark size=3.2pt, draw=black, fill=setC1] coordinates {(14,7)};
\addplot[black, line width=0.9pt] coordinates {(15,5) (15,7)};
\addplot[only marks, mark=*, mark size=3.2pt, draw=black, fill=setC3] coordinates {(15,5)};
\addplot[only marks, mark=*, mark size=3.2pt, draw=black, fill=setC1] coordinates {(15,7)};
\addplot[black, line width=0.9pt] coordinates {(16,3) (16,8)};
\addplot[only marks, mark=*, mark size=3.2pt, draw=black, fill=setC5] coordinates {(16,3)};
\addplot[only marks, mark=*, mark size=3.2pt, draw=black, fill=setC0] coordinates {(16,8)};
\addplot[only marks, mark=*, mark size=3.2pt, draw=black, fill=setC4] coordinates {(17,4)};
\addplot[black, line width=0.9pt] coordinates {(18,1) (18,8)};
\addplot[only marks, mark=*, mark size=3.2pt, draw=black, fill=setC7] coordinates {(18,1)};
\addplot[only marks, mark=*, mark size=3.2pt, draw=black, fill=setC0] coordinates {(18,8)};
\addplot[only marks, mark=*, mark size=3.2pt, draw=black, fill=setC6] coordinates {(19,2)};
\addplot[black, line width=0.9pt] coordinates {(20,3) (20,6)};
\addplot[only marks, mark=*, mark size=3.2pt, draw=black, fill=setC5] coordinates {(20,3)};
\addplot[only marks, mark=*, mark size=3.2pt, draw=black, fill=setC2] coordinates {(20,6)};
\end{axis}

\begin{axis}[
  name=leftbars,
  at={(matrix.west)},
  anchor=east,
  xshift=-3mm,
  width=5.5cm, height=6.3cm,
    xbar,
  xmode=log,
  xmin=1000,
  xmax=150000,
  ymin=0.5, ymax=8.5,
  ytick={1,...,8},
  yticklabels={Text Headline, Table Head, Text List Unordered, Image Caption, Table, Table Row, Image, Text},
  yticklabel style={align=right},
  axis x line*=bottom,
  axis y line*=left,
  xmajorgrids,
  enlargelimits=0.02,
  bar width=8.0pt,
    log ticks with fixed point,
    xticklabel style={/pgf/number format/fixed},
]
\addplot[xbar, fill=setC0, draw=none] coordinates {(146973,10.2)};
\addplot[xbar, fill=setC1, draw=none] coordinates {(71889,8.55)};
\addplot[xbar, fill=setC2, draw=none] coordinates {(66024,6.9)};
\addplot[xbar, fill=setC3, draw=none] coordinates {(31623,5.3)};
\addplot[xbar, fill=setC4, draw=none] coordinates {(18981,3.65)};
\addplot[xbar, fill=setC5, draw=none] coordinates {(7538,2.05)};
\addplot[xbar, fill=setC6, draw=none] coordinates {(6566,0.4)};
\addplot[xbar, fill=setC7, draw=none] coordinates {(1697,-1.2)};
\end{axis}

\end{tikzpicture}
    \caption{Co-occurrence of query evidence types and evidence type distribution. The most common occurring combinations are text, table row, and image with text. }
    \label{fig:question_evidence_type_co_occurrence}
\end{figure}

\subsection{Annotation Validation and Adjudication}
\label{app:annotation_quality}

Table~\ref{tab:annotation_validation_24_cells} reports the human audit over all 24 source-by-evidence-structure combinations. There are 50 Cross-Doc entries and 30 entries from every other structure per source. In total, 746/800 entries are fully correct (93.25\%), while 54/800 contain an observed issue (6.75\%). The issue counts are 13 formulation issues, 20 incorrect evidence sets, 17 sets containing an unnecessary evidence page, and four non-presence queries without directly retrievable positive evidence.

These rates describe the audited entries before corrections. The reported 95\% Wilson interval (91.3--94.8\%) is a nominal interval for the audited-entry proportion rather than a population-wide or post-correction accuracy estimate. Cross-Doc correctness is 177/200 (88.5\%), with the lowest source-specific rate observed for PubMed Cross-Doc (40/50, 80.0\%). The Cross-Doc audit contains 200 entries representing 199 unique queries; all counts in Table~\ref{tab:annotation_validation_24_cells} refer to audit entries. This retained-annotation audit is distinct from the 1,438-example query-cleaning evaluation in Appendix~\ref{subsec:appendix_question_cleaning_evaluation_dataset}.

\begin{table}[t]
\centering
\small
\setlength{\tabcolsep}{4pt}
\renewcommand{\arraystretch}{0.95}
\begin{tabular}{@{}llrrrrr@{}}
\toprule
Source & Evidence structure & Correct/Total & F & E & U & N \\
\midrule
\multirow{6}{*}{Wikipedia} & Page-Single & 29/30 & 1 & 0 & 0 & 0 \\
 & Page-Multi & 30/30 & 0 & 0 & 0 & 0 \\
 & Doc-Adjacent & 26/30 & 3 & 1 & 0 & 0 \\
 & Doc-Near & 29/30 & 1 & 0 & 0 & 0 \\
 & Doc-Far & 28/30 & 0 & 0 & 2 & 0 \\
 & Cross-Doc & 45/50 & 4 & 0 & 1 & 0 \\
\midrule
\multirow{6}{*}{arXiv} & Page-Single & 29/30 & 1 & 0 & 0 & 0 \\
 & Page-Multi & 30/30 & 0 & 0 & 0 & 0 \\
 & Doc-Adjacent & 28/30 & 0 & 1 & 1 & 0 \\
 & Doc-Near & 27/30 & 0 & 1 & 2 & 0 \\
 & Doc-Far & 29/30 & 1 & 0 & 0 & 0 \\
 & Cross-Doc & 48/50 & 0 & 2 & 0 & 0 \\
\midrule
\multirow{6}{*}{PubMed} & Page-Single & 29/30 & 1 & 0 & 0 & 0 \\
 & Page-Multi & 30/30 & 0 & 0 & 0 & 0 \\
 & Doc-Adjacent & 28/30 & 0 & 0 & 1 & 1 \\
 & Doc-Near & 28/30 & 0 & 0 & 2 & 0 \\
 & Doc-Far & 30/30 & 0 & 0 & 0 & 0 \\
 & Cross-Doc & 40/50 & 0 & 10 & 0 & 0 \\
\midrule
\multirow{6}{*}{SEC} & Page-Single & 30/30 & 0 & 0 & 0 & 0 \\
 & Page-Multi & 29/30 & 1 & 0 & 0 & 0 \\
 & Doc-Adjacent & 26/30 & 0 & 1 & 3 & 0 \\
 & Doc-Near & 29/30 & 0 & 0 & 1 & 0 \\
 & Doc-Far & 25/30 & 0 & 1 & 4 & 0 \\
 & Cross-Doc & 44/50 & 0 & 3 & 0 & 3 \\
\midrule
Total & & 746/800 & 13 & 20 & 17 & 4 \\
\bottomrule
\end{tabular}
\caption{\textbf{Stratified annotation audit}. F: incorrectly formulated or insufficiently identifying query; E: incorrect evidence; U: unnecessary evidence page; N: non-presence query without directly retrievable positive evidence.}
\label{tab:annotation_validation_24_cells}
\end{table}

Incorrect or unnecessary annotated pages can make RecallAll and nDCGAll mark a retrieval as incomplete even when answer-relevant evidence was retrieved. The lower Cross-Doc quality therefore remains relevant to interpreting its retrieval scores. A separate census covers all 47,013 Cross-Doc queries and 100,504 evidence elements. It finds 39,600 fully consistent queries (84.23\%), 5,860 queries with an element spanning rendered pages (12.46\%), 839 with an out-of-range stored bounding box (1.78\%), and 714 with stored/configured pages differing from the bounding-box-derived page (1.52\%). No mappings, element IDs, page images, configurations, or evidence documents are missing in these checks. 

After applying the accepted corrections, we re-evaluated the twelve frozen retrieval runs. The complete method order and best method were unchanged for all twelve metric--cutoff combinations under Micro, structure-balanced Macro, source-averaged Macro, and domain-averaged Macro. This held within the nine-method full-corpus comparison and the separately evaluated three-method.

\section{Evaluation Result Details}
Below, we present more detailed evaluation results on our benchmark. We show especially results grouped by different query categories, such as by query evidence structure type and data source type, or by evidence type. Additionally, we conduct further experiments to measure retrieval performance on our benchmark. We have experimented with the 40GB A100 and 94GB H100-equipped computing nodes. Each retrieval experiment per retrieval model has consumed approximately 100 hours of compute time on A100s and H100s.

\subsection{Overall Results}
The overall benchmark results refer to the evaluation of our whole benchmark across all query types, data source types, and evidence types. Table~\ref{tab:benchmark_results_overall} presents the complete results for the visual, OCR-based, native-text, and hybrid retrieval methods. The abbreviations Ctx, Rerank, and Hierarchical refer to document-context scoring, reranking, and hierarchical retrieval, respectively. B+Ctx combines the ranked lists from base retrieval and document-context scoring. Q-Split 1--3 denote the three query-splitting variants, with Q-Split 3 corresponding to Q-Split in Table~\ref{tab:overall_metric_summary}. ColNomic (Ensemble) combines the reranked, context, and base retrieval lists. Within each aggregation protocol and metric column, the highest, second-highest, and third-highest scores are shown in bold blue, bold green, and bold black, respectively.

ColNomic (Ensemble) is the best-performing configuration under both aggregation protocols, achieving nDCGAll@10 of 69.28 and RecallAll@10 of 80.64 under Macro (75.64 and 86.23 under Micro, respectively). Under Macro, ColNomic (Ctx) obtains higher nDCGAll@10 and RecallAll@10 than ColNomic (Rerank), while under Micro, ColNomic (Rerank) obtains higher scores for both metrics.

\begin{table*}[t]
\centering
\providecolor{bestcell}{RGB}{0,0,255}
\providecolor{secondbestcell}{RGB}{33,145,140}
\providecolor{thirdbestcell}{RGB}{0,0,0}
\resizebox{\textwidth}{!}{
\setlength{\tabcolsep}{2pt}
\renewcommand{\arraystretch}{0.94}
\begin{tabular}{@{}llrrr rrr rrr rrr@{}}
\toprule
\multirow{2}{*}{Model} & \multirow{2}{*}{Aggregate}
& \multicolumn{3}{c}{nDCG@k} & \multicolumn{3}{c}{Recall@k}
& \multicolumn{3}{c}{nDCGAll@k} & \multicolumn{3}{c}{RecallAll@k} \\
\cmidrule(lr){3-5}\cmidrule(lr){6-8}\cmidrule(lr){9-11}\cmidrule(l){12-14}
& & @3 & @5 & @10 & @3 & @5 & @10 & @3 & @5 & @10 & @3 & @5 & @10 \\
\midrule
\multirow{2}{*}{ColSmol 500} & Macro & 49.43 & 53.36 & 56.26 & 52.89 & 61.04 & 68.62 & 35.30 & 42.11 & 47.33 & 39.01 & 49.10 & 58.15 \\
 & Micro & 54.98 & 57.76 & 59.95 & 61.35 & 67.70 & 73.97 & 50.63 & 53.87 & 56.58 & 57.03 & 63.45 & 69.81 \\
\multirow{2}{*}{ColModernVBERT} & Macro & 47.33 & 50.91 & 53.72 & 50.48 & 57.93 & 65.30 & 33.28 & 39.31 & 44.29 & 36.73 & 45.70 & 54.35 \\
 & Micro & 51.69 & 54.39 & 56.55 & 57.73 & 63.92 & 70.11 & 47.45 & 50.57 & 53.20 & 53.51 & 59.73 & 65.96 \\
\multirow{2}{*}{GME Qwen2-VL} & Macro & 55.27 & 59.51 & 62.72 & 59.35 & 68.10 & 76.42 & 41.08 & 48.82 & 55.10 & 45.33 & 56.70 & 67.34 \\
 & Micro & 60.18 & 63.16 & 65.44 & 67.20 & 73.94 & 80.35 & 55.34 & 59.01 & 62.11 & 62.37 & 69.38 & 76.22 \\
\midrule
\multirow{2}{*}{Jina V4 SV} & Macro & 48.31 & 52.61 & 55.80 & 52.40 & 61.33 & 69.69 & 34.70 & 42.09 & 48.03 & 38.80 & 49.86 & 60.24 \\
 & Micro & 52.55 & 55.75 & 58.24 & 59.58 & 66.87 & 74.01 & 48.17 & 51.93 & 55.10 & 55.16 & 62.60 & 70.00 \\
\multirow{2}{*}{Jina V4 MV} & Macro & 60.92 & 64.99 & 67.79 & 64.43 & 72.78 & 79.99 & 45.69 & 53.56 & 59.59 & 49.61 & 60.92 & 70.77 \\
 & Micro & 67.24 & 69.83 & 71.74 & 73.36 & 79.17 & 84.47 & 61.80 & 65.15 & 67.98 & 68.01 & 74.16 & 80.03 \\
\midrule
\multirow{2}{*}{ColQwen3} & Macro & 65.54 & 69.09 & 71.36 & 68.15 & 75.35 & 81.13 & 50.39 & 57.86 & 63.20 & 53.74 & 64.11 & 72.42 \\
 & Micro & 72.54 & \textbf{\textcolor{thirdbestcell}{74.53}} & \textbf{\textcolor{thirdbestcell}{75.89}} & 77.44 & 81.80 & 85.49 & 67.13 & 69.92 & \textbf{\textcolor{thirdbestcell}{72.15}} & \textbf{\textcolor{thirdbestcell}{72.21}} & 77.00 & 81.24 \\
\midrule
\multirow{2}{*}{ColPali v1.3} & Macro & 42.03 & 45.47 & 48.26 & 45.04 & 52.18 & 59.52 & 28.93 & 34.55 & 39.41 & 32.14 & 40.58 & 49.13 \\
 & Micro & 46.10 & 48.74 & 50.92 & 51.76 & 57.80 & 64.08 & 42.21 & 45.20 & 47.85 & 47.87 & 53.90 & 60.26 \\
\multirow{2}{*}{ColPali v1.3 (Ctx)} & Macro & 47.99 & 53.19 & 57.46 & 52.21 & 63.01 & 74.20 & 33.51 & 42.55 & 50.87 & 37.63 & 51.13 & 65.69 \\
 & Micro & 50.46 & 54.12 & 57.22 & 57.45 & 65.84 & 74.76 & 45.48 & 49.71 & 53.48 & 52.47 & 60.92 & 69.90 \\
\multirow{2}{*}{ColPali v1.3 (B+Ctx)} & Macro & 47.28 & 52.03 & 56.15 & 51.28 & 61.14 & 71.98 & 32.67 & 40.65 & 48.69 & 36.66 & 48.59 & 62.55 \\
 & Micro & 50.29 & 53.66 & 56.69 & 57.22 & 64.96 & 73.68 & 45.51 & 49.39 & 53.14 & 52.45 & 60.22 & 69.12 \\
\multirow{2}{*}{ColPali v1.3 (Rerank)} & Macro & 57.25 & 59.73 & 60.92 & 58.96 & 63.94 & 66.94 & 43.30 & 48.42 & 51.12 & 45.88 & 52.93 & 57.16 \\
 & Micro & 62.97 & 64.26 & 64.95 & 66.38 & 69.19 & 71.03 & 58.31 & 60.18 & 61.33 & 61.93 & 65.05 & 67.17 \\
\midrule
\multirow{2}{*}{ColQwen2.5} & Macro & 58.68 & 62.56 & 65.15 & 62.08 & 70.02 & 76.72 & 44.15 & 51.48 & 56.72 & 47.93 & 58.51 & 67.23 \\
 & Micro & 64.94 & 67.41 & 69.21 & 70.84 & 76.37 & 81.40 & 59.84 & 62.97 & 65.51 & 65.82 & 71.63 & 77.04 \\
\multirow{2}{*}{ColQwen2.5 (Ctx)} & Macro & 61.22 & 65.91 & 69.30 & 65.20 & 74.82 & 83.56 & 46.17 & 55.36 & 62.80 & 50.39 & 63.52 & 75.78 \\
 & Micro & 66.02 & 68.97 & 71.15 & 72.52 & 79.16 & 85.28 & 60.13 & 63.84 & 66.85 & 66.77 & 73.67 & 80.10 \\
\multirow{2}{*}{ColQwen2.5 (B+Ctx)} & Macro & 61.56 & 66.11 & 69.40 & 65.81 & 75.15 & 83.63 & 46.51 & 55.48 & 62.74 & 50.95 & 63.80 & 75.68 \\
 & Micro & 66.66 & 69.56 & 71.72 & 73.58 & 80.10 & 86.11 & 60.97 & 64.67 & 67.81 & 67.96 & 74.83 & 81.41 \\
\multirow{2}{*}{ColQwen2.5 (Q-Split)} & Macro & 58.13 & 62.29 & 65.15 & 62.30 & 70.81 & 78.20 & 44.04 & 51.69 & 57.48 & 48.21 & 59.39 & 69.19 \\
 & Micro & 63.99 & 66.72 & 68.74 & 70.76 & 76.87 & 82.52 & 59.25 & 62.73 & 65.64 & 65.95 & 72.44 & 78.63 \\
\multirow{2}{*}{ColQwen2.5 (Rerank)} & Macro & \textbf{\textcolor{thirdbestcell}{65.82}} & 69.22 & 71.15 & 68.55 & 75.45 & 80.31 & \textbf{\textcolor{thirdbestcell}{51.33}} & 58.45 & 62.88 & \textbf{\textcolor{thirdbestcell}{54.69}} & 64.60 & 71.62 \\
 & Micro & \textbf{\textcolor{thirdbestcell}{72.61}} & 74.49 & 75.65 & 77.26 & 81.38 & 84.50 & \textbf{\textcolor{thirdbestcell}{67.33}} & \textbf{\textcolor{thirdbestcell}{70.02}} & 71.94 & 72.14 & 76.73 & 80.36 \\
\midrule
\multirow{2}{*}{ColNomic} & Macro & 63.85 & 67.61 & 70.05 & 66.84 & 74.53 & 80.78 & 48.73 & 56.38 & 61.82 & 52.36 & 63.06 & 71.76 \\
 & Micro & 70.46 & 72.75 & 74.35 & 75.95 & 81.05 & 85.46 & 65.06 & 68.17 & 70.68 & 70.71 & 76.24 & 81.24 \\
\multirow{2}{*}{ColNomic (Ctx)} & Macro & 64.85 & 69.37 & 72.57 & 68.26 & \textbf{\textcolor{thirdbestcell}{77.51}} & 85.72 & 49.33 & 58.64 & \textbf{\textcolor{thirdbestcell}{66.03}} & 53.28 & 66.31 & 78.19 \\
 & Micro & 69.75 & 72.51 & 74.48 & 75.70 & 81.88 & 87.38 & 63.50 & 67.15 & 70.07 & 69.69 & 76.28 & 82.26 \\
\multirow{2}{*}{ColNomic (B+Ctx)} & Macro & 65.69 & \textbf{\textcolor{thirdbestcell}{70.25}} & \textbf{\textcolor{secondbestcell}{73.34}} & \textbf{\textcolor{thirdbestcell}{69.63}} & \textbf{\textcolor{secondbestcell}{78.96}} & \textbf{\textcolor{secondbestcell}{86.89}} & 50.31 & \textbf{\textcolor{thirdbestcell}{59.77}} & \textbf{\textcolor{secondbestcell}{67.06}} & 54.63 & \textbf{\textcolor{secondbestcell}{67.91}} & \textbf{\textcolor{secondbestcell}{79.60}} \\
 & Micro & 71.02 & 73.77 & 75.71 & \textbf{\textcolor{thirdbestcell}{77.75}} & \textbf{\textcolor{secondbestcell}{83.88}} & \textbf{\textcolor{secondbestcell}{89.24}} & 65.11 & 68.87 & 71.96 & 71.95 & \textbf{\textcolor{secondbestcell}{78.66}} & \textbf{\textcolor{secondbestcell}{84.81}} \\
\multirow{2}{*}{ColNomic (Rerank)} & Macro & \textbf{\textcolor{secondbestcell}{66.82}} & \textbf{\textcolor{secondbestcell}{70.50}} & \textbf{\textcolor{thirdbestcell}{72.76}} & \textbf{\textcolor{secondbestcell}{69.77}} & 77.24 & 83.00 & \textbf{\textcolor{secondbestcell}{52.25}} & \textbf{\textcolor{secondbestcell}{59.84}} & 65.00 & \textbf{\textcolor{secondbestcell}{55.77}} & \textbf{\textcolor{thirdbestcell}{66.37}} & 74.62 \\
 & Micro & \textbf{\textcolor{secondbestcell}{73.77}} & \textbf{\textcolor{secondbestcell}{75.87}} & \textbf{\textcolor{secondbestcell}{77.29}} & \textbf{\textcolor{secondbestcell}{78.72}} & \textbf{\textcolor{thirdbestcell}{83.33}} & 87.15 & \textbf{\textcolor{secondbestcell}{68.44}} & \textbf{\textcolor{secondbestcell}{71.42}} & \textbf{\textcolor{secondbestcell}{73.76}} & \textbf{\textcolor{secondbestcell}{73.52}} & \textbf{\textcolor{thirdbestcell}{78.63}} & 83.11 \\
\multirow{2}{*}{ColNomic (Ensemble)} & Macro & \textbf{\textcolor{bestcell}{68.42}} & \textbf{\textcolor{bestcell}{72.68}} & \textbf{\textcolor{bestcell}{75.45}} & \textbf{\textcolor{bestcell}{72.03}} & \textbf{\textcolor{bestcell}{80.67}} & \textbf{\textcolor{bestcell}{87.73}} & \textbf{\textcolor{bestcell}{53.58}} & \textbf{\textcolor{bestcell}{62.60}} & \textbf{\textcolor{bestcell}{69.28}} & \textbf{\textcolor{bestcell}{57.43}} & \textbf{\textcolor{bestcell}{70.04}} & \textbf{\textcolor{bestcell}{80.64}} \\
 & Micro & \textbf{\textcolor{bestcell}{75.00}} & \textbf{\textcolor{bestcell}{77.39}} & \textbf{\textcolor{bestcell}{79.03}} & \textbf{\textcolor{bestcell}{80.59}} & \textbf{\textcolor{bestcell}{85.85}} & \textbf{\textcolor{bestcell}{90.30}} & \textbf{\textcolor{bestcell}{69.43}} & \textbf{\textcolor{bestcell}{72.89}} & \textbf{\textcolor{bestcell}{75.64}} & \textbf{\textcolor{bestcell}{75.02}} & \textbf{\textcolor{bestcell}{80.95}} & \textbf{\textcolor{bestcell}{86.23}} \\
\midrule
\multirow{2}{*}{OCR-BM25} & Macro & 51.30 & 55.40 & 58.47 & 54.99 & 63.50 & 71.53 & 37.56 & 44.59 & 50.24 & 41.45 & 51.91 & 61.69 \\
 & Micro & 56.06 & 59.16 & 61.53 & 62.70 & 69.76 & 76.48 & 51.55 & 55.20 & 58.26 & 58.18 & 65.36 & 72.34 \\
\multirow{2}{*}{OCR-dense (BGE-M3)} & Macro & 22.90 & 24.97 & 26.59 & 24.58 & 28.90 & 33.16 & 14.14 & 16.92 & 19.35 & 15.92 & 20.26 & 24.70 \\
 & Micro & 23.37 & 24.95 & 26.24 & 26.66 & 30.28 & 34.00 & 20.81 & 22.45 & 23.86 & 24.10 & 27.55 & 31.12 \\
\multirow{2}{*}{DOM-BM25} & Macro & 47.81 & 51.49 & 54.34 & 51.12 & 58.75 & 66.19 & 35.24 & 41.67 & 46.96 & 38.82 & 48.37 & 57.44 \\
 & Micro & 51.11 & 53.86 & 56.00 & 57.07 & 63.34 & 69.41 & 47.04 & 50.31 & 53.07 & 53.01 & 59.44 & 65.72 \\
\midrule
\multirow{2}{*}{Ctx + OCR-BM25 (RRF)} & Macro & 62.88 & 67.70 & 71.23 & 67.11 & 77.01 & \textbf{\textcolor{thirdbestcell}{86.10}} & 47.87 & 57.14 & 65.13 & 52.30 & 65.60 & \textbf{\textcolor{thirdbestcell}{78.67}} \\
 & Micro & 67.74 & 70.97 & 73.26 & 75.00 & 82.26 & \textbf{\textcolor{thirdbestcell}{88.62}} & 62.27 & 66.34 & 69.75 & 69.53 & 77.13 & \textbf{\textcolor{thirdbestcell}{84.24}} \\
\multirow{2}{*}{Rerank + OCR-BM25 (RRF)} & Macro & 61.38 & 65.84 & 69.10 & 65.41 & 74.60 & 82.96 & 46.78 & 54.95 & 61.86 & 51.07 & 62.90 & 74.39 \\
 & Micro & 67.30 & 70.39 & 72.61 & 74.29 & 81.25 & 87.37 & 62.16 & 66.07 & 69.34 & 69.16 & 76.47 & 83.30 \\
\bottomrule
\end{tabular}
}
\caption[Overall Retrieval Benchmark Results]{Overall retrieval benchmark results with the nDCG@k~\cite{jarvelin2002}, Recall@k, nDCGAll@k~\cite{jarvelin2002}, and RecallAll@k as evaluation metrics. The best models are highlighted with \textbf{\textcolor{bestcell}{blue}}, the second best with \textbf{\textcolor{secondbestcell}{green}}, and the third best with \textbf{\textcolor{thirdbestcell}{bold}} text. The numbers below the evaluation metrics represent the values of parameter k.}
\label{tab:benchmark_results_overall}
\end{table*}

Among the base visual retrievers, ColQwen3~\cite{huang2025tomoroqwen3} achieves the highest
nDCGAll@10 under both Macro and Micro, followed by ColNomic~\cite{nomicembedmultimodal2025}.
The Jina V4~\cite{guenther2025jinaembeddingsv4universalembeddingsmultimodal} multi-vector model outperforms its single-vector counterpart by 11.6 and 10.5 percentage points in Macro nDCGAll@10 and RecallAll@10, respectively (12.9 and 10.0 under Micro). For ColQwen2.5, Q-Split achieves higher RecallAll for all three values of $k$. Its RecallAll@10 scores are 69.19 under Macro and 78.63 under Micro, compared with 67.23 and 77.04 for the base model, respectively.

OCR-BM25 achieves the highest Macro nDCGAll@10 and RecallAll@10 among the evaluated text baselines, with scores of 50.24 and 61.69, respectively. The corresponding scores are 19.35 and 24.70 for OCR-dense (BGE-M3)~\cite{bge-m3}, and 46.96 and 57.44 for DOM-BM25.

RRF fusion of OCR-BM25 with ColNomic (Ctx) increases RecallAll@10 from 78.19 to 78.67 under Macro and from 82.26 to 84.24 under Micro. The corresponding nDCGAll@10 scores decrease from 66.03 to 65.13 and from 70.07 to 69.75, respectively. For ColNomic (Rerank), fusion with OCR-BM25 decreases Macro RecallAll@10 from 74.62 to 74.39 and increases Micro RecallAll@10 from 83.11 to 83.30, with lower nDCGAll@10 under both aggregation protocols.

\subsection{Source- and Domain-Averaged Macro Results}
\label{app:retrieval_full_results}

We report source-averaged and domain-averaged Macro for nine
full-corpus configurations: the ColNomic base, Ctx, Rerank, and
Ensemble settings; OCR-BM25, OCR-dense (BGE-M3), and DOM-BM25;
and the two ColNomic--OCR-BM25 RRF hybrids. These aggregations
assign equal weight to data sources or domains, while retaining
equal weight across evidence structures within each group.
Structure-balanced Macro remains the primary aggregation protocol.

For a source or domain $g$, let $Q_{g,e}$ denote the set of queries
with evidence structure $e$, and let $m(q)$ denote the query-level
metric. We compute the structure-balanced mean within each group as
\begin{equation}
 M_g(m)=\frac{1}{6}\sum_{e=1}^{6}
 \frac{1}{|Q_{g,e}|}\sum_{q\in Q_{g,e}}m(q),
\end{equation}
where all four sources and all 23 domains contain queries from
each of the six evidence structures. The two aggregate scores are
then defined as
\begin{align}
 \mathrm{SourceMacro}(m)
 &=\frac{100}{4}\sum_{s=1}^{4}M_s(m),\\
 \mathrm{DomainMacro}(m)
 &=\frac{100}{23}\sum_{d=1}^{23}M_d(m).
\end{align}

Tables~\ref{tab:source_averaged_macro} and
\ref{tab:domain_averaged_macro} report nDCG, Recall, nDCGAll,
and RecallAll for $k\in\{3,5,10\}$. ColNomic (Ensemble) achieves
the highest scores among the nine configurations for all metrics
and values of $k$ under both aggregations. Its nDCGAll@10 and
RecallAll@10 are 69.78 and 80.85 under source-averaged Macro,
and 75.26 and 85.93 under domain-averaged Macro, respectively.
The relative performance of Ctx and Rerank differs between the
two aggregations: ColNomic (Ctx) obtains higher RecallAll@10
under source-averaged Macro, whereas
ColNomic (Rerank) obtains higher RecallAll@10 under
domain-averaged Macro.

\begin{table*}[t]
\centering
\scriptsize
\setlength{\tabcolsep}{2.0pt}
\renewcommand{\arraystretch}{0.95}
\begin{tabular}{@{}lrrr rrr rrr rrr@{}}
\toprule
\multirow{2}{*}{Model / Setting}
& \multicolumn{3}{c}{nDCG@k}
& \multicolumn{3}{c}{Recall@k}
& \multicolumn{3}{c}{nDCGAll@k}
& \multicolumn{3}{c}{RecallAll@k} \\
\cmidrule(lr){2-4}
\cmidrule(lr){5-7}
\cmidrule(lr){8-10}
\cmidrule(l){11-13}
& @3 & @5 & @10
& @3 & @5 & @10
& @3 & @5 & @10
& @3 & @5 & @10 \\
\midrule
ColNomic
& 64.68 & 68.47 & 70.89
& 67.65 & 75.36 & 81.58
& 49.33 & 57.13 & 62.63
& 52.99 & 63.85 & 72.56 \\

ColNomic (Ctx)
& 65.11 & 69.58 & 72.74
& 68.45 & 77.58 & 85.71
& 49.55 & 58.77 & 66.07
& 53.51 & 66.34 & 78.01 \\

ColNomic (Rerank)
& 67.61 & 71.33 & 73.59
& 70.44 & 77.97 & 83.72
& 52.82 & 60.59 & 65.82
& 56.31 & 67.09 & 75.39 \\

ColNomic (Ensemble)
& \textbf{69.02} & \textbf{73.22} & \textbf{75.97}
& \textbf{72.43} & \textbf{80.95} & \textbf{87.93}
& \textbf{54.10} & \textbf{63.08} & \textbf{69.78}
& \textbf{57.88} & \textbf{70.37} & \textbf{80.85} \\

OCR-BM25
& 52.04 & 56.18 & 59.29
& 55.69 & 64.28 & 72.39
& 38.00 & 45.14 & 50.95
& 41.92 & 52.49 & 62.45 \\

OCR-dense (BGE-M3)
& 23.39 & 25.54 & 27.22
& 25.14 & 29.61 & 34.06
& 14.50 & 17.33 & 19.84
& 16.36 & 20.79 & 25.40 \\

DOM-BM25
& 48.80 & 52.52 & 55.44
& 52.21 & 59.94 & 67.54
& 35.91 & 42.39 & 47.86
& 39.61 & 49.23 & 58.54 \\

Ctx + OCR-BM25 (RRF)
& 63.28 & 68.00 & 71.53
& 67.42 & 77.13 & 86.19
& 48.21 & 57.36 & 65.40
& 52.63 & 65.70 & 78.75 \\

Rerank + OCR-BM25 (RRF)
& 62.07 & 66.51 & 69.79
& 66.06 & 75.20 & 83.61
& 47.22 & 55.46 & 62.58
& 51.53 & 63.40 & 75.13 \\
\bottomrule
\end{tabular}
\caption{\textbf{Source-averaged Macro retrieval results} (\%).
Each score is the equal-weight mean of the four source-specific
six-structure Macro scores. Bold indicates the highest score
in each column.}
\label{tab:source_averaged_macro}
\end{table*}

\begin{table*}[t]
\centering
\scriptsize
\setlength{\tabcolsep}{2.0pt}
\renewcommand{\arraystretch}{0.95}
\begin{tabular}{@{}lrrr rrr rrr rrr@{}}
\toprule
\multirow{2}{*}{Model / Setting}
& \multicolumn{3}{c}{nDCG@k}
& \multicolumn{3}{c}{Recall@k}
& \multicolumn{3}{c}{nDCGAll@k}
& \multicolumn{3}{c}{RecallAll@k} \\
\cmidrule(lr){2-4}
\cmidrule(lr){5-7}
\cmidrule(lr){8-10}
\cmidrule(l){11-13}
& @3 & @5 & @10
& @3 & @5 & @10
& @3 & @5 & @10
& @3 & @5 & @10 \\
\midrule
ColNomic
& 70.06 & 74.15 & 76.58
& 73.02 & 81.32 & 87.51
& 54.93 & 63.96 & 69.97
& 58.58 & 70.98 & 80.31 \\

ColNomic (Ctx)
& 67.87 & 72.58 & 75.65
& 70.71 & 80.28 & 88.15
& 52.05 & 62.50 & 70.11
& 55.70 & 69.86 & 81.76 \\

ColNomic (Rerank)
& 72.22 & 76.21 & 78.55
& 74.95 & 83.00 & 88.91
& 57.74 & 66.77 & 72.47
& 61.15 & 73.40 & 82.34 \\

ColNomic (Ensemble)
& \textbf{72.83} & \textbf{77.18} & \textbf{79.82}
& \textbf{75.83} & \textbf{84.58} & \textbf{91.22}
& \textbf{58.28} & \textbf{68.50} & \textbf{75.26}
& \textbf{61.85} & \textbf{75.56} & \textbf{85.93} \\

OCR-BM25
& 58.06 & 62.23 & 65.51
& 61.86 & 70.47 & 78.91
& 43.78 & 51.68 & 58.38
& 47.87 & 59.16 & 70.42 \\

OCR-dense (BGE-M3)
& 24.09 & 26.25 & 28.04
& 25.76 & 30.25 & 34.94
& 15.07 & 18.09 & 20.82
& 16.89 & 21.59 & 26.58 \\

DOM-BM25
& 56.75 & 60.81 & 63.87
& 60.57 & 69.00 & 76.93
& 42.61 & 50.38 & 56.73
& 46.75 & 58.10 & 68.63 \\

Ctx + OCR-BM25 (RRF)
& 67.80 & 72.22 & 75.59
& 71.62 & 80.67 & 89.29
& 53.60 & 62.70 & 70.95
& 57.88 & 70.49 & 83.51 \\

Rerank + OCR-BM25 (RRF)
& 67.43 & 71.84 & 75.08
& 71.15 & 80.16 & 88.43
& 53.20 & 62.06 & 69.46
& 57.44 & 69.81 & 81.78 \\
\bottomrule
\end{tabular}
\caption{\textbf{Domain-averaged Macro retrieval results} (\%).
Each score is the equal-weight mean of the 23 domain-specific
six-structure Macro scores. Bold indicates the highest score
in each column.}
\label{tab:domain_averaged_macro}
\end{table*}

\subsection{Results by Base Model}

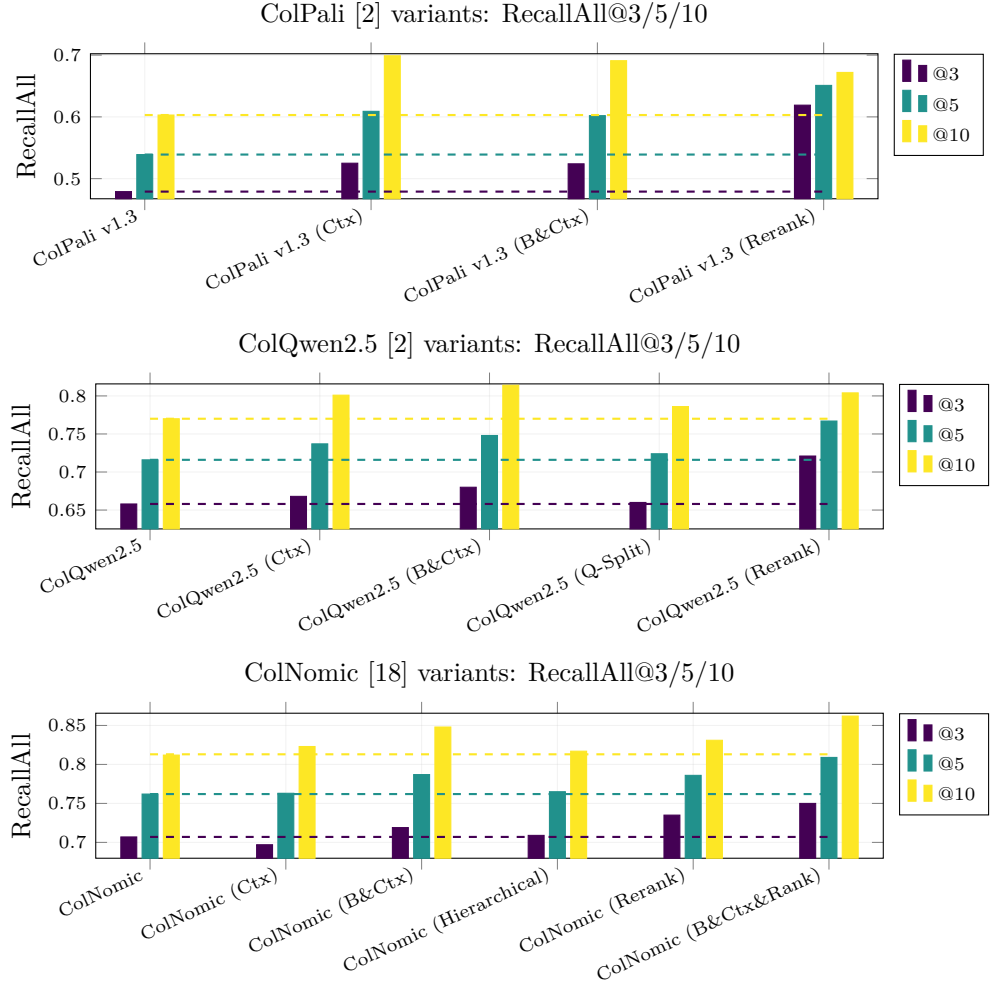
\begin{figure}[htbp]
    \centering
\begin{tikzpicture}
\begin{axis}[
  title={ColPali \cite{faysse2024colpali} variants: RecallAll@3/5/10},
  width=12cm,
  height=3.5cm,
  ybar,
  bar width=6pt,
  ymin=0.467410,
  ymax=0.702251,
  ylabel={RecallAll},
  symbolic x coords={ColPali v1.3,ColPali v1.3 (Ctx),ColPali v1.3 (B\&Ctx),ColPali v1.3 (Rerank)},
  xtick=data,
  xticklabel style={rotate=25, anchor=east, font=\scriptsize},
  yticklabel style={font=\scriptsize},
  enlarge x limits=0.08,
  clip=false,
  grid=both,
  major grid style={opacity=0.2},
  minor grid style={opacity=0.1},
  legend style={font=\scriptsize, at={(1.02,1)}, anchor=north west},
  legend cell align=left,
]
\addplot+[draw=viridisA, fill=viridisA] coordinates {
(ColPali v1.3,0.479)
(ColPali v1.3 (Ctx),0.525)
(ColPali v1.3 (B\&Ctx),0.524)
(ColPali v1.3 (Rerank),0.619)
};
\addplot+[draw=viridisB, fill=viridisB] coordinates {
(ColPali v1.3,0.539)
(ColPali v1.3 (Ctx),0.609)
(ColPali v1.3 (B\&Ctx),0.602)
(ColPali v1.3 (Rerank),0.651)
};
\addplot+[draw=viridisC, fill=viridisC] coordinates {
(ColPali v1.3,0.603)
(ColPali v1.3 (Ctx),0.699)
(ColPali v1.3 (B\&Ctx),0.691)
(ColPali v1.3 (Rerank),0.672)
};

\addplot+[forget plot, sharp plot, thick, dashed, draw=viridisA, fill=none]
    coordinates {(ColPali v1.3,0.479) (ColPali v1.3 (Rerank),0.479)};
\addplot+[forget plot, sharp plot, thick, dashed, draw=viridisB, fill=none]
    coordinates {(ColPali v1.3,0.539) (ColPali v1.3 (Rerank),0.539)};
\addplot+[forget plot, sharp plot, thick, dashed, draw=viridisC, fill=none]
    coordinates {(ColPali v1.3,0.603) (ColPali v1.3 (Rerank),0.603)};

\legend{@3,@5,@10}
\end{axis}
\end{tikzpicture}

\begin{tikzpicture}
\begin{axis}[
  title={ColQwen2.5 \cite{faysse2024colpali} variants: RecallAll@3/5/10},
  width=12cm,
  height=3.5cm,
  ybar,
  bar width=6pt,
  ymin=0.625222,
  ymax=0.815805,
  ylabel={RecallAll},
  symbolic x coords={ColQwen2.5,ColQwen2.5 (Ctx),ColQwen2.5 (B\&Ctx),ColQwen2.5 (Q-Split),ColQwen2.5 (Rerank)},
  xtick=data,
  xticklabel style={rotate=25, anchor=east, font=\scriptsize},
  yticklabel style={font=\scriptsize},
  enlarge x limits=0.08,
  clip=false,
  grid=both,
  major grid style={opacity=0.2},
  minor grid style={opacity=0.1},
  legend style={font=\scriptsize, at={(1.02,1)}, anchor=north west},
  legend cell align=left,
]
\addplot+[draw=viridisA, fill=viridisA] coordinates {
(ColQwen2.5,0.658)
(ColQwen2.5 (Ctx),0.668)
(ColQwen2.5 (B\&Ctx),0.68)
(ColQwen2.5 (Q-Split),0.66)
(ColQwen2.5 (Rerank),0.721)
};
\addplot+[draw=viridisB, fill=viridisB] coordinates {
(ColQwen2.5,0.716)
(ColQwen2.5 (Ctx),0.737)
(ColQwen2.5 (B\&Ctx),0.748)
(ColQwen2.5 (Q-Split),0.724)
(ColQwen2.5 (Rerank),0.767)
};
\addplot+[draw=viridisC, fill=viridisC] coordinates {
(ColQwen2.5,0.77)
(ColQwen2.5 (Ctx),0.801)
(ColQwen2.5 (B\&Ctx),0.814)
(ColQwen2.5 (Q-Split),0.786)
(ColQwen2.5 (Rerank),0.804)
};

\addplot+[forget plot, sharp plot, thick, dashed, draw=viridisA, fill=none]
    coordinates {(ColQwen2.5,0.658) (ColQwen2.5 (Rerank),0.658)};
\addplot+[forget plot, sharp plot, thick, dashed, draw=viridisB, fill=none]
    coordinates {(ColQwen2.5,0.716) (ColQwen2.5 (Rerank),0.716)};
\addplot+[forget plot, sharp plot, thick, dashed, draw=viridisC, fill=none]
    coordinates {(ColQwen2.5,0.77) (ColQwen2.5 (Rerank),0.77)};

\legend{@3,@5,@10}
\end{axis}
\end{tikzpicture}

\begin{tikzpicture}
\begin{axis}[
  title={ColNomic \cite{nomicembedmultimodal2025} variants: RecallAll@3/5/10},
  width=12cm,
  height=3.5cm,
  ybar,
  bar width=6pt,
  ymin=0.679678,
  ymax=0.865701,
  ylabel={RecallAll},
  symbolic x coords={ColNomic,ColNomic (Ctx),ColNomic (B\&Ctx),ColNomic (Hierarchical),ColNomic (Rerank),ColNomic (B\&Ctx\&Rank)},
  xtick=data,
  xticklabel style={rotate=25, anchor=east, font=\scriptsize},
  yticklabel style={font=\scriptsize},
  enlarge x limits=0.08,
  clip=false,
  grid=both,
  major grid style={opacity=0.2},
  minor grid style={opacity=0.1},
  legend style={font=\scriptsize, at={(1.02,1)}, anchor=north west},
  legend cell align=left,
]
\addplot+[draw=viridisA, fill=viridisA] coordinates {
(ColNomic,0.707)
(ColNomic (Ctx),0.697)
(ColNomic (B\&Ctx),0.719)
(ColNomic (Hierarchical),0.709)
(ColNomic (Rerank),0.735)
(ColNomic (B\&Ctx\&Rank),0.75)
};
\addplot+[draw=viridisB, fill=viridisB] coordinates {
(ColNomic,0.762)
(ColNomic (Ctx),0.763)
(ColNomic (B\&Ctx),0.787)
(ColNomic (Hierarchical),0.765)
(ColNomic (Rerank),0.786)
(ColNomic (B\&Ctx\&Rank),0.809)
};
\addplot+[draw=viridisC, fill=viridisC] coordinates {
(ColNomic,0.812)
(ColNomic (Ctx),0.823)
(ColNomic (B\&Ctx),0.848)
(ColNomic (Hierarchical),0.817)
(ColNomic (Rerank),0.831)
(ColNomic (B\&Ctx\&Rank),0.862)
};

\addplot+[forget plot, sharp plot, thick, dashed, draw=viridisA, fill=none]
    coordinates {(ColNomic,0.707) (ColNomic (B\&Ctx\&Rank),0.707)};
\addplot+[forget plot, sharp plot, thick, dashed, draw=viridisB, fill=none]
    coordinates {(ColNomic,0.762) (ColNomic (B\&Ctx\&Rank),0.762)};
\addplot+[forget plot, sharp plot, thick, dashed, draw=viridisC, fill=none]
    coordinates {(ColNomic,0.813) (ColNomic (B\&Ctx\&Rank),0.813)};

\legend{@3,@5,@10}
\end{axis}
\end{tikzpicture}

    \caption[Overall Base Model Variants Comparison]{Overall RecallAll@k results grouped by ColPali \cite{faysse2024colpali}, ColQwen2.5 \cite{faysse2024colpali}, and ColNomic~\cite{nomicembedmultimodal2025} variants for $k \in \{3, 5, 10\}$. B denotes base model and Ctx stands for document context embeddings}
    \label{fig:overall_results_colpali_colqwen_colnomic_variants_bar_chart}
\end{figure}

In Figure \ref{fig:overall_results_colpali_colqwen_colnomic_variants_bar_chart}, we show further evaluation results in micro-average grouped by the three base models ColPali~v1.3~\cite{faysse2024colpali}, ColQwen2.5 \cite{faysse2024colpali}, and ColNomic \cite{nomicembedmultimodal2025}. For ColPali v1.3 \cite{faysse2024colpali}, the base model has the lowest performance. Our assumed faster hierarchical approach shows a slightly higher performance across all k parameter values. However, the difference is more visible for our other approaches. For $k=3$, ColPali v1.3 (Rerank) is the best performing approach by RecallAll@k, even surpassing the RecallAll@10 of the base model. In general, ColPali v1.3 (Ctx) and ColPali v1.3 (Normal\&Ctx) display similar RecallAll@k scores. The relative performance, the performance gains for increased retrieved pages are stronger for our document context approach than for reranking. While RecallAll@5 is still the highest for ColPali v1.3 (Rerank), for RecallAll@10, ColPali v1.3 (Ctx) surpasses it.

The results are similar for the ColQwen2.5 \cite{faysse2024colpali} variants. One difference to the ColPali~v1.3~\cite{faysse2024colpali} variants is that ColQwen2.5 (Normal\&Ctx) improves upon ColQwen2.5 (Ctx), while the ColPali v1.3 ensemble performs slightly worse than the pure document context approach. One possible explanation for this might be that the ColQwen2.5 base model is relatively closer to the ColQwen2.5 (Ctx) model than ColPali v1.3 is to ColPali v1.3 (Ctx). Furthermore, we test for ColQwen2.5 query splitting approaches. Both ColQwen2.5 (Q-Split 1) and ColQwen2.5 (Q-Split 2) do overall perform worse in terms of RecallAll@k than the base ColQwen2.5 model \cite{faysse2024colpali}, whereas ColQwen2.5 (Q-Split 3) has a higher RecallAll@k. As for ColPali v1.3 \cite{faysse2024colpali}, the RecallAll@k increases the strongest for the approaches that include document context embeddings.

In the case of ColNomic \cite{nomicembedmultimodal2025} variants, the document context approach ColNomic (Ctx) has a slightly lower performance for $k=3$. However, as for the other two base models, ColPali~v1.3~\cite{faysse2024colpali} and ColQwen2.5 \cite{faysse2024colpali}, rising k values improve the performance stronger. As for ColQwen2.5~\cite{faysse2024colpali}, ColNomic (Ctx) profits from the usage of an ensemble with the retrieval results from the base model. ColNomic (Hierarchical) also exhibits the same performance as the base model, although it requires fewer computing resources during page retrieval. In contrast to the other two base models, reranking displays even for $k=5$ no better RecallAll@k values than ColNomic (Normal\&Ctx). Additionally, we include in this analysis the ensemble combination of the ColNomic base model with ColNomic (Ctx) and ColNomic (Rerank), which shows the best RecallAll@k for all parameter k values.

\subsection{Results by Query Evidence Structure Type and Data Source Type}
Furthermore, we evaluate the benchmark results in micro-average grouped by query evidence structure and data source type to determine whether the query evidence structure type performance differences are consistent for different data sources. Table \ref{tab:retrieval_combo_source_question_transposed} displays the RecallAll@5 metric for all pairs of query evidence structure types and data source types.

Across all data sources and models, we observe that multi-page queries are more challenging than single-page queries. Retrieval on Cross-Doc queries exhibits an even lower RecallAll@5 indicating that Cross-Doc queries are fundamentally more challenging than queries about cohesive document pages. Additionally, Table \ref{tab:retrieval_combo_source_question_transposed} reveals a slight tendency of Doc-Near queries being more challenging than Doc-Adjacent queries, and Doc-Far being more challenging than Doc-Near queries. 

For single-page queries, the best performing model is consistently ColNomic (Normal\&Ctx\&Rank). In contrast to this, the second-best model depends on the data source. For Wikipedia, arXiv, and PubMed, ColNomic (Rerank) is the second-best model, whereas for SEC Proxy Reports, ColNomic (Ctx) is the second-best. In the case of ColQwen2.5~\cite{faysse2024colpali}, we observe the same behavior for the reranking and document context embeddings approach, with the reranking model performing worse on the SEC Proxy Reports data source. A potential cause for this might be the missing context on pages closer to the end of a document in SEC Proxy Reports caused by missing company names on pages.

We observe a similar model behavior for multi-page queries. Document context embeddings lead to a higher RecallAll@5 for Wikipedia and SEC Proxy Reports, while for arXiv and PubMed reranking is more favorable. A possible explanation for this approach behavior could be that the scientific article data source queries are more clearly specified and reference articles more directly, which might lower the positive impact of document context embeddings.

Additionally, Table \ref{tab:retrieval_combo_source_question_transposed} shows that the RecallAll@5 differs strongly between different data sources for Cross-Doc queries. Whereas the best-performing model achieves a RecallAll@5 of 50.9\% for Wikipedia, the best metric values for SEC Proxy Reports are 26.6\%, for arXiv 18.0\%, and for PubMed 28.0\%. One possible explanation for the low RecallAll@5 scores might be that the queries are not well specified. This hypothesis could be supported by the significant drop in the RecallAll@5 from multi-page to Cross-Doc queries. Another explanation could be that cross-document retrieval is fundamentally harder, especially as document page counts grow, such as in SEC Proxy Reports and scientific articles. A third option might be that different query parts average themselves out in the late interaction max-similarity scoring. This is partly supported by the RecallAll@5 improvement when using the Q-Split 3 approach on ColQwen2.5 \cite{faysse2024colpali}, where the top k retrieved pages are an ensemble over the sub-queries. However, the still low score by absolute terms contradicts that this is the sole explanation. We conduct further analysis in the Appendix Subsections~\ref{appendix_subsec:page_count_reduction_analysis} and~\ref{appendix_subsec:cross_document_retrieval_error_analysis}.
Another observation for Cross-Doc queries is that document context embeddings are affecting the retrieval performance negatively in contrast to multi-page queries. The use of reranking improves over the base models ColQwen2.5 \cite{faysse2024colpali} and ColNomic \cite{nomicembedmultimodal2025}, however, does exceed the ColQwen2.5 (Q-Split) RecallAll@5 only for the Wikipedia data source.
\begin{table}[htbp]
\centering
\scriptsize
\setlength{\tabcolsep}{4pt}
\renewcommand{\arraystretch}{1.15}
\renewcommand{\rot}[1]{\rotatebox[origin=c]{90}{\parbox{2.5cm}{\centering #1}}}
\setlength{\tabcolsep}{3pt}
\begin{tabular}{ll|cccccc}
\toprule
 & Model 
& \rot{Page-Single}
& \rot{Page-Multi}
& \rot{Doc-Adjacent}
& \rot{Doc-Near}
& \rot{Doc-Far}
& \rot{Cross-Doc} \\
\midrule

\multirow{9}{*}{\rotatebox[origin=c]{90}{\parbox{2.5cm}{\centering Wikipedia}}}
& ColQwen3 \cite{huang2025tomoroqwen3}         & \textbf{\textcolor{thirdbestcell}{90.5}} & 90.3 & 57.1 & 49.7 & 49.8 & 45.2 \\
& ColQwen2.5 \cite{faysse2024colpali}          & 84.9 & 85.6 & 48.9 & 44.3 & 46.7 & 41.0 \\
& ColQwen2.5 (Ctx)                             & 83.8 & 85.0 & \textbf{\textcolor{thirdbestcell}{62.0}} & \textbf{\textcolor{secondbestcell}{59.9}} & 57.3 & 32.9 \\
& ColQwen2.5 (Q-Split)                       & 84.3 & 84.5 & 48.2 & 45.1 & 47.5 & 45.8 \\
& ColQwen2.5 (Rerank)                          & 89.9 & \textbf{\textcolor{thirdbestcell}{90.4}} & 56.0 & 49.8 & 58.4 & \textbf{\textcolor{bestcell}{50.9}} \\
& ColNomic \cite{nomicembedmultimodal2025}    & 89.4 & 89.7 & 54.3 & 50.0 & 53.3 & 48.5 \\
& ColNomic (Ctx)                               & 86.6 & 87.9 & \textbf{\textcolor{bestcell}{67.2}} & \textbf{\textcolor{bestcell}{64.4}} & \textbf{\textcolor{bestcell}{67.1}} & 34.2 \\
& ColNomic (Rerank)                            & \textbf{\textcolor{secondbestcell}{91.1}} & \textbf{\textcolor{secondbestcell}{91.3}} & 58.1 & 51.4 & \textbf{\textcolor{thirdbestcell}{59.6}} & \textbf{\textcolor{secondbestcell}{50.8}} \\
& ColNomic (Normal\&Ctx\&Rank)                 & \textbf{\textcolor{bestcell}{92.5}} & \textbf{\textcolor{bestcell}{93.0}} & \textbf{\textcolor{secondbestcell}{63.5}} & \textbf{\textcolor{thirdbestcell}{59.5}} & \textbf{\textcolor{secondbestcell}{62.4}} & \textbf{\textcolor{thirdbestcell}{49.8}} \\
\midrule

\multirow{9}{*}{\rotatebox[origin=c]{90}{\parbox{2.5cm}{\centering SEC Proxy Reports}}}
& ColQwen3 \cite{huang2025tomoroqwen3}         & \textbf{\textcolor{thirdbestcell}{83.4}} & 85.4 & 52.5 & 48.5 & 43.7 & 15.3 \\
& ColQwen2.5 \cite{faysse2024colpali}          & 71.0 & 73.8 & 36.3 & 37.4 & 35.3 & 10.0 \\
& ColQwen2.5 (Ctx)                             & 81.7 & 83.7 & \textbf{\textcolor{thirdbestcell}{57.3}} & \textbf{\textcolor{thirdbestcell}{54.9}} & \textbf{\textcolor{thirdbestcell}{51.5}} & 13.9 \\
& ColQwen2.5 (Q-Split)                       & 70.8 & 73.6 & 37.5 & 40.6 & 36.0 & \textbf{\textcolor{thirdbestcell}{17.5}} \\
& ColQwen2.5 (Rerank)                          & 78.3 & 80.6 & 47.3 & 45.9 & 42.9 & 15.3 \\
& ColNomic \cite{nomicembedmultimodal2025}    & 79.5 & 82.1 & 43.7 & 45.4 & 37.5 & 13.8 \\
& ColNomic (Ctx)                               & \textbf{\textcolor{secondbestcell}{87.1}} & \textbf{\textcolor{secondbestcell}{88.8}} & \textbf{\textcolor{bestcell}{62.1}} & \textbf{\textcolor{secondbestcell}{57.5}} & \textbf{\textcolor{bestcell}{57.3}} & \textbf{\textcolor{secondbestcell}{17.9}} \\
& ColNomic (Rerank)                            & 83.4 & \textbf{\textcolor{thirdbestcell}{85.7}} & 50.4 & 50.3 & 45.9 & 17.0 \\
& ColNomic (Normal\&Ctx\&Rank)                 & \textbf{\textcolor{bestcell}{89.0}} & \textbf{\textcolor{bestcell}{90.2}} & \textbf{\textcolor{secondbestcell}{59.8}} & \textbf{\textcolor{bestcell}{59.7}} & \textbf{\textcolor{secondbestcell}{53.7}} & \textbf{\textcolor{bestcell}{26.6}} \\
\midrule

\multirow{9}{*}{\rotatebox[origin=c]{90}{\parbox{2.5cm}{\centering arXiv}}}
& ColQwen3 \cite{huang2025tomoroqwen3}         & 93.2 & 94.6 & 74.3 & \textbf{\textcolor{secondbestcell}{77.1}} & 71.1 & 11.5 \\
& ColQwen2.5 \cite{faysse2024colpali}          & 90.5 & 93.5 & 68.5 & 68.8 & 65.5 & 7.4 \\
& ColQwen2.5 (Ctx)                             & 90.1 & 93.2 & 74.4 & 71.9 & 63.3 & 3.5 \\
& ColQwen2.5 (Q-Split)                       & 89.4 & 92.3 & 65.2 & 66.2 & 63.6 & \textbf{\textcolor{bestcell}{18.0}} \\
& ColQwen2.5 (Rerank)                          & \textbf{\textcolor{thirdbestcell}{94.1}} & \textbf{\textcolor{thirdbestcell}{96.1}} & \textbf{\textcolor{thirdbestcell}{74.8}} & \textbf{\textcolor{thirdbestcell}{76.0}} & \textbf{\textcolor{thirdbestcell}{72.2}} & 12.4 \\
& ColNomic \cite{nomicembedmultimodal2025}    & 93.0 & 95.5 & 73.4 & 72.1 & 69.4 & 11.2 \\
& ColNomic (Ctx)                               & 91.5 & 94.4 & 74.7 & 70.5 & 65.7 & 4.2 \\
& ColNomic (Rerank)                            & \textbf{\textcolor{secondbestcell}{95.2}} & \textbf{\textcolor{secondbestcell}{96.9}} & \textbf{\textcolor{secondbestcell}{77.6}} & 75.4 & \textbf{\textcolor{secondbestcell}{73.1}} & \textbf{\textcolor{secondbestcell}{15.1}} \\
& ColNomic (Normal\&Ctx\&Rank)                 & \textbf{\textcolor{bestcell}{95.8}} & \textbf{\textcolor{bestcell}{97.1}} & \textbf{\textcolor{bestcell}{79.5}} & \textbf{\textcolor{bestcell}{77.3}} & \textbf{\textcolor{bestcell}{75.5}} & \textbf{\textcolor{thirdbestcell}{14.1}} \\
\midrule

\multirow{9}{*}{\rotatebox[origin=c]{90}{\parbox{2.5cm}{\centering PubMed}}}
& ColQwen3 \cite{huang2025tomoroqwen3}         & 93.4 & 94.8 & 75.4 & 69.2 & 67.7 & 18.7 \\
& ColQwen2.5 \cite{faysse2024colpali}          & 92.7 & 94.7 & 73.9 & 71.8 & 66.9 & 14.3 \\
& ColQwen2.5 (Ctx)                             & 92.4 & 94.6 & 77.0 & \textbf{\textcolor{thirdbestcell}{76.0}} & 63.3 & 5.1 \\
& ColQwen2.5 (Q-Split)                       & 91.4 & 93.7 & 71.8 & 72.4 & 62.5 & \textbf{\textcolor{bestcell}{28.0}} \\
& ColQwen2.5 (Rerank)                          & \textbf{\textcolor{thirdbestcell}{95.3}} & \textbf{\textcolor{thirdbestcell}{96.7}} & \textbf{\textcolor{thirdbestcell}{79.3}} & 75.3 & \textbf{\textcolor{thirdbestcell}{72.4}} & 19.6 \\
& ColNomic \cite{nomicembedmultimodal2025}    & 94.3 & 96.1 & 77.1 & 74.3 & 70.0 & 18.5 \\
& ColNomic (Ctx)                               & 92.9 & 95.2 & 78.0 & 74.7 & 66.7 & 5.6 \\
& ColNomic (Rerank)                            & \textbf{\textcolor{secondbestcell}{95.7}} & \textbf{\textcolor{secondbestcell}{97.1}} & \textbf{\textcolor{secondbestcell}{79.8}} & \textbf{\textcolor{secondbestcell}{76.1}} & \textbf{\textcolor{secondbestcell}{72.9}} & \textbf{\textcolor{secondbestcell}{20.0}} \\
& ColNomic (Normal\&Ctx\&Rank)                 & \textbf{\textcolor{bestcell}{96.3}} & \textbf{\textcolor{bestcell}{97.4}} & \textbf{\textcolor{bestcell}{82.9}} & \textbf{\textcolor{bestcell}{79.6}} & \textbf{\textcolor{bestcell}{73.7}} & \textbf{\textcolor{thirdbestcell}{20.0}} \\
\bottomrule
\end{tabular}

\caption[Retrieval Results by Data Source and Query Evidence Structure Type]{Retrieval results by data source and query evidence structure type with the RecallAll@5 evaluation metric. The best models are highlighted with \textbf{\textcolor{bestcell}{blue}}, the second best with \textbf{\textcolor{secondbestcell}{green}}, and the third best has \textbf{\textcolor{thirdbestcell}{bold}} text.}
\label{tab:retrieval_combo_source_question_transposed}
\end{table}

\subsection{Results by Evidence Type}
Furthermore, we compare the benchmark results grouped by evidence type as shown in Table \ref{tab:retrieval_by_evidence_type}. We choose as evidence groups text, table, and image which refer to the cases that only this evidence is the ground truth. We exclude combinations of evidence types, such as tables with text, to remove the impact of inter-dependencies. 
Overall, nDCGAll@k and RecallAll@k are similar for the text and image evidence types, with a slight tendency for image evidence being harder to find. In contrast to those, the retrieval performance is consistently lower for table-only evidence. Across all evidence type and model pairs, ColNomic (Normal\&Ctx\&Rank) exhibits the highest nDCGAll@k and RecallAll@k values.
In case of text-only evidence, the second-best model depends on the metric and the number of retrieved pages. For lower k values, ColQwen3 \cite{huang2025tomoroqwen3} ranks as the second-best model, while for higher k values, ColNomic (Rerank) is the second-best model.
For both table-only and image-only evidence, ColNomic (Rerank) is the second-best model except for RecallAll@k, where ColNomic (Ctx) scores higher.
\begin{table}[htbp]
\centering
\scriptsize
\begin{tabular}{ll|ccc|ccc}
\toprule
Evidence Type & Model & \multicolumn{3}{c}{nDCGAll@k \cite{jarvelin2002}} & \multicolumn{3}{c}{RecallAll@k} \\
\cmidrule(lr){3-5} \cmidrule(lr){6-8}
 &  & 3 & 5 & 10 & 3 & 5 & 10 \\
\midrule
\multirow{9}{*}{\rotatebox[origin=c]{90}{\parbox{2cm}{\centering Text}}} & ColQwen3 \cite{huang2025tomoroqwen3} & \textbf{\textcolor{secondbestcell}{68.5}} & \textbf{\textcolor{secondbestcell}{71.1}} & \textbf{\textcolor{thirdbestcell}{73.3}} & \textbf{\textcolor{secondbestcell}{73.5}} & \textbf{\textcolor{thirdbestcell}{78.1}} & \textbf{\textcolor{thirdbestcell}{82.3}} \\
 & ColQwen2.5 \cite{faysse2024colpali} & 59.3 & 62.4 & 65.0 & 65.6 & 71.6 & 77.4 \\
 & ColQwen2.5 (Ctx) & 59.0 & 62.4 & 65.3 & 65.8 & 72.4 & 78.9 \\
 & ColQwen2.5 (Q-Split) & 58.3 & 61.8 & 64.8 & 65.4 & 72.2 & 78.9 \\
 & ColQwen2.5 (Rerank) & 67.1 & 69.7 & 71.8 & 72.1 & 76.8 & 80.8 \\
 & ColNomic \cite{nomicembedmultimodal2025} & 64.8 & 67.7 & 70.3 & 70.8 & 76.3 & 81.6 \\
 & ColNomic (Ctx) & 62.6 & 65.9 & 68.6 & 68.8 & 75.1 & 81.0 \\
 & ColNomic (Rerank) & \textbf{\textcolor{thirdbestcell}{68.2}} & \textbf{\textcolor{thirdbestcell}{71.0}} & \textbf{\textcolor{secondbestcell}{73.5}} & \textbf{\textcolor{thirdbestcell}{73.5}} & \textbf{\textcolor{secondbestcell}{78.6}} & \textbf{\textcolor{secondbestcell}{83.5}} \\
 & ColNomic (Normal\&Ctx\&Rank) & \textbf{\textcolor{bestcell}{68.9}} & \textbf{\textcolor{bestcell}{72.1}} & \textbf{\textcolor{bestcell}{74.8}} & \textbf{\textcolor{bestcell}{74.7}} & \textbf{\textcolor{bestcell}{80.4}} & \textbf{\textcolor{bestcell}{85.7}} \\
\midrule
\multirow{9}{*}{\rotatebox[origin=c]{90}{\parbox{2cm}{\centering Table}}} & ColQwen3 \cite{huang2025tomoroqwen3} & 58.2 & 61.5 & 64.1 & 62.9 & 68.3 & 73.4 \\
 & ColQwen2.5 \cite{faysse2024colpali} & 51.9 & 55.4 & 58.3 & 57.1 & 63.1 & 69.0 \\
 & ColQwen2.5 (Ctx) & 53.7 & 58.0 & 61.7 & 59.5 & 67.0 & 74.2 \\
 & ColQwen2.5 (Q-Split) & 53.1 & 56.7 & 59.8 & 58.6 & 65.0 & 71.5 \\
 & ColQwen2.5 (Rerank) & \textbf{\textcolor{thirdbestcell}{59.5}} & \textbf{\textcolor{thirdbestcell}{62.6}} & 64.7 & \textbf{\textcolor{thirdbestcell}{64.0}} & 69.1 & 73.0 \\
 & ColNomic \cite{nomicembedmultimodal2025} & 57.5 & 61.1 & 64.0 & 62.5 & 68.6 & 74.1 \\
 & ColNomic (Ctx) & 57.0 & 61.5 & \textbf{\textcolor{thirdbestcell}{65.3}} & 62.6 & \textbf{\textcolor{thirdbestcell}{70.0}} & \textbf{\textcolor{secondbestcell}{77.0}} \\
 & ColNomic (Rerank) & \textbf{\textcolor{secondbestcell}{60.7}} & \textbf{\textcolor{secondbestcell}{64.1}} & \textbf{\textcolor{secondbestcell}{66.8}} & \textbf{\textcolor{secondbestcell}{65.4}} & \textbf{\textcolor{secondbestcell}{71.1}} & \textbf{\textcolor{thirdbestcell}{76.3}} \\
 & ColNomic (Normal\&Ctx\&Rank) & \textbf{\textcolor{bestcell}{62.6}} & \textbf{\textcolor{bestcell}{67.0}} & \textbf{\textcolor{bestcell}{70.4}} & \textbf{\textcolor{bestcell}{68.1}} & \textbf{\textcolor{bestcell}{75.3}} & \textbf{\textcolor{bestcell}{81.7}} \\
\midrule
\multirow{9}{*}{\rotatebox[origin=c]{90}{\parbox{2cm}{\centering Image}}} & ColQwen3 \cite{huang2025tomoroqwen3} & \textbf{\textcolor{thirdbestcell}{64.3}} & \textbf{\textcolor{thirdbestcell}{67.5}} & \textbf{\textcolor{thirdbestcell}{69.8}} & \textbf{\textcolor{thirdbestcell}{71.3}} & 77.3 & 82.0 \\
 & ColQwen2.5 \cite{faysse2024colpali} & 54.7 & 58.0 & 60.6 & 62.1 & 68.6 & 74.3 \\
 & ColQwen2.5 (Ctx) & 56.6 & 60.9 & 64.0 & 65.3 & 74.1 & 81.7 \\
 & ColQwen2.5 (Q-Split) & 54.3 & 58.0 & 60.9 & 62.3 & 69.6 & 75.9 \\
 & ColQwen2.5 (Rerank) & 63.7 & 66.4 & 68.2 & 69.7 & 74.6 & 77.8 \\
 & ColNomic \cite{nomicembedmultimodal2025} & 61.7 & 65.0 & 67.6 & 68.7 & 74.8 & 80.0 \\
 & ColNomic (Ctx) & 61.2 & 65.4 & 68.4 & 69.4 & \textbf{\textcolor{thirdbestcell}{77.9}} & \textbf{\textcolor{secondbestcell}{84.9}} \\
 & ColNomic (Rerank) & \textbf{\textcolor{secondbestcell}{66.1}} & \textbf{\textcolor{secondbestcell}{69.2}} & \textbf{\textcolor{secondbestcell}{71.3}} & \textbf{\textcolor{secondbestcell}{72.7}} & \textbf{\textcolor{secondbestcell}{78.3}} & \textbf{\textcolor{thirdbestcell}{82.4}} \\
 & ColNomic (Normal\&Ctx\&Rank) & \textbf{\textcolor{bestcell}{66.8}} & \textbf{\textcolor{bestcell}{70.7}} & \textbf{\textcolor{bestcell}{73.4}} & \textbf{\textcolor{bestcell}{74.1}} & \textbf{\textcolor{bestcell}{81.1}} & \textbf{\textcolor{bestcell}{86.9}} \\
\bottomrule
\end{tabular}
\caption[Retrieval Results by Evidence Type]{Retrieval results by evidence type with the nDCGAll@k \cite{jarvelin2002} and RecallAll@k as evaluation metrics. The best models are highlighted with \textbf{\textcolor{bestcell}{blue}}, the second best with \textbf{\textcolor{secondbestcell}{green}}, and the third best has \textbf{\textcolor{thirdbestcell}{bold}} text. The numbers below the evaluation metrics represent the values of parameter k. The three evidence groups text, table, and image refer to the cases that only this evidence is the ground truth. Combinations are not shown to remove the impact of inter-dependencies.}
\label{tab:retrieval_by_evidence_type}
\end{table}

\subsection{Results by Domain Type}
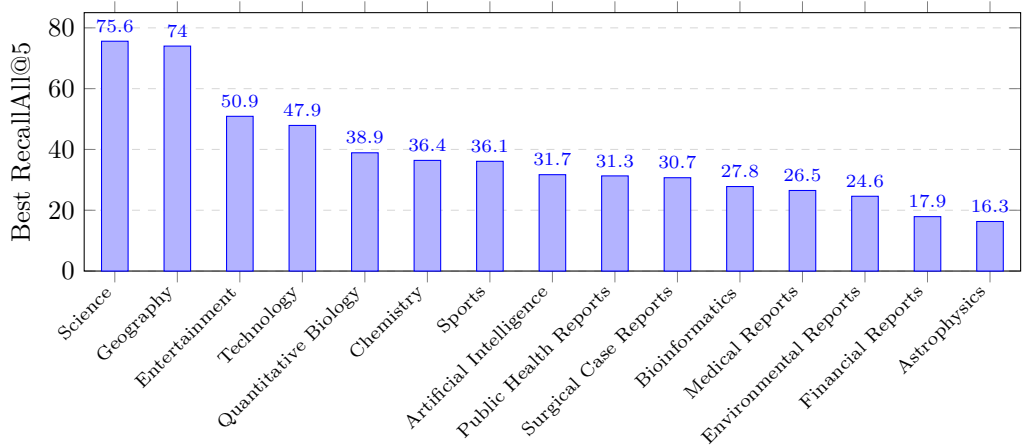
\begin{figure}[htbp]
\centering
\begin{tikzpicture}
\begin{axis}[
    ybar,
    width=\linewidth,
    height=5.0cm,
    bar width=10pt,
    ymin=0,
    ymax=85,
    ylabel={Best RecallAll@5},
    symbolic x coords={
        Science,
        Geography,
        Entertainment,
        Technology,
        Quantitative Biology,
        Chemistry,
        Sports,
        Artificial Intelligence,
        Public Health Reports,
        Surgical Case Reports,
        Bioinformatics,
        Medical Reports,
        Environmental Reports,
        Financial Reports,
        Astrophysics
    },
    xtick=data,
    x tick label style={
        rotate=45,
        anchor=east,
        font=\scriptsize
    },
    ymajorgrids=true,
    grid style={dashed,gray!30},
    nodes near coords,
    nodes near coords align={vertical},
    every node near coord/.append style={font=\scriptsize},
    enlarge x limits=0.035,
]
\addplot coordinates {
    (Science,75.6)
    (Geography,74.0)
    (Entertainment,50.9)
    (Technology,47.9)
    (Quantitative Biology,38.9)
    (Chemistry,36.4)
    (Sports,36.1)
    (Artificial Intelligence,31.7)
    (Public Health Reports,31.3)
    (Surgical Case Reports,30.7)
    (Bioinformatics,27.8)
    (Medical Reports,26.5)
    (Environmental Reports,24.6)
    (Financial Reports,17.9)
    (Astrophysics,16.3)
};
\end{axis}
\end{tikzpicture}
\caption{
Best cross-document retrieval RecallAll@5 for domains with at least 100 Cross-Doc queries. For each domain, the value is the maximum over all evaluated models.
}
\label{fig:cross_doc_recall_domains_ge_100}
\end{figure}

In Figure \ref{fig:cross_doc_recall_domains_ge_100}, we show further benchmark evaluation results in micro-average grouped by domain type. We select all domains with at least 100 Cross-Doc queries and take the maximum RecallAll@5 of the following six models: ColQwen3, ColQwen2.5, ColQwen2.5 (Ctx), ColQwen2.5 (Q-Split), ColQwen2.5 (Reranking), ColNomic, ColNomic (Ctx), and  ColNomic (Reranking). The evaluation results show that there is a large different between the domains. While retrieval on the \textit{Science} domain, sourced from Wikipedia, achieves a RecallAll@5 of 75.6\%, retrieval on the \textit{Financial Reports} and \textit{Astrophysics} domain less than 20\%.

\subsection{Document Context Embeddings Vector Count Impact}
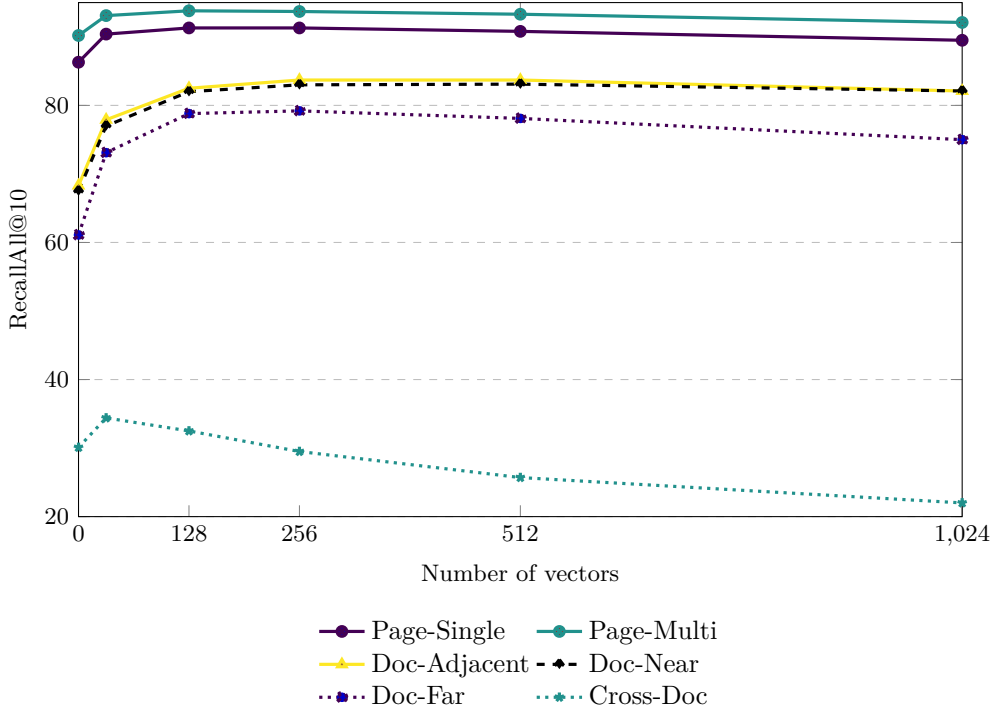
\begin{figure}[htbp]
\centering
\begin{tikzpicture}
\begin{axis}[
    width=0.95\linewidth,
    height=0.6\linewidth,
    xlabel={Number of vectors},
    ylabel={RecallAll@10},
    xmin=0, xmax=1024,
    ymin=20, ymax=95,
    xtick={0,128,256,512,1024},
    ymajorgrids=true,
    grid style=dashed,
    legend style={
        at={(0.5,-0.18)},
        anchor=north,
        legend columns=2,
        draw=none
    },
    tick label style={font=\small},
    label style={font=\small},
    legend cell align=left,
]

\addplot+[very thick, mark=otimes*, mark size=1.8pt, color=viridisA] coordinates {
    (0,86.3) (32, 90.4) (128,91.3) (256,91.3) (512,90.8) (1024,89.5)
};
\addlegendentry{Page-Single}

\addplot+[very thick, mark=otimes*, mark size=1.8pt, color=viridisB] coordinates {
    (0,90.2) (32, 93.1) (128,93.8) (256,93.7) (512,93.3) (1024,92.1)
};
\addlegendentry{Page-Multi}

\addplot+[very thick, mark=triangle*, mark size=1.8pt, color=viridisC] coordinates {
    (0,68.3) (32, 77.9) (128,82.5) (256,83.7) (512,83.7) (1024,82.1)
};
\addlegendentry{Doc-Adjacent}

\addplot+[very thick, mark=diamond*, mark size=1.8pt, dashed] coordinates {
    (0,67.5) (32, 77.0) (128,82.0) (256,83.0) (512,83.1) (1024,82.1)
};
\addlegendentry{Doc-Near}

\addplot+[very thick, mark=otimes*, mark size=1.8pt, dotted, color=viridisA] coordinates {
    (0,61.1) (32, 73.1) (128,78.8) (256,79.2) (512,78.1) (1024,75.0)
};
\addlegendentry{Doc-Far}

\addplot+[very thick, mark=star, mark size=1.8pt, dotted, color=viridisB] coordinates {
    (0,30.1) (32, 34.4) (128,32.5) (256,29.5) (512,25.7) (1024,22.0)
};
\addlegendentry{Cross-Doc}

\end{axis}
\end{tikzpicture}
\caption[Document Context Embeddings Vectors Count Impact]{Analysis of vectors count impact in document context embeddings measured by RecallAll@10 for each query evidence structure type for ColQwen2.5 \cite{faysse2024colpali} as base model. We omit the cross-document line for reasons of clarity since the low RecallAll@10 values would distort the axis.}
\label{fig:vectors_count_impact_recallall10_by_question_type}
\end{figure}

To find out the relation between document context embedding vectors count $n$ and retrieval performance, we analyze the retrieval performance for $n \in \{32, 128, 256, 512, 1024\}$. We display the results of this analysis in Figure \ref{fig:vectors_count_impact_recallall10_by_question_type}, where each line contains the results for all $n$ values for a specific query evidence structure type. As the evaluation metric, we choose RecallAll@10. The results show that the optimal value for $n$ depends on the query evidence structure type. For the single-page queries Page-Single and Page-Multi, ColQwen2.5 reaches the optimal performance between 128 and 256 document context embedding vectors. Whereas for Doc-Adjacent and Doc-Near queries, the optimum is between 256 and 512 vectors. The optimum for Doc-Distant queries is around 256 document context embedding vectors. In contrast to single-page and multi-page queries, we observe the optimum for Cross-Doc queries to be around $n=32$. One explanation for the lower optimal $n$ for Cross-Doc queries might be that document context embeddings favor the pages of the closest document, especially for queries that mention another document only with a few tokens. However, we observe that document context embeddings improve the RecallAll@10 across all query evidence structure types for a certain $n$ in case of our benchmark.

\subsection{Page Count Reduction Analysis}
\label{appendix_subsec:page_count_reduction_analysis}
To identify potential causes for our benchmark dataset being challenging, we evaluate the base models ColPali v1.3 \cite{faysse2024colpali}, ColQwen2.5 \cite{faysse2024colpali}, and ColNomic \cite{nomicembedmultimodal2025} on the reduced document dataset with micro-average results. We reduce the corpus size from 388,548 document page images to 38,551 by removing all pages that are not part of the evidence for at least one query. Table \ref{tab:retrieval_combo_source_question_reduced_dataset} displays the evaluation results for the combination of query evidence structure type and data source. In general, we observe an increase in RecallAll@5 across all models for every query evidence structure and data source type pair.

The RecallAll@5 is in the case of single-page queries for arXiv and PubMed close to perfect retrieval. Since those RecallAll@5 are already for the full dataset, this might indicate that the large dataset and in especially many pages per document cause the missing percentage points. For Wikipedia and SEC Proxy Reports, the RecallAll@5 gap to perfect retrieval still remains, however, still leaves room for improvement. Thus, we hypothesize that, especially for SEC Proxy Reports, there are factors other than high page count which prohibit perfect retrieval for single-page queries.

\begin{table}[htbp]
\centering
\scriptsize
\begin{tabular}{ll|cccc}
\toprule
Evidence Structure & Model & Wikipedia & SEC Proxy Reports & arXiv & PubMed \\
\midrule
\multirow{6}{*}{{\parbox{2cm}{\centering Page-Single}}} & ColPali & 63.9 & 41.5 & 70.4 & 79.4 \\
 & ColQwen & 84.9 & 71.0 & 90.5 & 92.7 \\
 & ColNomic & \textbf{\textcolor{thirdbestcell}{89.4}} & \textbf{\textcolor{thirdbestcell}{79.5}} & \textbf{\textcolor{thirdbestcell}{93.0}} & \textbf{\textcolor{thirdbestcell}{94.3}} \\
 & ColPali Reduced & 78.8 & 53.6 & 85.0 & 89.4 \\
 & ColQwen Reduced & \textbf{\textcolor{secondbestcell}{92.7}} & \textbf{\textcolor{secondbestcell}{81.1}} & \textbf{\textcolor{secondbestcell}{96.9}} & \textbf{\textcolor{secondbestcell}{97.2}} \\
 & ColNomic Reduced & \textbf{\textcolor{bestcell}{95.5}} & \textbf{\textcolor{bestcell}{86.8}} & \textbf{\textcolor{bestcell}{98.1}} & \textbf{\textcolor{bestcell}{97.9}} \\
\midrule
\multirow{6}{*}{{\parbox{2cm}{\centering Page-Multi}}} & ColPali & 67.3 & 44.6 & 78.3 & 85.3 \\
 & ColQwen & 85.6 & 73.8 & 93.5 & 94.7 \\
 & ColNomic & \textbf{\textcolor{thirdbestcell}{89.7}} & \textbf{\textcolor{thirdbestcell}{82.1}} & \textbf{\textcolor{thirdbestcell}{95.5}} & \textbf{\textcolor{thirdbestcell}{96.1}} \\
 & ColPali Reduced & 80.6 & 56.9 & 89.6 & 92.4 \\
 & ColQwen Reduced & \textbf{\textcolor{secondbestcell}{92.6}} & \textbf{\textcolor{secondbestcell}{83.4}} & \textbf{\textcolor{secondbestcell}{98.0}} & \textbf{\textcolor{secondbestcell}{97.9}} \\
 & ColNomic Reduced & \textbf{\textcolor{bestcell}{95.1}} & \textbf{\textcolor{bestcell}{88.7}} & \textbf{\textcolor{bestcell}{99.0}} & \textbf{\textcolor{bestcell}{98.6}} \\
\midrule
\multirow{6}{*}{{\parbox{2cm}{\centering Doc-Adjacent}}} & ColPali & 30.3 & 13.8 & 43.1 & 57.2 \\
 & ColQwen & 48.9 & 36.3 & 68.5 & 73.9 \\
 & ColNomic & \textbf{\textcolor{thirdbestcell}{54.3}} & \textbf{\textcolor{thirdbestcell}{43.7}} & \textbf{\textcolor{thirdbestcell}{73.4}} & 77.1 \\
 & ColPali Reduced & 51.3 & 26.9 & 66.9 & \textbf{\textcolor{thirdbestcell}{77.3}} \\
 & ColQwen Reduced & \textbf{\textcolor{secondbestcell}{67.6}} & \textbf{\textcolor{secondbestcell}{57.3}} & \textbf{\textcolor{secondbestcell}{87.8}} & \textbf{\textcolor{secondbestcell}{89.0}} \\
 & ColNomic Reduced & \textbf{\textcolor{bestcell}{73.2}} & \textbf{\textcolor{bestcell}{61.7}} & \textbf{\textcolor{bestcell}{91.4}} & \textbf{\textcolor{bestcell}{91.3}} \\
\midrule
\multirow{6}{*}{{\parbox{2cm}{\centering Doc-Near}}} & ColPali & 29.2 & 15.0 & 44.3 & 53.5 \\
 & ColQwen & 44.3 & 37.4 & 68.8 & 71.8 \\
 & ColNomic & 50.0 & \textbf{\textcolor{thirdbestcell}{45.4}} & \textbf{\textcolor{thirdbestcell}{72.1}} & 74.3 \\
 & ColPali Reduced & \textbf{\textcolor{thirdbestcell}{51.9}} & 31.7 & 67.7 & \textbf{\textcolor{thirdbestcell}{75.5}} \\
 & ColQwen Reduced & \textbf{\textcolor{secondbestcell}{61.8}} & \textbf{\textcolor{secondbestcell}{58.2}} & \textbf{\textcolor{secondbestcell}{90.1}} & \textbf{\textcolor{secondbestcell}{89.3}} \\
 & ColNomic Reduced & \textbf{\textcolor{bestcell}{69.4}} & \textbf{\textcolor{bestcell}{63.3}} & \textbf{\textcolor{bestcell}{90.9}} & \textbf{\textcolor{bestcell}{90.5}} \\
\midrule
\multirow{6}{*}{{\parbox{2cm}{\centering Doc-Far}}} & ColPali & 30.6 & 11.3 & 43.2 & 52.2 \\
 & ColQwen & 46.7 & 35.3 & 65.5 & 66.9 \\
 & ColNomic & 53.3 & \textbf{\textcolor{thirdbestcell}{37.5}} & \textbf{\textcolor{thirdbestcell}{69.4}} & 70.0 \\
 & ColPali Reduced & \textbf{\textcolor{thirdbestcell}{57.6}} & 25.0 & 69.0 & \textbf{\textcolor{thirdbestcell}{74.2}} \\
 & ColQwen Reduced & \textbf{\textcolor{secondbestcell}{70.2}} & \textbf{\textcolor{secondbestcell}{53.4}} & \textbf{\textcolor{secondbestcell}{91.8}} & \textbf{\textcolor{secondbestcell}{87.4}} \\
 & ColNomic Reduced & \textbf{\textcolor{bestcell}{78.4}} & \textbf{\textcolor{bestcell}{57.4}} & \textbf{\textcolor{bestcell}{94.6}} & \textbf{\textcolor{bestcell}{89.1}} \\
\midrule
\multirow{6}{*}{{\parbox{2cm}{\centering Cross-Doc}}} & ColPali & 32.4 & 1.6 & 3.6 & 10.4 \\
 & ColQwen & 41.0 & 10.0 & 7.4 & 14.3 \\
 & ColNomic & \textbf{\textcolor{secondbestcell}{48.5}} & \textbf{\textcolor{thirdbestcell}{13.8}} & \textbf{\textcolor{thirdbestcell}{11.2}} & 18.5 \\
 & ColPali Reduced & 39.0 & 3.5 & 8.5 & \textbf{\textcolor{thirdbestcell}{20.4}} \\
 & ColQwen Reduced & \textbf{\textcolor{thirdbestcell}{46.7}} & \textbf{\textcolor{secondbestcell}{16.1}} & \textbf{\textcolor{secondbestcell}{15.2}} & \textbf{\textcolor{secondbestcell}{25.5}} \\
 & ColNomic Reduced & \textbf{\textcolor{bestcell}{54.8}} & \textbf{\textcolor{bestcell}{19.7}} & \textbf{\textcolor{bestcell}{20.8}} & \textbf{\textcolor{bestcell}{31.4}} \\
\bottomrule
\end{tabular}
\caption{RecallAll@5 by data source and query evidence structure type comparing the the retrieval on the full and reduced dataset. The model names with suffix 'Reduced' are the models which are evaluated on the reduced dataset, whereas the other models are evaluated on the whole document dataset. We evaluate the three base models ColPali \cite{faysse2024colpali}, ColQwen2.5 \cite{faysse2024colpali}, and ColNomic \cite{nomicembedmultimodal2025}. The best models are highlighted with \textbf{\textcolor{bestcell}{blue}}, the second best with \textbf{\textcolor{secondbestcell}{green}}, and the third best has \textbf{\textcolor{thirdbestcell}{bold}} text.}
\label{tab:retrieval_combo_source_question_reduced_dataset}
\end{table}

In the case of multi-page queries, we observe a strong increase in RecallAll@5 for all models. We observe across all four data sources room for improvement also with retrieval on the reduced dataset. This does apply especially for the Wikipedia and SEC Proxy Reports queries. For example, the best base model ColNomic \cite{nomicembedmultimodal2025} has still an absolute RecallAll@5 gap of 38.3\% for Doc-Adjacent. Therefore, this strongly indicates that there are other factors than the large page count in the dataset that make the benchmark challenging.

For Wikipedia Cross-Doc queries, the results show a similar pattern to that for multi-page queries, but with a lower RecallAll@5 increase in the reduced dataset retrieval. This could indicate that Cross-Doc query retrieval is harder than multi-page queries, not only because of the larger number of similar pages to a query, but also because of other factors such as the lack of document diversity in the top k retrieval results. For the other three datasets, the RecallAll@5 values are in absolute terms very low for the full dataset with no model crossing 19\%. After our page count reduction, this changes for the three data sources. The best base model ColNomic \cite{nomicembedmultimodal2025} achieves on arXiv queries 20.8\% and on PubMed queries 31.4\% after the reductions. In absolute terms, those values are still low, however it shows that one cause of the very low RecallAll@5 values for the whole dataset can be attributed to the large page count. The missing RecallAll@5 percentage points might be caused by a fundamental problem in the late-interaction max-similarity calculation, still a large number of possible pages, or wrongly formulated Cross-Doc queries. One argument for the problem in the late-interaction max-similarity calculation is that the formula sums over the maximum-similarity of each token. When a query primarily consists of tokens that reference one specific document, the pages from the other documents might receive a much lower score. This is supported by the difference of the RecallAll@5 to the much higher Recall@5. For example, ColNomic \cite{nomicembedmultimodal2025} has for PubMed queries on the reduced dataset a RecallAll@5 of 31.4\% and a Recall@5 of 57.7\%. For SEC Proxy reports, this difference is relatively even larger with a RecallAll@5 of 19.7\% and a Recall@5 of 45.6\%. However, at the same time this might also indicate that some Cross-Doc queries are incorrectly created or categorized.

\subsection{Cross-Document Retrieval Error Analysis}
\label{appendix_subsec:cross_document_retrieval_error_analysis}
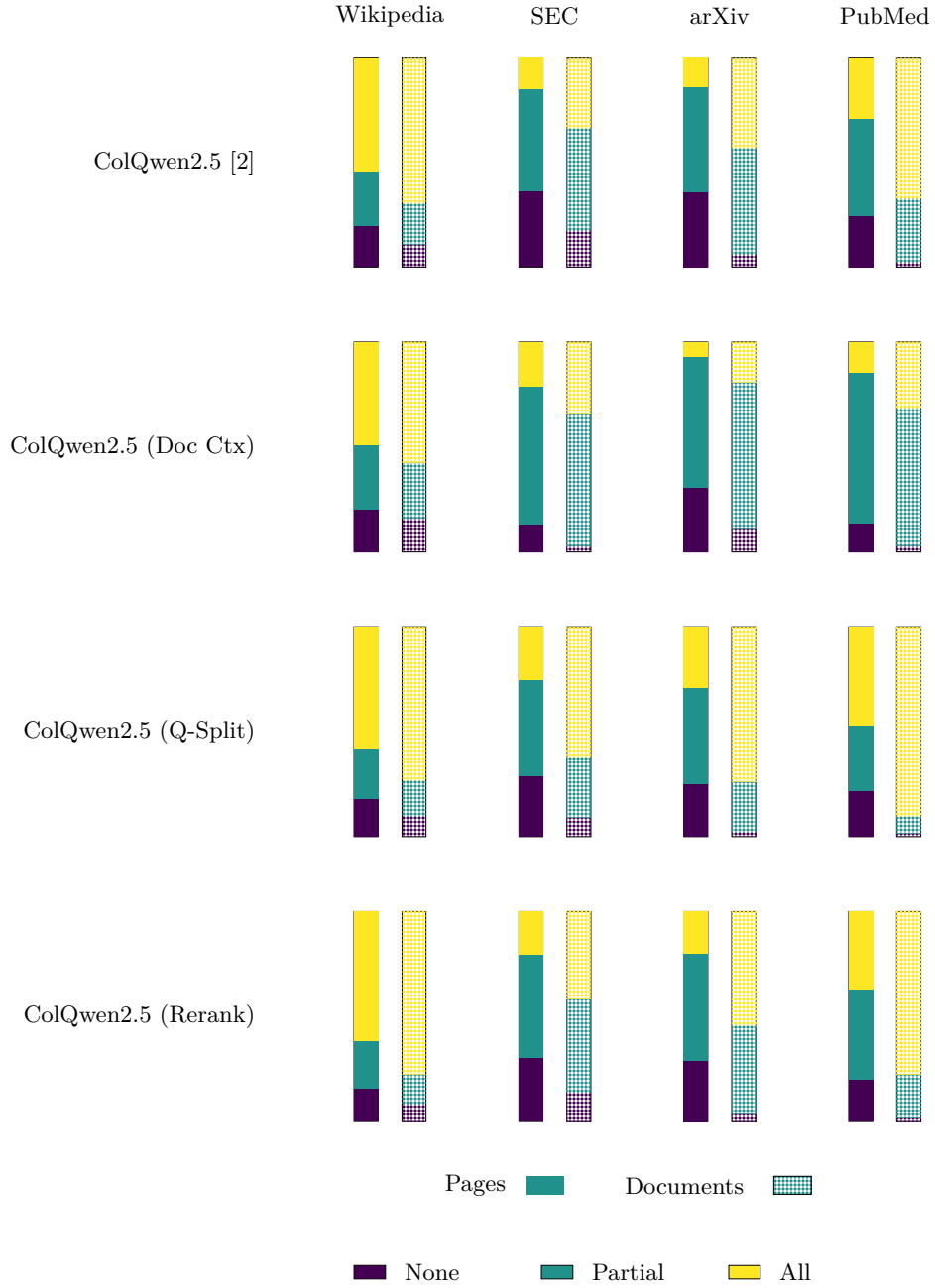
\begin{figure}[htbp]
    \centering
\begin{tikzpicture}[x=1cm,y=1cm]
\small

\definecolor{viridisA}{RGB}{68,1,84}
\definecolor{viridisB}{RGB}{33,145,140}
\definecolor{viridisC}{RGB}{253,231,37}

\node[font=\small] at (0.480,14.750) {Wikipedia};
\node[font=\small] at (2.680,14.750) {SEC};
\node[font=\small] at (4.880,14.750) {arXiv};
\node[font=\small] at (7.080,14.750) {PubMed};

\node[anchor=east,font=\small] at (-1.200,12.800) {ColQwen2.5 \cite{faysse2024colpali}};
\draw[line width=0.4pt] (0.000,11.400) rectangle (0.320,14.200);
\draw[line width=0.4pt] (0.640,11.400) rectangle (0.960,14.200);
\filldraw[fill=viridisA,draw=none] (0.000,11.400) rectangle (0.320,11.941);
\filldraw[fill=viridisB,draw=none] (0.000,11.941) rectangle (0.320,12.681);
\filldraw[fill=viridisC,draw=none] (0.000,12.681) rectangle (0.320,14.200);
\filldraw[pattern={Dots[distance=1.5pt, radius=0.6pt]},pattern color=viridisA,draw=none] (0.640,11.400) rectangle (0.960,11.699);
\filldraw[pattern={Dots[distance=1.5pt, radius=0.6pt]},pattern color=viridisB,draw=none] (0.640,11.699) rectangle (0.960,12.241);
\filldraw[pattern={Dots[distance=1.5pt, radius=0.6pt]},pattern color=viridisC,draw=none] (0.640,12.241) rectangle (0.960,14.200);
\draw[line width=0.4pt] (2.200,11.400) rectangle (2.520,14.200);
\draw[line width=0.4pt] (2.840,11.400) rectangle (3.160,14.200);
\filldraw[fill=viridisA,draw=none] (2.200,11.400) rectangle (2.520,12.416);
\filldraw[fill=viridisB,draw=none] (2.200,12.416) rectangle (2.520,13.773);
\filldraw[fill=viridisC,draw=none] (2.200,13.773) rectangle (2.520,14.200);
\filldraw[pattern={Dots[distance=1.5pt, radius=0.6pt]},pattern color=viridisA,draw=none] (2.840,11.400) rectangle (3.160,11.893);
\filldraw[pattern={Dots[distance=1.5pt, radius=0.6pt]},pattern color=viridisB,draw=none] (2.840,11.893) rectangle (3.160,13.245);
\filldraw[pattern={Dots[distance=1.5pt, radius=0.6pt]},pattern color=viridisC,draw=none] (2.840,13.245) rectangle (3.160,14.200);
\draw[line width=0.4pt] (4.400,11.400) rectangle (4.720,14.200);
\draw[line width=0.4pt] (5.040,11.400) rectangle (5.360,14.200);
\filldraw[fill=viridisA,draw=none] (4.400,11.400) rectangle (4.720,12.395);
\filldraw[fill=viridisB,draw=none] (4.400,12.395) rectangle (4.720,13.803);
\filldraw[fill=viridisC,draw=none] (4.400,13.803) rectangle (4.720,14.200);
\filldraw[pattern={Dots[distance=1.5pt, radius=0.6pt]},pattern color=viridisA,draw=none] (5.040,11.400) rectangle (5.360,11.576);
\filldraw[pattern={Dots[distance=1.5pt, radius=0.6pt]},pattern color=viridisB,draw=none] (5.040,11.576) rectangle (5.360,12.988);
\filldraw[pattern={Dots[distance=1.5pt, radius=0.6pt]},pattern color=viridisC,draw=none] (5.040,12.988) rectangle (5.360,14.200);
\draw[line width=0.4pt] (6.600,11.400) rectangle (6.920,14.200);
\draw[line width=0.4pt] (7.240,11.400) rectangle (7.560,14.200);
\filldraw[fill=viridisA,draw=none] (6.600,11.400) rectangle (6.920,12.078);
\filldraw[fill=viridisB,draw=none] (6.600,12.078) rectangle (6.920,13.380);
\filldraw[fill=viridisC,draw=none] (6.600,13.380) rectangle (6.920,14.200);
\filldraw[pattern={Dots[distance=1.5pt, radius=0.6pt]},pattern color=viridisA,draw=none] (7.240,11.400) rectangle (7.560,11.470);
\filldraw[pattern={Dots[distance=1.5pt, radius=0.6pt]},pattern color=viridisB,draw=none] (7.240,11.470) rectangle (7.560,12.307);
\filldraw[pattern={Dots[distance=1.5pt, radius=0.6pt]},pattern color=viridisC,draw=none] (7.240,12.307) rectangle (7.560,14.200);

\node[anchor=east,font=\small] at (-1.200,9.000) {ColQwen2.5 (Doc Ctx)};
\draw[line width=0.4pt] (0.000,7.600) rectangle (0.320,10.400);
\draw[line width=0.4pt] (0.640,7.600) rectangle (0.960,10.400);
\filldraw[fill=viridisA,draw=none] (0.000,7.600) rectangle (0.320,8.172);
\filldraw[fill=viridisB,draw=none] (0.000,8.172) rectangle (0.320,9.027);
\filldraw[fill=viridisC,draw=none] (0.000,9.027) rectangle (0.320,10.400);
\filldraw[pattern={Dots[distance=1.5pt, radius=0.6pt]},pattern color=viridisA,draw=none] (0.640,7.600) rectangle (0.960,8.047);
\filldraw[pattern={Dots[distance=1.5pt, radius=0.6pt]},pattern color=viridisB,draw=none] (0.640,8.047) rectangle (0.960,8.793);
\filldraw[pattern={Dots[distance=1.5pt, radius=0.6pt]},pattern color=viridisC,draw=none] (0.640,8.793) rectangle (0.960,10.400);
\draw[line width=0.4pt] (2.200,7.600) rectangle (2.520,10.400);
\draw[line width=0.4pt] (2.840,7.600) rectangle (3.160,10.400);
\filldraw[fill=viridisA,draw=none] (2.200,7.600) rectangle (2.520,7.961);
\filldraw[fill=viridisB,draw=none] (2.200,7.961) rectangle (2.520,9.811);
\filldraw[fill=viridisC,draw=none] (2.200,9.811) rectangle (2.520,10.400);
\filldraw[pattern={Dots[distance=1.5pt, radius=0.6pt]},pattern color=viridisA,draw=none] (2.840,7.600) rectangle (3.160,7.677);
\filldraw[pattern={Dots[distance=1.5pt, radius=0.6pt]},pattern color=viridisB,draw=none] (2.840,7.677) rectangle (3.160,9.434);
\filldraw[pattern={Dots[distance=1.5pt, radius=0.6pt]},pattern color=viridisC,draw=none] (2.840,9.434) rectangle (3.160,10.400);
\draw[line width=0.4pt] (4.400,7.600) rectangle (4.720,10.400);
\draw[line width=0.4pt] (5.040,7.600) rectangle (5.360,10.400);
\filldraw[fill=viridisA,draw=none] (4.400,7.600) rectangle (4.720,8.457);
\filldraw[fill=viridisB,draw=none] (4.400,8.457) rectangle (4.720,10.204);
\filldraw[fill=viridisC,draw=none] (4.400,10.204) rectangle (4.720,10.400);
\filldraw[pattern={Dots[distance=1.5pt, radius=0.6pt]},pattern color=viridisA,draw=none] (5.040,7.600) rectangle (5.360,7.907);
\filldraw[pattern={Dots[distance=1.5pt, radius=0.6pt]},pattern color=viridisB,draw=none] (5.040,7.907) rectangle (5.360,9.870);
\filldraw[pattern={Dots[distance=1.5pt, radius=0.6pt]},pattern color=viridisC,draw=none] (5.040,9.870) rectangle (5.360,10.400);
\draw[line width=0.4pt] (6.600,7.600) rectangle (6.920,10.400);
\draw[line width=0.4pt] (7.240,7.600) rectangle (7.560,10.400);
\filldraw[fill=viridisA,draw=none] (6.600,7.600) rectangle (6.920,7.985);
\filldraw[fill=viridisB,draw=none] (6.600,7.985) rectangle (6.920,9.990);
\filldraw[fill=viridisC,draw=none] (6.600,9.990) rectangle (6.920,10.400);
\filldraw[pattern={Dots[distance=1.5pt, radius=0.6pt]},pattern color=viridisA,draw=none] (7.240,7.600) rectangle (7.560,7.687);
\filldraw[pattern={Dots[distance=1.5pt, radius=0.6pt]},pattern color=viridisB,draw=none] (7.240,7.687) rectangle (7.560,9.523);
\filldraw[pattern={Dots[distance=1.5pt, radius=0.6pt]},pattern color=viridisC,draw=none] (7.240,9.523) rectangle (7.560,10.400);

\node[anchor=east,font=\small] at (-1.200,5.200) {ColQwen2.5 (Q-Split)};
\draw[line width=0.4pt] (0.000,3.800) rectangle (0.320,6.600);
\draw[line width=0.4pt] (0.640,3.800) rectangle (0.960,6.600);
\filldraw[fill=viridisA,draw=none] (0.000,3.800) rectangle (0.320,4.299);
\filldraw[fill=viridisB,draw=none] (0.000,4.299) rectangle (0.320,4.978);
\filldraw[fill=viridisC,draw=none] (0.000,4.978) rectangle (0.320,6.600);
\filldraw[pattern={Dots[distance=1.5pt, radius=0.6pt]},pattern color=viridisA,draw=none] (0.640,3.800) rectangle (0.960,4.065);
\filldraw[pattern={Dots[distance=1.5pt, radius=0.6pt]},pattern color=viridisB,draw=none] (0.640,4.065) rectangle (0.960,4.539);
\filldraw[pattern={Dots[distance=1.5pt, radius=0.6pt]},pattern color=viridisC,draw=none] (0.640,4.539) rectangle (0.960,6.600);
\draw[line width=0.4pt] (2.200,3.800) rectangle (2.520,6.600);
\draw[line width=0.4pt] (2.840,3.800) rectangle (3.160,6.600);
\filldraw[fill=viridisA,draw=none] (2.200,3.800) rectangle (2.520,4.604);
\filldraw[fill=viridisB,draw=none] (2.200,4.604) rectangle (2.520,5.886);
\filldraw[fill=viridisC,draw=none] (2.200,5.886) rectangle (2.520,6.600);
\filldraw[pattern={Dots[distance=1.5pt, radius=0.6pt]},pattern color=viridisA,draw=none] (2.840,3.800) rectangle (3.160,4.050);
\filldraw[pattern={Dots[distance=1.5pt, radius=0.6pt]},pattern color=viridisB,draw=none] (2.840,4.050) rectangle (3.160,4.861);
\filldraw[pattern={Dots[distance=1.5pt, radius=0.6pt]},pattern color=viridisC,draw=none] (2.840,4.861) rectangle (3.160,6.600);
\draw[line width=0.4pt] (4.400,3.800) rectangle (4.720,6.600);
\draw[line width=0.4pt] (5.040,3.800) rectangle (5.360,6.600);
\filldraw[fill=viridisA,draw=none] (4.400,3.800) rectangle (4.720,4.506);
\filldraw[fill=viridisB,draw=none] (4.400,4.506) rectangle (4.720,5.791);
\filldraw[fill=viridisC,draw=none] (4.400,5.791) rectangle (4.720,6.600);
\filldraw[pattern={Dots[distance=1.5pt, radius=0.6pt]},pattern color=viridisA,draw=none] (5.040,3.800) rectangle (5.360,3.861);
\filldraw[pattern={Dots[distance=1.5pt, radius=0.6pt]},pattern color=viridisB,draw=none] (5.040,3.861) rectangle (5.360,4.538);
\filldraw[pattern={Dots[distance=1.5pt, radius=0.6pt]},pattern color=viridisC,draw=none] (5.040,4.538) rectangle (5.360,6.600);
\draw[line width=0.4pt] (6.600,3.800) rectangle (6.920,6.600);
\draw[line width=0.4pt] (7.240,3.800) rectangle (7.560,6.600);
\filldraw[fill=viridisA,draw=none] (6.600,3.800) rectangle (6.920,4.404);
\filldraw[fill=viridisB,draw=none] (6.600,4.404) rectangle (6.920,5.285);
\filldraw[fill=viridisC,draw=none] (6.600,5.285) rectangle (6.920,6.600);
\filldraw[pattern={Dots[distance=1.5pt, radius=0.6pt]},pattern color=viridisA,draw=none] (7.240,3.800) rectangle (7.560,3.851);
\filldraw[pattern={Dots[distance=1.5pt, radius=0.6pt]},pattern color=viridisB,draw=none] (7.240,3.851) rectangle (7.560,4.074);
\filldraw[pattern={Dots[distance=1.5pt, radius=0.6pt]},pattern color=viridisC,draw=none] (7.240,4.074) rectangle (7.560,6.600);

\node[anchor=east,font=\small] at (-1.200,1.400) {ColQwen2.5 (Rerank)};
\draw[line width=0.4pt] (0.000,0.000) rectangle (0.320,2.800);
\draw[line width=0.4pt] (0.640,0.000) rectangle (0.960,2.800);
\filldraw[fill=viridisA,draw=none] (0.000,0.000) rectangle (0.320,0.443);
\filldraw[fill=viridisB,draw=none] (0.000,0.443) rectangle (0.320,1.072);
\filldraw[fill=viridisC,draw=none] (0.000,1.072) rectangle (0.320,2.800);
\filldraw[pattern={Dots[distance=1.5pt, radius=0.6pt]},pattern color=viridisA,draw=none] (0.640,0.000) rectangle (0.960,0.238);
\filldraw[pattern={Dots[distance=1.5pt, radius=0.6pt]},pattern color=viridisB,draw=none] (0.640,0.238) rectangle (0.960,0.634);
\filldraw[pattern={Dots[distance=1.5pt, radius=0.6pt]},pattern color=viridisC,draw=none] (0.640,0.634) rectangle (0.960,2.800);
\draw[line width=0.4pt] (2.200,0.000) rectangle (2.520,2.800);
\draw[line width=0.4pt] (2.840,0.000) rectangle (3.160,2.800);
\filldraw[fill=viridisA,draw=none] (2.200,0.000) rectangle (2.520,0.853);
\filldraw[fill=viridisB,draw=none] (2.200,0.853) rectangle (2.520,2.230);
\filldraw[fill=viridisC,draw=none] (2.200,2.230) rectangle (2.520,2.800);
\filldraw[pattern={Dots[distance=1.5pt, radius=0.6pt]},pattern color=viridisA,draw=none] (2.840,0.000) rectangle (3.160,0.398);
\filldraw[pattern={Dots[distance=1.5pt, radius=0.6pt]},pattern color=viridisB,draw=none] (2.840,0.398) rectangle (3.160,1.633);
\filldraw[pattern={Dots[distance=1.5pt, radius=0.6pt]},pattern color=viridisC,draw=none] (2.840,1.633) rectangle (3.160,2.800);
\draw[line width=0.4pt] (4.400,0.000) rectangle (4.720,2.800);
\draw[line width=0.4pt] (5.040,0.000) rectangle (5.360,2.800);
\filldraw[fill=viridisA,draw=none] (4.400,0.000) rectangle (4.720,0.808);
\filldraw[fill=viridisB,draw=none] (4.400,0.808) rectangle (4.720,2.242);
\filldraw[fill=viridisC,draw=none] (4.400,2.242) rectangle (4.720,2.800);
\filldraw[pattern={Dots[distance=1.5pt, radius=0.6pt]},pattern color=viridisA,draw=none] (5.040,0.000) rectangle (5.360,0.102);
\filldraw[pattern={Dots[distance=1.5pt, radius=0.6pt]},pattern color=viridisB,draw=none] (5.040,0.102) rectangle (5.360,1.277);
\filldraw[pattern={Dots[distance=1.5pt, radius=0.6pt]},pattern color=viridisC,draw=none] (5.040,1.277) rectangle (5.360,2.800);
\draw[line width=0.4pt] (6.600,0.000) rectangle (6.920,2.800);
\draw[line width=0.4pt] (7.240,0.000) rectangle (7.560,2.800);
\filldraw[fill=viridisA,draw=none] (6.600,0.000) rectangle (6.920,0.561);
\filldraw[fill=viridisB,draw=none] (6.600,0.561) rectangle (6.920,1.760);
\filldraw[fill=viridisC,draw=none] (6.600,1.760) rectangle (6.920,2.800);
\filldraw[pattern={Dots[distance=1.5pt, radius=0.6pt]},pattern color=viridisA,draw=none] (7.240,0.000) rectangle (7.560,0.051);
\filldraw[pattern={Dots[distance=1.5pt, radius=0.6pt]},pattern color=viridisB,draw=none] (7.240,0.051) rectangle (7.560,0.627);
\filldraw[pattern={Dots[distance=1.5pt, radius=0.6pt]},pattern color=viridisC,draw=none] (7.240,0.627) rectangle (7.560,2.800);

\node[anchor=west,font=\small] at (1.1,-0.850) {Pages};
\filldraw[fill=viridisB,draw=none] (2.3,-0.970) rectangle (2.8,-0.730);

\node[anchor=west,font=\small] at (3.5,-0.850) {Documents};
\filldraw[pattern={Dots[distance=1.5pt, radius=0.6pt]},pattern color=viridisB] (5.6,-0.970) rectangle (6.1,-0.730);

\filldraw[fill=viridisA,draw=black,line width=0.2pt] (0.000,-2.100) rectangle (0.420,-1.920);
\node[anchor=west] at (0.560,-2.010) {None};
\filldraw[fill=viridisB,draw=black,line width=0.2pt] (2.500,-2.100) rectangle (2.920,-1.920);
\node[anchor=west] at (3.060,-2.010) {Partial};
\filldraw[fill=viridisC,draw=black,line width=0.2pt] (5.000,-2.100) rectangle (5.420,-1.920);
\node[anchor=west] at (5.560,-2.010) {All};
\end{tikzpicture}

    \caption{Analysis of Cross-Doc query retrieval results for each data source and the models ColQwen2.5 \cite{faysse2024colpali}, ColQwen2.5 (Doc Ctx), ColQwen2.5 (Q-Split), and ColQwen2.5 (Rerank). Instead of just deciding whether a retrieval is successful, we breakdown the results into page-level (filled bars) and document-level (dotted bars) evaluation. Additionally, each level is divided into the three classes that all evidences could be retrieved, partially, or none which is indicated by the colors.}
    \label{fig:cross_document_colqwen_retrieval_results_breakdown}
\end{figure}

We analyze in this subsection what potential causes might be for the low performance in micro-average on Cross-Doc queries. Figure \ref{fig:cross_document_colqwen_retrieval_results_breakdown} shows a detailed cross-document retrieval performance breakdown for each data source for the models ColQwen2.5 \cite{faysse2024colpali}, ColQwen2.5 (Doc Ctx), ColQwen2.5 (Q-Split), and ColQwen2.5 (Rerank). The retrieval performance breakdown is not just binary, but also considers the retrieval on a page-level (filled bars) and document-level (dotted bars). By page-level, we mean in this case whether the evidence pages are correctly retrieved, and by document-level, whether the documents of the pages are correctly retrieved. Additionally, we measure for both levels whether all, partial, or no evidence could be retrieved. The aim of this analysis is to find out whether the retrieval models even struggle to find the correct documents. As Figure \ref{fig:cross_document_colqwen_retrieval_results_breakdown} shows, all models have both for the page-level and document-level retrieval a high percentage of retrieving at least one correct evidence item. This speaks against the possibility that there might be multiple other documents which are evidence for a query. As already noted in previous sections, Cross-Doc query retrieval on the Wikipedia data source exhibits a relatively high performance in contrast to the three other data sources SEC Proxy Statements, arXiv, and PubMed. We observe that the breakdown distribution heavily depends on the retrieval model, while the breakdown distribution is consistent across the three other data sources for a given retrieval model. The base model ColQwen2.5 \cite{faysse2024colpali} and ColQwen2.5 (Rerank) display a similar performance for SEC Proxy Statements and arXiv articles with roughly the same percentage of complete and partial document-level retrieval. For page-level retrieval, this proportion shifts towards partial evidence retrieval. ColQwen2.5 (Doc Ctx) has even a higher percentage of partial retrieval both for page-level and document-level retrieval. A likely explanation for this behavior is that the document context embeddings bias the page retrieval to the pages of the most similar document to the query. ColQwen2.5 (Q-Split) has a high document-level retrieval success percentage, whereas at the page-level the retrieval percentage of all ground truth evidence is about half of the document-level retrieval percentage. The relatively high page-level partial evidence retrieval might be due to long documents. The lowest successful complete evidence retrieval for SEC Proxy Statements, which have the highest document page count supports this. However, this might also indicate that there are alternative document pages containing the evidence for a query. Overall, the query splitting achieves the best retrieval performance for the ColQwen2.5 \cite{faysse2024colpali} base model.

\subsection{Difference between Query Generators}
We compare queries generated by GPT-4.1-mini (59,983 queries)
and Qwen3-VL-32B-Instruct~\cite{bai2025qwen3} (211,777 queries)
to examine the effect of query generation on retrieval performance.
All methods retrieve from the complete 388,548-page corpus.
Within each generator subset, Macro assigns equal weight to the
six evidence structures, while Micro averages over all queries.
Both protocols retain each generator's original query distribution.

\begin{figure}[htbp]
\centering

\providecolor{colorColPali}{RGB}{228,26,28}
\providecolor{colorColQwen25}{RGB}{55,126,184}
\providecolor{colorColNomic}{RGB}{77,175,74}
\providecolor{colorColSmol}{RGB}{152,78,163}
\providecolor{colorColModernVBERT}{RGB}{255,127,0}

\begin{tikzpicture}
\begin{axis}[
    width=0.82\linewidth,
    height=0.82\linewidth,
    xlabel={Macro RecallAll@10 (GPT-4.1-mini)},
    ylabel={Macro RecallAll@10 (Qwen3-VL-32B-Instruct)},
    xmin=47, xmax=81,
    ymin=47, ymax=81,
    axis equal image,
    grid=both,
    major grid style={line width=0.35pt, draw=gray!35},
    minor grid style={line width=0.2pt, draw=gray!20},
    tick align=outside,
    tick style={black},
    axis line style={black},
    label style={font=\small},
    tick label style={font=\small},
    legend style={
        at={(0.99,0.01)},
        anchor=south east,
        font=\scriptsize,
        draw=black,
        fill=white
    },
    clip=false
]

\addplot[
    domain=47:81,
    samples=2,
    thick,
    dashed,
    black
] {x};
\addlegendentry{$y=x$}

\addplot+[
    only marks,
    mark=*,
    mark size=2.6pt,
    draw=black,
    fill=colorColQwen25,
    mark options={draw=black,fill=colorColQwen25}
] coordinates {
    (71.7,66.4)
    (76.8,76.8)
    (76.7,76.8)
    (76.7,75.6)
    (74.8,73.6)
    (73.2,68.4)
    (75.9,70.8)
};
\addlegendentry{ColQwen2.5}

\addplot+[
    only marks,
    mark=*,
    mark size=2.6pt,
    draw=black,
    fill=colorColNomic,
    mark options={draw=black,fill=colorColNomic}
] coordinates {
    (76.2,70.9)
    (79.2,78.0)
    (78.6,73.9)
};
\addlegendentry{ColNomic}

\addplot+[
    only marks,
    mark=*,
    mark size=2.6pt,
    draw=black,
    fill=colorColPali,
    mark options={draw=black,fill=colorColPali}
] coordinates {
    (54.7,48.2)
    (69.0,65.1)
    (63.2,56.1)
};
\addlegendentry{ColPali v1.3}

\addplot+[
    only marks,
    mark=*,
    mark size=2.6pt,
    draw=black,
    fill=colorColSmol,
    mark options={draw=black,fill=colorColSmol}
] coordinates {
    (62.2,57.4)
};
\addlegendentry{ColSmol 500}

\addplot+[
    only marks,
    mark=*,
    mark size=2.6pt,
    draw=black,
    fill=colorColModernVBERT,
    mark options={draw=black,fill=colorColModernVBERT}
] coordinates {
    (59.5,53.5)
};
\addlegendentry{ColModernVBERT}


\node[anchor=east, font=\scriptsize] at (axis cs:62.2,57.4) {\hspace{2pt}ColSmol 500};
\node[anchor=west, font=\scriptsize] at (axis cs:59.5,53.5) {\hspace{2pt}ColModernVBERT};
\node[anchor=east, font=\scriptsize] at (axis cs:54.7,48.2) {\hspace{2pt}ColPali v1.3};
\node[anchor=west, font=\scriptsize] at (axis cs:71.7,66.4) {\hspace{2pt}ColQwen2.5};
\node[anchor=west, font=\scriptsize] at (axis cs:76.2,70.9) {\hspace{2pt}ColNomic};
\node[anchor=north, font=\scriptsize] at (axis cs:69.0,65.1) {\hspace{2pt}ColPali v1.3 (Ctx512)};
\node[anchor=south east, font=\scriptsize] at (axis cs:76.8,76.8) {\hspace{2pt}ColQwen2.5 (Ctx128)};
\node[anchor=east, font=\scriptsize] at (axis cs:76.7,76.8) {\hspace{2pt}ColQwen2.5 (Ctx256)};
\node[anchor=north east, font=\scriptsize] at (axis cs:76.7,75.6) {\hspace{2pt}ColQwen2.5 (Ctx512)};
\node[anchor=east, font=\scriptsize] at (axis cs:74.8,73.6) {\hspace{2pt}ColQwen2.5 (Ctx1024)};
\node[anchor=south east, font=\scriptsize] at (axis cs:79.2,78.0) {\hspace{2pt}ColNomic (Ctx512)};
\node[anchor=west, font=\scriptsize] at (axis cs:73.2,68.4) {\hspace{2pt}ColQwen2.5 (Q-Split)};
\node[anchor=east, font=\scriptsize] at (axis cs:63.2,56.1) {\hspace{2pt}ColPali v1.3 (Rerank)};
\node[anchor=east, font=\scriptsize] at (axis cs:75.9,70.8) {\hspace{2pt}ColQwen2.5 (Rerank)};
\node[anchor=west, font=\scriptsize] at (axis cs:78.6,73.9) {\hspace{2pt}ColNomic (Rerank)};

\end{axis}
\end{tikzpicture}
\caption{Comparison of structure-balanced Macro RecallAll@10
between queries generated by GPT-4.1-mini and
Qwen3-VL-32B-Instruct. Each point represents a retrieval
configuration evaluated on the same full corpus. Colors indicate
the model family, and the dashed line denotes equal scores.}
\label{fig:recallall10_openai_vs_qwen}
\end{figure}
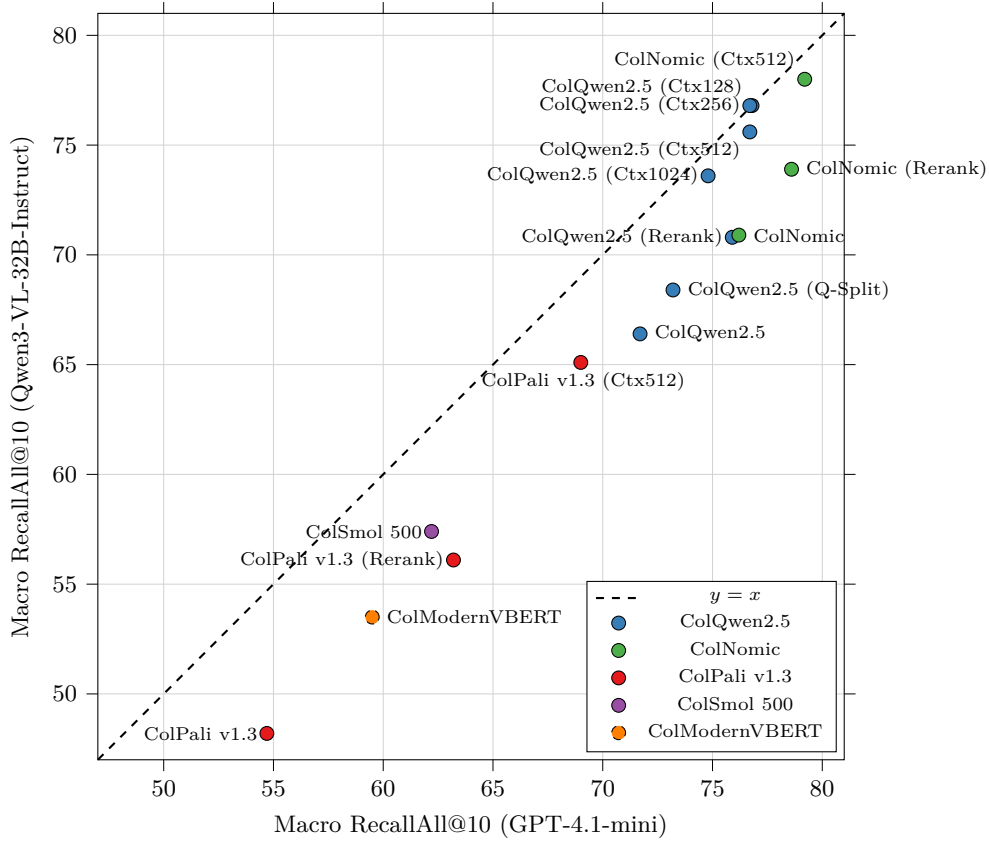

Figure~\ref{fig:recallall10_openai_vs_qwen} compares Macro
RecallAll@10 across the visual retrieval configurations.
Most configurations obtain higher scores on GPT-4.1-mini queries,
with differences ranging from $-0.1$ to $7.1$ percentage points.
For ColPali, ColQwen2.5, and ColNomic, document-context retrieval
exhibits smaller differences between generators than the corresponding
base and reranking settings. For example, the ColNomic difference
decreases from 5.3 percentage points for base retrieval and 4.8
for reranking to 1.2 for document-context retrieval.

Tables~\ref{tab:generator_gpt_full} and
\ref{tab:generator_qwen_full} report nDCG, Recall, nDCGAll,
and RecallAll for $k\in\{3,5,10\}$ for the same nine visual,
text, and hybrid configurations under both aggregation protocols.
At $k=10$, all nine configurations obtain higher nDCGAll and
RecallAll on GPT-4.1-mini queries than on Qwen3-VL-32B-Instruct
queries under both Macro and Micro. ColNomic (Ensemble) achieves
the highest scores among the nine configurations for both
generators at all evaluated values of $k$. Its Macro nDCGAll@10
and RecallAll@10 are 71.51 and 81.74 on GPT-4.1-mini queries,
compared with 68.85 and 80.46 on Qwen3-VL-32B-Instruct queries.
The corresponding Micro scores are 78.05 and 87.80, and 74.96
and 85.78, respectively.
OCR-BM25 achieves the highest nDCGAll@10 and RecallAll@10
among the three text baselines under both aggregation protocols
for each generator. Its Macro RecallAll@10 is 64.19 on
GPT-4.1-mini queries and 61.28 on Qwen3-VL-32B-Instruct queries.
For ColNomic (Ctx), RRF fusion with OCR-BM25 reduces Macro
RecallAll@10 from 79.20 to 79.14 on GPT-4.1-mini queries, but
increases it from 78.03 to 78.61 on Qwen3-VL-32B-Instruct queries.
Under Micro, the same fusion increases RecallAll@10 from 84.21
to 85.70 and from 81.70 to 83.82, respectively.
These comparisons retain the source and evidence-structure
composition of each generator subset. The 4,152-query subset
balanced across generators and evidence structures is evaluated
separately to examine the effect of query composition. Differences
in retrieval performance may reflect both query formulation and
dataset composition; they do not by themselves establish
differences in annotation correctness.

\begin{table*}[t]
\centering
\resizebox{\textwidth}{!}{
\setlength{\tabcolsep}{2pt}
\renewcommand{\arraystretch}{0.94}
\begin{tabular}{@{}llrrr rrr rrr rrr@{}}
\toprule
\multirow{2}{*}{Model / Setting} & \multirow{2}{*}{Aggregate}
& \multicolumn{3}{c}{nDCG@k} & \multicolumn{3}{c}{Recall@k}
& \multicolumn{3}{c}{nDCGAll@k} & \multicolumn{3}{c}{RecallAll@k} \\
\cmidrule(lr){3-5}\cmidrule(lr){6-8}\cmidrule(lr){9-11}\cmidrule(l){12-14}
& & @3 & @5 & @10 & @3 & @5 & @10 & @3 & @5 & @10 & @3 & @5 & @10 \\
\midrule
\multirow{2}{*}{ColNomic (Ensemble)} & Macro & 70.58 & 74.42 & 76.89 & 74.01 & 81.76 & 88.02 & 57.31 & 65.67 & 71.51 & 61.16 & 72.66 & 81.74 \\
 & Micro & 77.58 & 79.53 & 80.87 & 83.24 & 87.57 & 91.22 & 73.31 & 75.94 & 78.05 & 79.02 & 83.67 & 87.80 \\
\multirow{2}{*}{ColNomic} & Macro & 67.41 & 70.93 & 73.39 & 70.64 & 77.83 & 84.12 & 53.60 & 60.91 & 66.34 & 57.39 & 67.55 & 76.19 \\
 & Micro & 73.27 & 75.35 & 76.85 & 78.92 & 83.60 & 87.78 & 68.95 & 71.61 & 73.85 & 74.68 & 79.63 & 84.23 \\
\multirow{2}{*}{ColNomic (Ctx)} & Macro & 65.72 & 70.13 & 73.18 & 68.99 & 77.99 & 85.78 & 51.46 & 60.75 & 67.74 & 55.26 & 68.07 & 79.20 \\
 & Micro & 72.01 & 74.40 & 76.08 & 78.07 & 83.50 & 88.21 & 67.59 & 70.52 & 72.81 & 73.75 & 79.30 & 84.21 \\
\multirow{2}{*}{ColNomic (Rerank)} & Macro & 69.89 & 73.46 & 75.72 & 72.92 & 80.16 & 85.88 & 56.14 & 63.91 & 69.16 & 59.75 & 70.38 & 78.63 \\
 & Micro & 76.86 & 78.72 & 79.93 & 82.04 & 86.15 & 89.42 & 72.64 & 75.15 & 77.06 & 77.89 & 82.30 & 86.02 \\
\multirow{2}{*}{OCR-BM25} & Macro & 51.75 & 56.13 & 59.29 & 55.44 & 64.52 & 72.77 & 38.54 & 46.26 & 52.18 & 42.46 & 53.83 & 64.19 \\
 & Micro & 57.04 & 60.02 & 62.27 & 63.94 & 70.79 & 77.28 & 53.58 & 56.98 & 59.79 & 60.45 & 67.35 & 74.06 \\
\multirow{2}{*}{OCR-dense (BGE-M3)} & Macro & 25.75 & 28.31 & 30.38 & 27.72 & 33.02 & 38.44 & 16.35 & 19.88 & 23.27 & 18.27 & 23.72 & 29.85 \\
 & Micro & 24.76 & 26.45 & 27.95 & 28.23 & 32.16 & 36.50 & 22.56 & 24.31 & 25.92 & 25.99 & 29.75 & 33.92 \\
\multirow{2}{*}{DOM-BM25} & Macro & 49.41 & 53.56 & 56.64 & 52.85 & 61.43 & 69.47 & 36.58 & 43.78 & 49.53 & 40.29 & 50.84 & 60.79 \\
 & Micro & 52.75 & 55.46 & 57.60 & 59.10 & 65.31 & 71.47 & 49.33 & 52.41 & 55.11 & 55.65 & 61.90 & 68.24 \\
\multirow{2}{*}{ColNomic (Ctx) + OCR-BM25 (RRF)} & Macro & 63.55 & 68.21 & 71.42 & 67.87 & 77.44 & 85.69 & 49.81 & 59.19 & 66.28 & 54.30 & 67.67 & 79.14 \\
 & Micro & 69.89 & 72.74 & 74.68 & 77.31 & 83.80 & 89.25 & 65.89 & 69.27 & 71.94 & 73.27 & 79.90 & 85.70 \\
\multirow{2}{*}{ColNomic (Rerank) + OCR-BM25 (RRF)} & Macro & 62.82 & 67.33 & 70.69 & 67.25 & 76.50 & 85.09 & 49.32 & 57.54 & 64.57 & 53.87 & 65.72 & 77.54 \\
 & Micro & 69.52 & 72.50 & 74.59 & 76.91 & 83.68 & 89.48 & 65.58 & 69.15 & 72.04 & 72.93 & 79.85 & 86.10 \\
\bottomrule
\end{tabular}
}
\caption{\textbf{Retrieval results on GPT-4.1-mini queries} (\%; 59,983 queries). Macro gives equal weight to the six evidence structures within this subset, and Micro averages over its queries. All nine configurations retrieve from the same 388,548-page corpus.}
\label{tab:generator_gpt_full}
\end{table*}

\begin{table*}[t]
\centering
\resizebox{\textwidth}{!}{
\setlength{\tabcolsep}{2pt}
\renewcommand{\arraystretch}{0.94}
\begin{tabular}{@{}llrrr rrr rrr rrr@{}}
\toprule
\multirow{2}{*}{Model / Setting} & \multirow{2}{*}{Aggregate}
& \multicolumn{3}{c}{nDCG@k} & \multicolumn{3}{c}{Recall@k}
& \multicolumn{3}{c}{nDCGAll@k} & \multicolumn{3}{c}{RecallAll@k} \\
\cmidrule(lr){3-5}\cmidrule(lr){6-8}\cmidrule(lr){9-11}\cmidrule(l){12-14}
& & @3 & @5 & @10 & @3 & @5 & @10 & @3 & @5 & @10 & @3 & @5 & @10 \\
\midrule
\multirow{2}{*}{ColNomic (Ensemble)} & Macro & 67.97 & 72.32 & 75.15 & 71.63 & 80.46 & 87.68 & 52.85 & 62.02 & 68.85 & 56.71 & 69.57 & 80.46 \\
 & Micro & 74.26 & 76.79 & 78.51 & 79.84 & 85.37 & 90.04 & 68.32 & 72.03 & 74.96 & 73.88 & 80.17 & 85.78 \\
\multirow{2}{*}{ColNomic} & Macro & 63.15 & 66.96 & 69.39 & 66.12 & 73.90 & 80.13 & 47.82 & 55.53 & 60.95 & 51.43 & 62.24 & 70.93 \\
 & Micro & 69.67 & 72.02 & 73.64 & 75.11 & 80.32 & 84.80 & 63.96 & 67.19 & 69.78 & 69.59 & 75.28 & 80.39 \\
\multirow{2}{*}{ColNomic (Ctx)} & Macro & 64.64 & 69.19 & 72.42 & 68.08 & 77.40 & 85.73 & 48.87 & 58.20 & 65.69 & 52.87 & 65.97 & 78.03 \\
 & Micro & 69.11 & 71.97 & 74.03 & 75.03 & 81.42 & 87.14 & 62.34 & 66.19 & 69.29 & 68.53 & 75.42 & 81.70 \\
\multirow{2}{*}{ColNomic (Rerank)} & Macro & 66.18 & 69.88 & 72.16 & 69.12 & 76.64 & 82.43 & 51.47 & 59.03 & 64.19 & 54.98 & 65.58 & 73.86 \\
 & Micro & 72.89 & 75.07 & 76.54 & 77.78 & 82.54 & 86.51 & 67.25 & 70.36 & 72.83 & 72.28 & 77.59 & 82.28 \\
\multirow{2}{*}{OCR-BM25} & Macro & 51.25 & 55.30 & 58.36 & 54.95 & 63.36 & 71.36 & 37.44 & 44.33 & 49.93 & 41.33 & 51.61 & 61.28 \\
 & Micro & 55.78 & 58.91 & 61.32 & 62.35 & 69.46 & 76.25 & 50.97 & 54.70 & 57.83 & 57.53 & 64.80 & 71.85 \\
\multirow{2}{*}{OCR-dense (BGE-M3)} & Macro & 22.34 & 24.32 & 25.84 & 23.96 & 28.11 & 32.11 & 13.71 & 16.34 & 18.59 & 15.47 & 19.59 & 23.69 \\
 & Micro & 22.98 & 24.53 & 25.75 & 26.21 & 29.75 & 33.29 & 20.31 & 21.92 & 23.28 & 23.57 & 26.92 & 30.32 \\
\multirow{2}{*}{DOM-BM25} & Macro & 47.55 & 51.14 & 53.95 & 50.84 & 58.29 & 65.62 & 35.06 & 41.34 & 46.53 & 38.62 & 47.96 & 56.86 \\
 & Micro & 50.65 & 53.41 & 55.55 & 56.50 & 62.79 & 68.83 & 46.39 & 49.72 & 52.50 & 52.26 & 58.74 & 65.01 \\
\multirow{2}{*}{ColNomic (Ctx) + OCR-BM25 (RRF)} & Macro & 62.71 & 67.57 & 71.17 & 66.94 & 76.93 & 86.19 & 47.47 & 56.74 & 64.89 & 51.90 & 65.24 & 78.61 \\
 & Micro & 67.13 & 70.47 & 72.86 & 74.34 & 81.82 & 88.44 & 61.24 & 65.51 & 69.13 & 68.47 & 76.35 & 83.82 \\
\multirow{2}{*}{ColNomic (Rerank) + OCR-BM25 (RRF)} & Macro & 61.08 & 65.53 & 68.77 & 65.04 & 74.21 & 82.53 & 46.29 & 54.44 & 61.32 & 50.54 & 62.35 & 73.78 \\
 & Micro & 66.67 & 69.80 & 72.05 & 73.54 & 80.56 & 86.77 & 61.19 & 65.20 & 68.58 & 68.09 & 75.51 & 82.51 \\
\bottomrule
\end{tabular}
}
\caption{\textbf{Retrieval results on Qwen3-VL-32B-Instruct queries} (\%; 211,777 queries). Macro gives equal weight to the six evidence structures within this subset, and Micro averages over its queries. All nine configurations retrieve from the same 388,548-page corpus.}
\label{tab:generator_qwen_full}
\end{table*}


\end{document}